%% file: main.tex
\documentclass[11pt, a4paper, logo, copyright]{googlecloud}

\pdftrailerid{redacted}

\usepackage{dsfont}
\usepackage[numbers, sort&compress, super, comma]{natbib}

\usepackage[utf8]{inputenc} %
\usepackage[T1]{fontenc}    %
\usepackage{hyperref}       %
\hypersetup{
  colorlinks,
  citecolor=blue,
  linkcolor=red,
  urlcolor=blue}
\usepackage{nicefrac}       %
\usepackage{microtype}      %
\usepackage{wrapfig} %
\usepackage{soul} %
\usepackage{tikz, adjustbox}
\usepackage[capitalize,noabbrev]{cleveref}
\usepackage{fancyvrb}
\usepackage[colorinlistoftodos]{todonotes}
\usepackage{rotating}
\usepackage{inconsolata}
\usepackage[skins,breakable,most]{tcolorbox}

\usepackage{caption}
\usepackage{subcaption}
\usepackage{enumitem}
\usepackage{flushend}
\usepackage{balance}
\usepackage{lineno}
\usepackage{amsmath,amssymb,amsfonts}
\usepackage{algorithm}
\usepackage{algpseudocode}
\usepackage{graphicx}
\usepackage{textcomp}
\usepackage[table]{xcolor}
\usepackage{siunitx}
\usepackage{colortbl}
\usepackage{amsthm}
\usepackage{url}
\usepackage{float}
\usepackage{multirow}
\usepackage{multicol}
\usepackage{color}
\usepackage{bm}
\usepackage{bbm}

\usepackage{pifont}
\usepackage{wasysym} 
\usepackage{threeparttable}
\usepackage{longtable}
\usepackage{booktabs}

\usepackage{array}
\newcolumntype{L}[1]{>{\raggedright\let\newline\\\arraybackslash\hspace{0pt}}m{#1}}
\newcolumntype{C}[1]{>{\centering\let\newline  \\\arraybackslash\hspace{0pt}}m{#1}}
\newcolumntype{R}[1]{>{\raggedleft\let\newline \\\arraybackslash\hspace{0pt}}m{#1}}

\usepackage{comment} 
\usepackage{mdframed}
\usepackage{fontawesome}
\usepackage{csquotes}
\usepackage{makecell}

\definecolor{beigecolor}{RGB}{253, 244, 204}
\definecolor{greencolor}{RGB}{228, 242, 217}
\definecolor{bluecolor}{RGB}{66, 133, 244}
\definecolor{orgcolor}{RGB}{255, 140, 15}
\definecolor{redcolor}{RGB}{234, 67, 53}
\definecolor{ggreen}{RGB}{52, 168, 83}
\definecolor{gyellow}{RGB}{251, 188, 5}

\newcolumntype{Y}[1]{>{\centering\arraybackslash}p{#1}}

\lstdefinestyle{mystyle}{
    backgroundcolor=\color{backcolour},   
    commentstyle=\color{codegreen},
    keywordstyle=\color{magenta},
    numberstyle=\tiny\color{codegray},
    stringstyle=\color{codepurple},
    basicstyle=\ttfamily\scriptsize,
    breakatwhitespace=false,         
    breaklines=true,                 
    captionpos=b,                    
    keepspaces=true,                 
    numbers=left,                    
    numbersep=5pt,                  
    showspaces=false,                
    showstringspaces=false,
    showtabs=false,                  
    tabsize=2,
    frame=none,
    aboveskip=1pt,
    belowskip=1pt,
}
\lstdefinestyle{plainins}{
    backgroundcolor=\color{white},   
    commentstyle=\color{codegreen},
    keywordstyle=\color{magenta},
    numberstyle=\tiny\color{codegray},
    stringstyle=\color{codepurple},
    basicstyle=\ttfamily\scriptsize,
    breakatwhitespace=false,         
    breaklines=true,                 
    captionpos=b,                    
    keepspaces=true,                 
    numbers=none,                    
    numbersep=5pt,                  
    showspaces=false,                
    showstringspaces=false,
    showtabs=false,                  
    tabsize=2,
    aboveskip=0pt,
    belowskip=0pt,
    frame=single
}
\lstdefinestyle{plainexam}{
    backgroundcolor=\color[HTML]{FFFCF3},   
    commentstyle=\color{codegreen},
    keywordstyle=\color{magenta},
    numberstyle=\tiny\color{codegray},
    stringstyle=\color{codepurple},
    basicstyle=\ttfamily\scriptsize,
    breakatwhitespace=false,         
    breaklines=true,                 
    captionpos=b,                    
    keepspaces=true,                 
    numbers=none,                    
    numbersep=5pt,                  
    showspaces=false,                
    showstringspaces=false,
    showtabs=false,                  
    tabsize=2,
    aboveskip=0pt,
    belowskip=0pt
}

\tcbset{
  aibox/.style={
    width=\linewidth,
    top=5pt,
    bottom=1pt,
    colback=blue!6!white,
    colframe=black,
    colbacktitle=black,
    enhanced,
    center,
    fontupper=\scriptsize,
    attach boxed title to top left={yshift=-0.1in,xshift=0.15in},
    boxed title style={boxrule=0pt,colframe=white,},
  }
}
\newtcolorbox{AIbox}[2][]{aibox,title=#2,#1}
\definecolor{lightblue}{rgb}{0.22,0.45,0.70}

\usepackage{mathrsfs}%
\usepackage[title]{appendix}%
\usepackage{manyfoot}%
\usepackage{algorithmicx}
\usepackage{xspace}

\renewcommand{\figurename}{Fig.}

\newcommand{\supfig}[1]{Supplementary Fig.~\ref{#1}}
\newcommand{\supfigs}[1]{Supplementary Figs.~\ref{#1}}

\newcommand{\suptab}[1]{Supplementary Table~\ref{#1}}
\newcommand{\suptabs}[1]{Supplementary Tables~\ref{#1}}

\DeclareCaptionLabelSeparator{pipe}{ $|$ }
\DeclareCaptionLabelFormat{boldsimple}{\textbf{#1 #2}}
\renewcommand{\thesubsection}{\arabic{subsection}}
\renewcommand{\thesubsubsection}{\thesubsection.\arabic{subsubsection}}

\newcommand{\ourmethod}{{ERA}\xspace}
\newcommand{\gflumodel}{\texttt{Google\_SAI-FluEns}}
\newcommand{\gcovidmodel}{\texttt{Google\_SAI-Ensemble}}
\newcommand{\grsvmodel}{\texttt{Google\_SAI-RSVEns}}

\usepackage{listings}
\usepackage{placeins}

\lstdefinestyle{pythonstyle}{
    language=Python,
    backgroundcolor=\color{black!5},
    commentstyle=\color{green!40!black},
    keywordstyle=\color{blue},
    stringstyle=\color{purple},
    numberstyle=\tiny\color{gray},
    basicstyle=\ttfamily\footnotesize,
    breaklines=true,
    captionpos=b,
    numbers=left,
    showstringspaces=false,
}
\lstdefinestyle{mypythonstyle}{
    language=Python,
    basicstyle=\ttfamily\fontsize{5pt}{5.5pt}\selectfont,
    keywordstyle=\color{blue},
    stringstyle=\color{purple},
    commentstyle=\color{green!40!black},
    breaklines=true,
    showstringspaces=false,
    numbers=left,
    numberstyle=\fontsize{3pt}{4pt}\selectfont\color{gray},
    numbersep=5pt,
    frame=tb, 
    framerule=0.5pt,
}

\begin{abstract}
Probabilistic forecasting of infectious diseases is crucial for public health monitoring and preparedness but has historically relied on labor-intensive manual model curation by expert modeling teams. Bespoke development is an obstacle to scaling models to granular geographic resolutions or emerging pathogens. Here, we present an autonomous system---Empirical Research Assistance (\ourmethod)---using Large Language Model-guided tree search to iteratively generate, evaluate, and optimize executable forecasting software. Mostly, \ourmethod was provided the kernel of methods to implement, and in a few instances it was given license to develop ideas from scratch.  In a fully prospective, real-time evaluation during the 2025–26 United States respiratory season, \ourmethod generated methodologically diverse models for influenza, COVID-19, and respiratory syncytial virus (RSV). Aggregating a selection of these automatically-generated models yielded an ensemble that ranked first overall in official Centers for Disease Control and Prevention (CDC) post-season evaluations and outperformed the gold-standard CDC forecast ensembles across all three pathogens in a real-time prediction challenge. Additionally, \ourmethod successfully navigated data-scarce scenarios for RSV. Controlled retrospective ablations allowed for additional explorations of the impact of optimization target and instruction-following on forecast accuracy. Preliminary experiments indicate that \ourmethod's search optimization metric influences behavior of output forecasts, and using higher-tier language models in-search improves predictive performance. By autonomously generating interpretable code that creates epidemiologically relevant predictions, this framework provides a path toward rapidly scaling high-quality disease forecasting.
\end{abstract}

\title{Prospective multi-pathogen disease forecasting using autonomous LLM-guided tree search}
\author[1,2]{Sarah Martinson}
\author[1,2]{Michael P. Brenner}
\author[3]{Martyna Plomecka}
\author[1]{Brian P. Williams}
\author[1,4,$\dagger$]{Nicholas G. Reich}
\author[1,$\dagger$]{Zahra Shamsi}
\affil[1]{Google Research}
\affil[2]{School of Engineering and Applied Sciences, Harvard University}
\affil[3]{Google Deepmind}
\affil[4]{University of Massachusetts Amherst}

\newcommand{\copyrightext}{\footerfont 
$\dagger$ To whom correspondence should be addressed: shamsiz@google.com, nick@umass.edu}

\begin{document}
\maketitle

\section*{Introduction}
\input{Sections/1-intro}

\section*{Results}
\input{Sections/2-results}
\section*{Discussion}
\input{Sections/3-discussion}

\section*{Methods} \label{sec:methods}
\input{Sections/4-methods}

\section*{Data Availability}
All surveillance data used in this study are publicly available. Hospital admission data were obtained from the CDC's National Healthcare Safety Network (NHSN) Weekly Hospital Respiratory Data reporting system~\cite{cdc_nhsn_hrd_2025}. Outpatient syndromic surveillance data were obtained from the CDC's ILINet system\cite{cdc_ilinet_2026}. Forecast submissions from all participating teams are publicly available through the CDC FluSight\cite{flusight}, COVIDHub\cite{cramer2022covidhub}, and RSVHub\cite{rsvhub} repositories.

\section*{Code Availability}
ERA-generated model source codes for models that were submitted prospectively and their weekly submissions are publicly available at the Google Research GitHub repository, GoogleHub: \url{https://github.com/google-research/google-research/tree/master/epi_forecasts}. The same repository also contains prompts used to seed relevant \ourmethod runs.
The prospective analysis in this paper pulled forecasts from this repository, as well as from FluSight, COVIDHub and RSVHub. Analysis code is publicly available at \url{https://github.com/sarahmart/epi-hub-analysis}. This work follows all EPIFORGE 2020 \cite{pollett_epiforge_2021} recommended reporting items for epidemic forecasting and prediction research.


\setcounter{figure}{0}
\setcounter{table}{0}
\renewcommand{\figurename}{Supplementary Figure}
\renewcommand{\tablename}{Supplementary Table}

\appendix
\input{Sections/6-supp-figs}

\newpage
\input{Sections/7-prompt_tables}
\input{Sections/8-model_inventories}

\bibliographystyle{naturemag}
\bibliography{references}

\section*{Acknowledgements}
We are grateful to our colleagues in Google Research and Google DeepMind for the environment in which to do this work.  We are grateful for our coauthors in the original \ourmethod paper \cite{aygun_ai_2025}, and in particular would like to thank Matthew Abraham, Erica Brand, Marc Coram, Lizzie Dorfman for help and important discussions, as well as John Platt, Yossi Matias and James Manyika for their support and encouragement.

\section*{Author Contributions}
S.M. and Z.S. designed the study, executed the ERA forecasting pipelines, and conducted the primary data analysis. M.P. developed and analyzed the LLM judge mechanisms and instruction-following criteria. N.G.R. provided domain-specific epidemiological expertise, guided the selection of reference models, and validated method fidelity. B.P.W. provided strategic direction on the ERA architecture. M.P.B. directed and advised on the overall project. All authors contributed to the writing and editing of the manuscript.

\section*{Competing Interests}
S.M. carried out this work as part of a student researchership at Google Research. 
M.P.B. holds appointments at both Harvard University and Google Research. M.P. is an employee of Google DeepMind. B.P.W. and Z.S. are employees of Google Research. 
N.G.R. serves as a faculty member at the University of Massachusetts Amherst and was paid as a scientific consultant by Google to provide expertise for and advice on the experiments presented in this paper. His effort on this project was supported by Google, and aligned with UMass policies on external consulting. His engagement with Google was disclosed to and approved by the UMass Research Compliance office.

\end{document}

%% file: Sections/1-intro.tex

Forecasting disease outbreaks is of critical importance for public health monitoring and response planning. In the past 15 years, the sophistication of probabilistic forecasting for epidemiology has increased dramatically in the United States (US). 
With some centralization of epidemiological forecasting efforts beginning prior to 2020\cite{viboud_rapidd_2018,del_valle_summary_2018,johansson_open_2019,lutz_applying_2019}, the COVID-19 pandemic accelerated research and integration of forecasts into public health communication and decision-making\cite{shea_harnessing_2020,reich_collaborative_2022,mellor_forecasting_2025}.  Much of the innovation in these fields has been driven by collaborative modeling challenges run by governmental agencies or academic groups. 
These open challenges, many of which follow an emerging set of data standards for the field\cite{kerr_software_2026}, accept submissions from any research team, focus on transparency and careful evaluation, and typically integrate with public health decision-makers\cite{reich_collaborative_2022}.
This allows different research groups with their own modeling expertise and opinions to contribute to a collective effort to improve disease forecasting.

The United States Centers for Disease Control and Prevention (CDC) runs such collaborative forecasting hubs for major seasonal respiratory diseases, including influenza (FluSight\cite{flusight}), COVID-19 (COVIDHub\cite{cramer2022covidhub}), and, more recently, respiratory syncytial virus (RSV; RSVHub\cite{rsvhub}). Each week, updated surveillance data are released and participating teams submit standardized probabilistic forecasts for upcoming weeks. These hubs share a common operational structure and state-level public health surveillance data source, with weekly submissions and jurisdictional targets, with the main outcome incident hospital admissions. Prior seasons have shown that models contributed by academic, governmental and industry research teams routinely outperform simple reference models such as seasonal average and persistence models, a result that has been shown consistently across infectious disease forecasting challenges in the US, United Kingdom and France~\cite{mcgowan_collaborative_2019,reich_collaborative_2019,cramer_evaluation_2022,paireau_ensemble_2022,sherratt_predictive_2023,mellor_forecasting_2025}.

There is a wide diversity of different models submitted by individual teams: while some implement explicit mechanistic models of disease transmission (e.g., compartmental state-space formulations representing different stages of infection), others use methods from classical statistical frameworks (e.g., auto-regressive or spatio-temporal regression models) or machine learning (ML; e.g., random forests or neural networks). Additionally, many models hybridize these approaches, taking inspiration from a mechanistic understanding of disease transmission without explicitly building a compartmental model but rather implementing these concepts via statistical or ML techniques. Careful post-hoc evaluation has not shown consistent patterns of one kind of model outperforming others~\cite{mcgowan_collaborative_2019,reich_collaborative_2019,cramer_evaluation_2022,mathis_evaluation_2024}. However, probabilistic ensemble forecasts that combine these individual contributed models consistently rank near the top of leaderboards~\cite{mcgowan_collaborative_2019,reich_accuracy_2019,paireau_ensemble_2022,cramer_evaluation_2022,lopez_challenges_2024,mathis_evaluation_2024}, and thus serve as state-of-the-art forecasts for public health stakeholders.

This paper examines whether Large Language Models (LLMs) in an agentic harness can create epidemiological prediction models at the same level of skill as those created by teams in the forecast hubs. If so, this could have widespread implications for epidemiological forecasting: the current methodology is labor intensive, requiring many groups to build and maintain individual operational models. Under the current system, expanding CDC-style hubs to different countries and infections is labor intensive, with the main bottleneck being the lack of humans who can code appropriate and accurate epidemiological models\cite{kerr_software_2026, reich_collaborative_2022}. Automation could make it possible to do this at a larger scale by allowing public health officials with epidemiological expertise but not expert-level programming skills to orchestrate creation of a diverse array of models for a specific predictive challenge. 

We build on our previous work\cite{aygun_ai_2025}, which showed that an LLM-based tree search algorithm called Empirical Research Assistance (\ourmethod) demonstrated competitive performance on COVID-19 forecasting tasks compared to models written by experts when evaluated {\sl retrospectively}. \ourmethod uses an agentic harness based on Monte Carlo Tree Search, inspired by AlphaGo\cite{silver2017mastering}. By framing the creation of empirical software as a `scorable task' to optimize, the system iteratively generated, evaluated, and refined Python code to minimize historical forecasting error. 
When seeded with high-level descriptions of established epidemiological methods extracted from the literature, the search autonomously discovered and hybridized modeling strategies, ultimately generating multiple models that performed on par with or better than the official COVIDHub ensemble in retrospective analyses\cite{aygun_ai_2025}.
However, rigorous evaluation in time-series forecasting settings such as this requires prospective, out-of-sample forecasting. Retrospective studies inherently risk data leakage and hindsight bias regarding a pathogen's overall trajectory, and often fail to capture the chaotic realities of live forecasting, such as data reporting delays, mid-season shifts in disease dynamics, holiday anomalies, and subsequent data revisions, sometimes called `backfill'\cite{cramer_evaluation_2022,sherratt_predictive_2023,mathis_evaluation_2024,reich_collaborative_2019}.   

For that reason, in this paper we report a fully \textit{prospective} analysis of \ourmethod-generated models during the 2025-26 forecasting season across all three CDC Forecast Hubs---FluSight\cite{flusight}, COVIDHub\cite{cramer2022covidhub}, and RSVHub\cite{rsvhub}. We executed \ourmethod tree searches seeded with 90 unique candidate model prompts, producing over 207,500 individual candidate models. From these, we selected 54 models (by validation score and method diversity) and published them in the public GoogleHub\footnote{\href{https://github.com/google-research/google-research/tree/master/epi_forecasts}{github.com/google-research/google-research/tree/master/epi\_forecasts}} on the Google Research GitHub repository. A subset of 19 models were further designated to serve as components of a Google Research ensemble, yielding one forecast per pathogen, each submitted weekly in real time to the relevant CDC hub with auditable time stamps (\supfig{fig:selection_diagram}). Human involvement in this process comprised setup and supervision of \ourmethod runs (S.M., M.P., Z.S.), broader guidance on the use of the system (M.P.B., B.P.W.), and epidemiological modeling expertise (N.G.R.).

Additionally, we adapted this forecasting setup for two exploratory \textit{retrospective} experiments: (1) analyzing how the choice of search optimization metric influences forecast behavior, and (2) exploring different configurations of \ourmethod, including different Gemini tiers and an in-search `judge,' which can provide necessary guidance for the single-objective \ourmethod search when strict methodological constraints are desired.

This paper demonstrates that \ourmethod can generate, adapt, and replicate epidemiological forecasting models across a range of methodologies that, when ensembled, outperform current state-of-the-art methods and the gold-standard CDC hub ensembles—with the resulting Google-SAI ensemble officially recognized by the CDC as the top-performing model of the 2025–26 season\cite{cdc2026flusight}.
This automated development strategy demonstrates the advantages of a wider and deeper iterative exploration of modeling approaches and problem design than current standard practice of manual, bespoke model development allows, while also highlighting the value of human expertise in guiding model selection and development.

%% file: Sections/2-results.tex

We first outline our prospective results, showing model performance on each of the CDC's forecasting competitions during the 2025–26 forecasting season, as well as model performance of the individual models submitted to GoogleHub. 
We then present results from two retrospective influenza forecasting ablation experiments: (1) changing the optimization metric used for evaluation during \ourmethod model search; and (2) examining instruction-following capability of \ourmethod under different configurations.

\subsection*{Prospective Model Performance} \label{sec:part1}

\subsubsection*{Task, evaluation, and model selection.}

\begin{figure}[htbp]
    \centering
    \includegraphics[width=\textwidth]{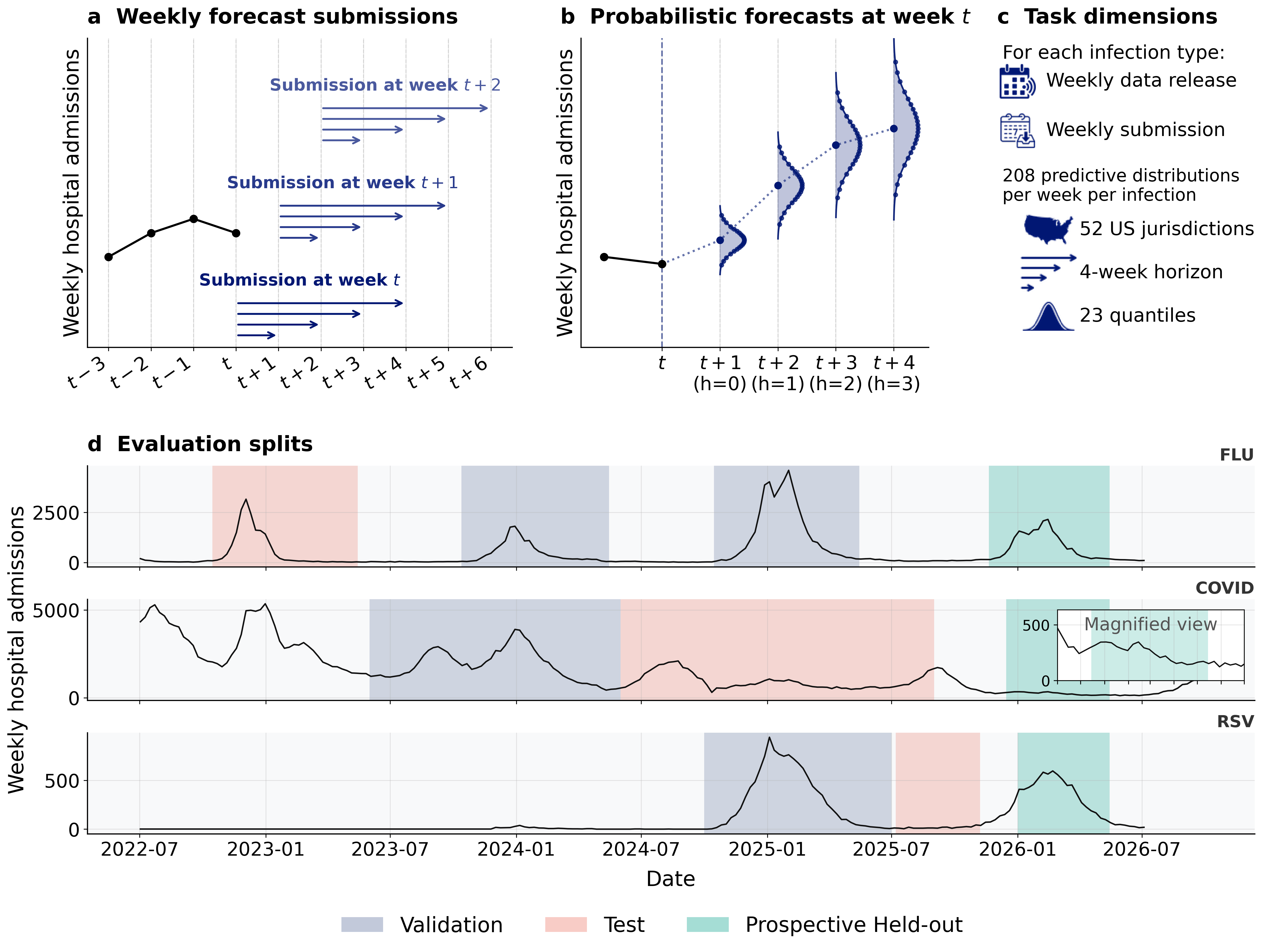}
    \caption{\textbf{Overview of the epidemiological forecasting task and evaluation framework.} \textbf{a} Rolling submission scheme. Historical surveillance data (black line) are available up to the forecast origin (vertical dashed lines). At each weekly submission point, forecasts are generated for four subsequent target weeks (horizontal arrows). \textbf{b} Depiction of a probabilistic forecast for one jurisdiction in a submission week. Taking the end of the observed data as the origin, the model produces a full predictive distribution via 23 quantiles for each of the four target horizons. The dotted line indicates the median trajectory. \textbf{c} Task dimensions and scale for each pathogen: 52 US jurisdictions, 4-week horizon, 23-quantile distributions, with weekly data updates and forecast submissions. \textbf{d} Sample time series of weekly hospital admissions for influenza (top), COVID-19 (middle), and RSV (bottom) for California, with non-overlapping temporal splits for validation (purple), retrospective testing (orange), and prospective evaluation (green). A non-chronological ordering was used for influenza forecast validation because the most recent season before the held-out set was deemed to be an overly challenging test set due to it being a unusually high influenza season (see Methods).}
    \label{fig:fig1}
\end{figure}

Each week, updated surveillance data on respiratory diseases in the US are released and participating teams submit standardized probabilistic forecasts to the CDC's FluSight\cite{flusight}, COVIDHub\cite{cramer2022covidhub} and RSVHub\cite{rsvhub}.
The main target for each prediction challenge is incident hospital admissions across 53 jurisdictions (50 states, Washington DC, Puerto Rico, and the entire US)\footnote{While the CDC hubs accept optional submissions for the entire US, our pipeline generates and evaluates forecasts for the 52 state and territory jurisdictions, excluding the national-level aggregate (see Methods for details).} at horizons of zero to three weeks ahead, with each forecast represented as a predictive distribution over 23 quantiles (Fig.~\ref{fig:fig1}a--c). Forecasts are evaluated against observations that are released (and occasionally revised) in future weeks, and are considered final after the end of the respiratory forecasting season. This setting is a leak-proof prospective test, as no model can access the outcomes it is predicting, and all submissions carry a publicly auditable time stamp on GitHub.

\begin{figure}[htbp]
    \centering
    \includegraphics[width=\textwidth]{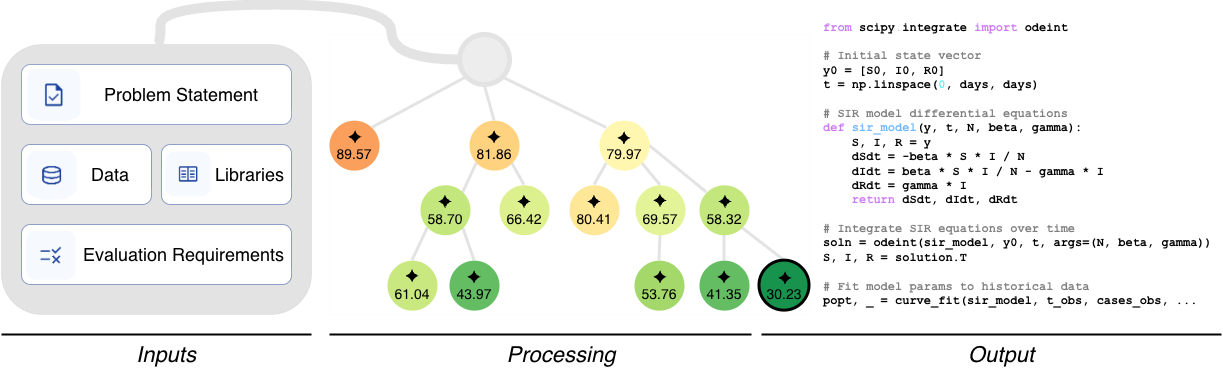}
    \caption{\textbf{The \ourmethod System Architecture.}
The \ourmethod workflow automates the translation of natural language scientific model statements into optimized executable code. \textit{Inputs:} \ourmethod ingests a natural language problem statement, datasets, and an evaluation harness. \textit{Processing:} An LLM-driven tree search explores a high-dimensional solution space, generating and refining Python implementations of epidemiological models. \textit{ Output:} \ourmethod returns the entire search tree, including Python code and a corresponding evaluation score for each node. \ourmethod identifies the best-scoring solution—a mathematically optimized model that minimizes the defined loss function while adhering to architectural constraints.}
    \label{fig:fig2}
\end{figure}

We treated this forecasting task as an automated model discovery problem, deploying \ourmethod to generate, evaluate, and iteratively refine code for candidate forecasting models. Given a natural language task specification, access to historical surveillance data, an evaluation harness with a defined objective function, and a computational sandbox with access to relevant scientific libraries, ERA uses an LLM-guided tree search to explore the space of possible modeling approaches autonomously (Fig.~\ref{fig:fig2}). At each node of the search tree, the LLM reads and modifies the previous model code, the modified model is executed and scored on a validation period, and the score is used to guide subsequent search branching decisions. To guard against overfitting to the validation period, all candidate nodes are additionally evaluated on a separate retrospective test set (Fig.~\ref{fig:fig1}d, \suptab{tab:dates}), and final models selected on the basis of combined performance across both periods (see Methods). Models were deployed prospectively starting in November 2025; real-time forecasts were submitted to the CDC hubs and pushed weekly to the GoogleHub repository for transparent evaluation over the season.

Models were scored during search using the Weighted Interval Score (WIS; equation~\eqref{eq:wis} in Methods), a proper scoring rule well-matched to the 23-quantile submission format, and one that approximates the Continuous Ranked Probability Score (CRPS), a well-known probabilistic evaluation metric \cite{Bracher2021}. All leaderboard comparisons reported in this paper use WIS applied to log-transformed targets and quantiles (referred to throughout as `log WIS'), a choice recommended for epidemiological forecast evaluation\cite{bosse_scoring_2023} and consistent with the official CDC hub evaluation framework.

To promote architectural diversity, we supplied ERA with a structured set of diverse prompts containing instructions drawn from the metadata of other CDC-submitting models, relevant epidemiological modeling literature, and suggestions from Gemini Deep Research (Methods). 

\begin{figure}[htbp]
    \centering
    \begin{minipage}[t]{0.35\textwidth}
        \raggedright
        \textbf{\textsf{a~~Example Real-time Forecasts for California}}\par
        \vspace{0.2cm}
        \includegraphics[width=\linewidth]{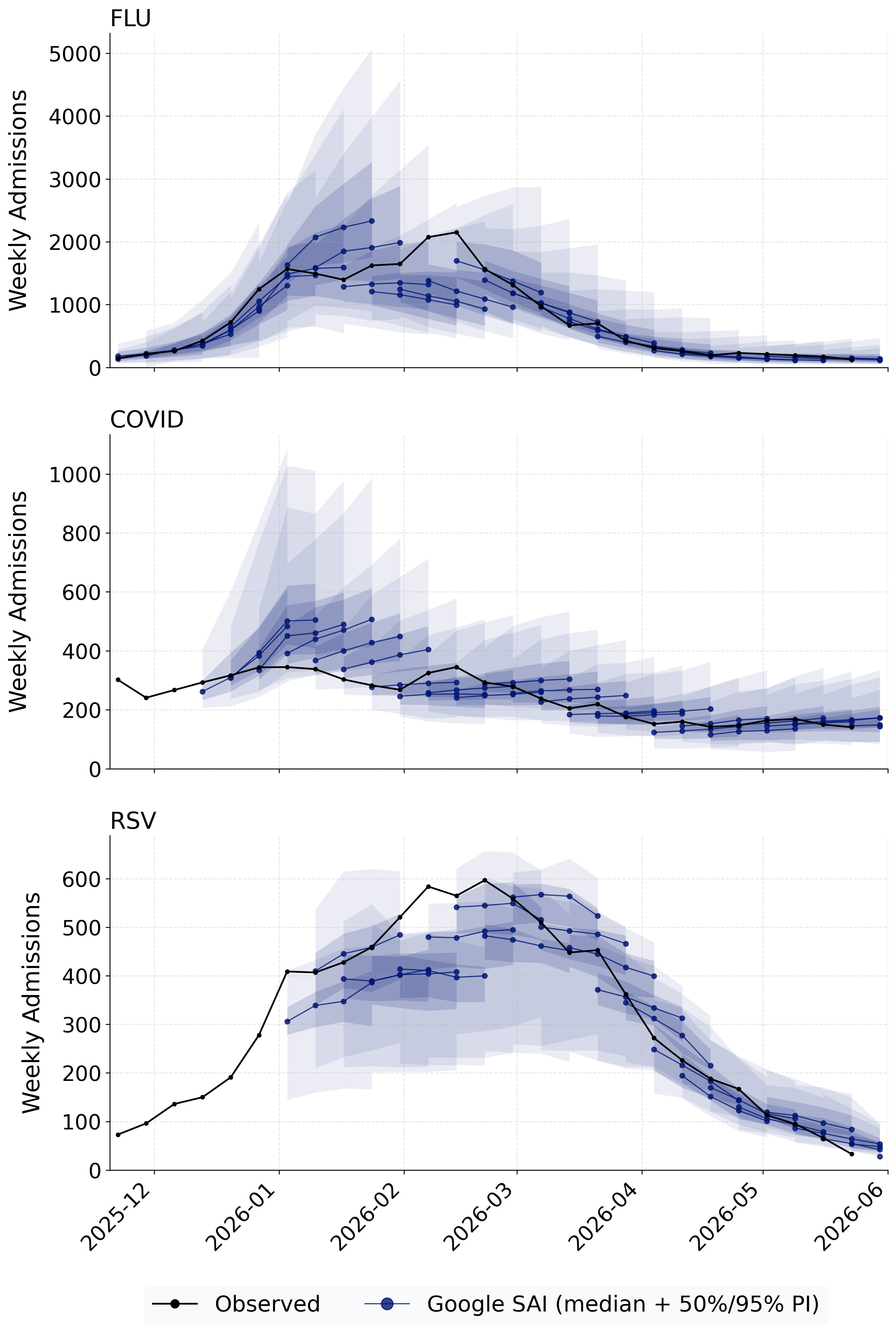}
    \end{minipage}%
    \hfill
    \begin{minipage}[t]{0.58\textwidth}
        \raggedright
        \textbf{\textsf{b~~Prospective Model Performance on CDC Forecasting Hubs for influenza, COVID-19 and RSV}}\par
        \vspace{0.2cm} 
        \includegraphics[width=\linewidth]{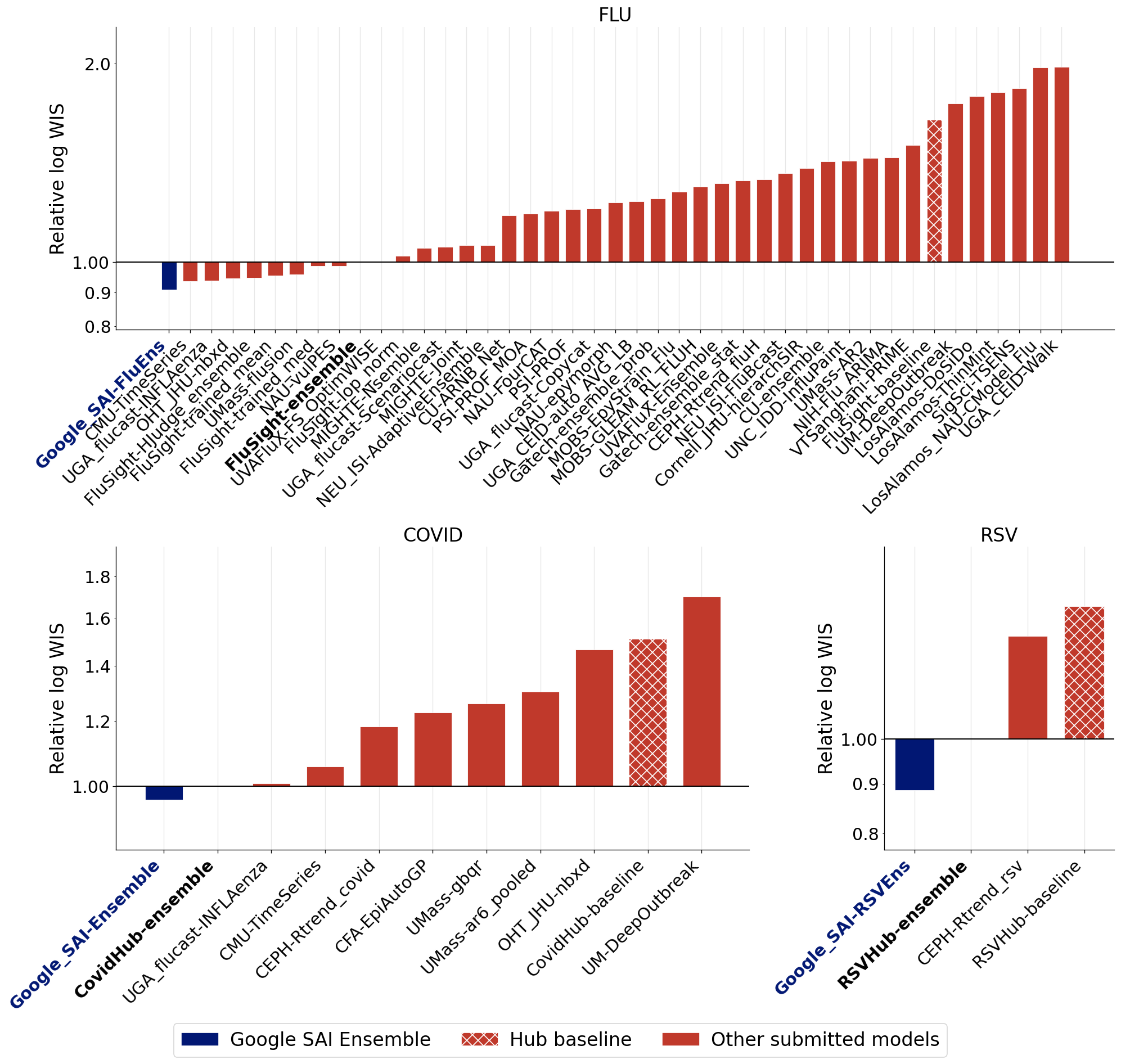}
    \end{minipage}
    
    \caption{\textbf{Prospective Performance of \ourmethod-Generated Ensembles.}
    \textbf{a} Weekly observed hospital admissions (black) and \texttt{Google-SAI} ensemble forecast 
    submissions for California, with median (dark blue) 50\%, and 90\% prediction intervals (shaded 
    regions), for influenza, COVID-19, and RSV. See~\supfigs{fig:flu_forecasts}, \ref{fig:covid_forecasts}, \ref{fig:rsv_forecasts} for forecasts for each jurisdiction.
    \textbf{b} Pairwise relative log WIS ranking of the \texttt{Google-SAI} ensembles (dark blue) among eligible CDC hub submissions (lower is better). For each pair of models $(i, j)$, the ratio of their mean log WIS scores is computed over tasks both models submitted; the relative score for model $i$ is the geometric mean of these pairwise ratios across all other models, rescaled so the CDC hub ensemble equals one (horizontal line; see Methods for additional details). Values $<1$ indicate better performance than the CDC ensemble. Models were included in this comparison if they submitted scorable predictions for at least $80\%$ of covered reference dates (from 22 November 2025 for influenza, 13 December 2025 for COVID-19 and 3 January 2026 for RSV to 27 May 2026), horizons, and jurisdictions. The \texttt{Google-SAI} ensembles achieved the best performance across all three pathogens, according to the pairwise relative log WIS metric.} 
    \label{fig:fig3}
\end{figure}

Around the beginning of the prospective submissions for each pathogen, we selected 34 influenza models, 18 COVID-19 models, and two RSV models from the full pool of generated candidates on the basis of retrospective validation and test performance (see Equation~\eqref{eq:selection_score} and Supplementary Section~\ref{sec:model_inventories} for test and validation scores for all models), with an additional emphasis on maintaining methodological diversity across the selected set. The significantly smaller RSV model pool reflects the pathogen's recent addition to forecast hubs and the limited number of available seasons of historical target data. We submitted predictions from these models weekly with time stamps to GoogleHub. From these, we selected a subset of models (seven for influenza, ten for COVID-19 and two for RSV), for inclusion in the three \texttt{Google-SAI} submissions--the ensemble models submitted by Google Research’s Science AI team to the CDC forecasting hubs. Ensembles are calculated by a direct quantile average of component forecasts (Methods). Figure~\ref{fig:fig3}a shows example forecasts for California for each of the three \texttt{Google-SAI} ensemble models. 

\subsubsection*{Ensembles of ERA-generated models perform competitively across pathogens.}

\texttt{Google-SAI} ensembles achieved top-tier performance across all three pathogens for the 2025-26 forecasting season. Figure~\ref{fig:fig3}b shows the \texttt{Google-SAI} ensembles achieving the best pairwise relative log WIS across all CDC hub-submitted models. \suptabs{tab:flu_model_summary},~\ref{tab:covid_model_summary},~\ref{tab:rsv_model_summary} provide a complete summary of performance of all models analyzed in this paper, including absolute and pairwise relative WIS and log WIS, mean absolute error (MAE), and coverage of 95\% prediction intervals; the \texttt{Google-SAI} ensembles display competitive performance across all four metrics and has the best on pairwise relative log-scale WIS, the official metric used to rank models by CDC\cite{cdc2026flusight}.
We start our analysis period from the first valid submission of the \texttt{Google-SAI} ensemble to each CDC hub (see Supplementary Sections~\ref{SIsec:flu}, \ref{SIsec:covid}, \ref{SIsec:rsv}). The task space for all analyses covers 52 jurisdictions (50 states, plus Washington DC and Puerto Rico) and four horizons (Fig.~\ref{fig:fig1}c). 
For influenza, we evaluate models on a task space including 28 reference dates. Eligible models were those submitting predictions for $\geq 80\%$ of tasks ($\geq 4,659$ tasks).
\gflumodel~ranked first among 43 eligible submissions (out of 57 total models submitting at least one prediction to FluSight) by pairwise relative season-average mean log WIS, with a score of 0.91 vs. 1.0 for the official \texttt{FluSight-ensemble}, which is created by taking the median across all eligible team submissions to the hub. \gflumodel also ranks first in the CDC's official FluSight evaluation of the 2025-26 season\cite{cdc2026flusight}.
For COVID-19, we consider 25 reference dates with the same task space coverage as for influenza, resulting in a minimum required task space of $4,160$ tasks. \gcovidmodel~ranked first across 11 eligible COVIDHub submissions (out of 17 total), also outperforming \texttt{CovidHub-ensemble} with a relative log WIS of 0.96.
RSV models are evaluated over 23 reference dates, with the same eligibility criteria, i.e., models must have scorable predictions for $\geq3,660$ tasks. \grsvmodel~ranked first among the four eligible RSVHub submissions (out of nine total), outperforming \texttt{RSVHub-ensemble} with a relative log WIS of 0.89.

Jurisdiction-level forecast plots for all three pathogens 
(\supfigs{fig:flu_forecasts}, \ref{fig:covid_forecasts}, \ref{fig:rsv_forecasts})
illustrate the behavior of these ensemble forecasts and of the ground truth observations across the analyzed reference dates.
Nearly all locations saw a clear `seasonal peak' for influenza during the forecasting period. For COVID and RSV, the magnitude of a seasonal rise and fall varied by location.
For influenza, the ensemble median roughly tracked the observed seasonal trajectory, although it tended to under-predict as the season initially progressed and over-predict after the season peak. Predictive intervals were wider during the rapid epidemic ascent in December 2025 and tightened through the peak and decline phases from January 2026 onward. 
COVID-19 forecasts were generated over a more gradual seasonal peak in most locations. 
RSV exhibited the most heterogeneous state-level dynamics, with greater variation in the timing and shape of epidemic curves, and correspondingly wider predictive intervals throughout the season.
Overall, wider prediction intervals indicate greater uncertainty about future epidemic trajectories, while narrower intervals indicate greater confidence in the range of likely outcomes.
Additionally, \supfigs{fig:flu_performace_by_location}, \ref{fig:covid_performace_by_location}, and \ref{fig:rsv_performace_by_location} show each of the \texttt{Google-SAI} models achieving clearly better-than-average performance across all jurisdictions as compared to other models submitting to the CDC hubs, with the exception only of Hawaii for COVID-19. 

This pattern of competitive consistency is preserved across all prediction horizons from zero to three weeks ahead (\suptabs{tab:flu_logwis_by_horizon}, \ref{tab:covid_logwis_by_horizon}, and \ref{tab:rsv_logwis_by_horizon}). 
For all three pathogens the \texttt{Google-SAI} ensembles track within the top cluster of submitted models by log WIS at every horizon, well below the CDC baselines, and maintain their relative positions without degradation at longer lead times.

WIS is the sum of three components describing forecast behavior: dispersion (width of the predicted distribution), overprediction, and underprediction \cite{bosse2022evaluating}. We plot the decomposed season average WIS for the top-performing eligible models submitting to each CDC forecasting hub in \supfigs{fig:flu_wis_decomp}, \ref{fig:covid_wis_decomp}, \ref{fig:rsv_wis_decomp}. 
Across pathogens, \texttt{Google-SAI} models exhibited a consistent asymmetry in their WIS decomposition, with underprediction contributing more to mean WIS than overprediction. This asymmetry was relatively weak for COVID-19, more apparent for influenza, and strongest for RSV, where underprediction accounted for nearly 60\% of the total score. 
This trend of insufficient coverage of unexpectedly high incidence appears across models for all three pathogens, potentially indicating a fundamental challenge in the prediction of the onset and exponential growth of outbreaks.

Standardized ranks provide an additional task-level view of relative performance, showing whether a model is consistently competitive across forecasts or performs well on average despite occasional poor rankings. For each task, a standardized rank of one indicates the best log WIS among submitting models, while a rank of zero indicates the worst (see Methods for details). The standardized log WIS rank distribution of FluSight-submitting models (\supfig{fig:flu_rank_distribution}) shows substantial variability in individual model performance. Among eligible models, excluding \texttt{FluSight-ensemble}, \gflumodel~had the highest first quartile rank, meaning it ranked near the bottom less often than any other individual model. Some models achieved high ranks more frequently, but also had more frequent low-ranking forecasts. Consistent with previous findings\cite{cramer_evaluation_2022}, \texttt{FluSight-ensemble} had a narrower standardized rank distribution than most individual models, indicating that it was rarely either the best or worst performer. This is expected for an ensemble that combines multiple component models, as ensembles tend to be more robust across forecasting conditions rather than dominant on every individual task. While \gflumodel~is also an 
ensemble, its component models are fewer in number (seven versus 43), which limits the diversity-driven stabilization that characterizes hub ensembles with many independent contributors. The standardized rank distributions for the \texttt{Google-SAI} ensembles for COVID-19 (\supfig{fig:covid_rank_distribution}) and RSV (\supfig{fig:rsv_rank_distribution}) show similar patterns. \gcovidmodel~ranked second among COVIDHub submissions, and \grsvmodel~ranked first among RSVHub submissions with the smallest first quartile of standardized ranks.

\subsubsection*{Discovery of diverse forecasting architectures.}

\begin{figure}[htbp]
    \centering
    \includegraphics[width=\textwidth]{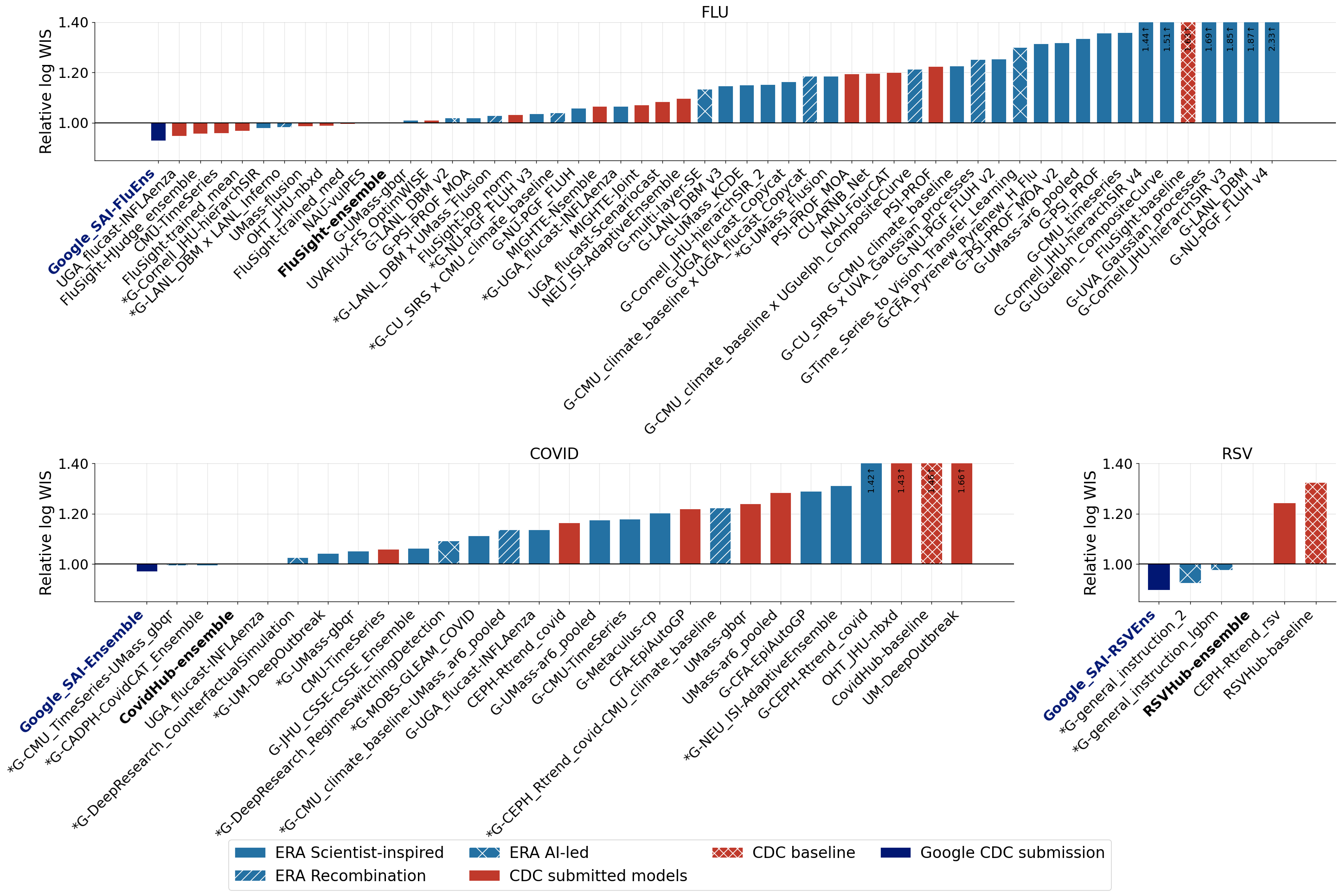}
    \caption{\textbf{Individual \ourmethod-generated Model Performance.}
    The \ourmethod pipeline generated a broad and diverse pool of models during the prospective season. 
    Each bar represents the season-average pairwise relative log WIS of an individual model. For each pair of models $(i, j)$, the ratio of their mean log WIS scores is computed over tasks both models submitted; the relative score for model $i$ is then the geometric mean of these pairwise ratios across all other models, rescaled so the CDC hub ensemble equals one (horizontal line). Values $<1$ indicate better performance than the CDC ensemble. 
    CDC-submitted models are restricted to the top 20 best-performing eligible models ($\geq 80 \%$ task coverage), plus the CDC hub ensembles, baselines, and \texttt{Google-SAI} ensemble submission (if they are not already selected). 
    The \ourmethod-generated component models (light blue) are hatched by method origin: `scientist-inspired' methods from hubs and relevant publications (solid blue), method `recombinations' (diagonal-hatched blue), and `AI-led' methods (cross-hatched blue). Additionally, CDC-submitted models (solid red) and CDC hub ensemble and baseline (diagonal-hatched red) are included. 
    Asterisks denote components of the \texttt{Google-SAI} ensemble submission (dark blue). Note that relative log WIS scores may be different than in other figures (e.g., in Figure \ref{fig:fig3}) because they are computed on different model sets.
    \textbf{Top} Influenza: 34 models generated, with two individual \ourmethod-generated models and \gflumodel~outperforming \texttt{FluSight-ensemble}. \textbf{Bottom left} COVID-19: 17 models generated with \gcovidmodel~outperforming \texttt{CovidHub-ensemble}. \textbf{Bottom right} RSV: two novel models generated, both outperforming \texttt{RSVHub-ensemble} in the hub's inaugural season.}
    \label{fig:fig4}
\end{figure}

All \texttt{Google-SAI} component models, along with additional candidate models, were submitted as individual forecasts to GoogleHub throughout the prospective season. 
These models were generated using a structured set of prompts containing summarized method instructions from three broad categories: (1) scientist-inspired methods adapted from existing literature and documentation on human-developed models, (2) recombinations that combine pairs of distinct methodological frameworks, and (3) AI-led methods, where an LLM suggests the modeling approach rather than following a human-specified methodology (see Methods for additional detail).
Figure~\ref{fig:fig4} shows the relative log WIS of every model in this repository against the respective CDC hub ensemble (values below one indicate better-than-CDC-ensemble performance), alongside several top-performing independently submitted models from each hub for context.

Across all three hubs, \texttt{Google-SAI} ensemble forecasts achieved the best relative log WIS scores across all internal component models and all top-performing hub-submitting models (Fig.~\ref{fig:fig4}). For influenza, two \ourmethod-generated component models from our prospective time-stamped repository additionally outperformed \texttt{FluSight-ensemble}. Of these, one is an adaptation of an existing `scientist-inspired' method (\texttt{G-Cornell\_JHU-hierarchSIR}), and the other a `recombination', a hybrid of \texttt{LANL-DBM and LANL-Inferno}. Both were included as components in \gflumodel. 
For COVID-19, two individual component models also outperformed \texttt{CovidHub-ensemble}: a scientist-inspired adaptation (\texttt{G-CADPH-CovidCAT\_Ensemble}) and a recombination (\texttt{CMU\_TimeSeries and  UMASS\_gbqr}). Similarly, for RSV, both individual AI-led models out-performed \texttt{RSVHub-Ensemble}.

Pairwise comparisons between \ourmethod-generated adaptations and their human-authored reference models reveal that \ourmethod-generated models can either ameliorate or degrade performance as compared to their original scientist-created counterparts. 
No single prompting strategy reliably yields only good models (Fig.~\ref{fig:fig4}). Prompts grounded in established domain literature (scientist-inspired) and those `recombining' different paradigms can provide models that forecast competitively in mature contexts such as influenza and COVID-19, but equally produce poorer-performing models. 
Methodological diversity is a well-established driver of ensemble forecast skill~\cite{cramer_evaluation_2022}. Our results suggest this principle extends to automated model generation: aggregating models across all three prompt strategies yields ensemble predictions that consistently outperform both individual components and CDC hub ensembles.

However, the above comment on diversity applies to the methodologies \textit{requested} in the scientist-inspired prompts, rather than necessarily those implemented by the generated models. 
To assess methodological fidelity, a subject-matter expert reviewed the 24 scientist-inspired \ourmethod-generated influenza models submitted to GoogleHub in the prospective study and classified each according to how faithfully it implemented the requested methodology (\suptab{tab:flu_model_summary}; see Methods). 
Of the 24 models, 17 successfully followed the instructions, four implemented the requested methodology only partially, and three failed to implement it. 
Partial or failed implementations were concentrated among prompts specifying more complex mechanistic or statistical approaches, with omitted or approximated components including Kalman filtering and explicit compartmental models, and some models instead implementing simpler ML approaches such as gradient boosting. 
We investigate this limitation further in a subsequent exploratory analysis. 
Importantly, the expert assessments for the component models selected for \gflumodel~(\suptab{tab:flu_model_summary}) confirm that, despite occasional instruction-following failures, the final ensemble retains methodological diversity across mechanistic, statistical, and machine learning paradigms.

\subsection*{Retrospective Experiments}

\subsubsection*{Internal optimization metrics shape forecast behavior.}

Beyond generating performant, production-ready models to predict outbreaks, the \ourmethod pipeline provides a systematic experimental mechanism to explore how problem framing impacts forecast behavior. 
To explore the effects of \ourmethod optimization metric on model output, we evaluated three candidate metrics drawn from the two principal families of proper scoring rules for probabilistic forecasting~\cite{Gneiting2007}. 
The first is CRPS, which measures the integrated squared difference between the predictive cumulative distribution function (CDF) and the empirical CDF of the observation. The WIS used in analyses in the previous section is a quantile-based approximation of CRPS~\cite{Bracher2021}. 
The second is the Logarithmic Score (Log Score), a classical proper scoring rule based on the log-probability assigned to the observed outcome; to frame this as an optimization loss, we evaluate the negative log-probability, where a lower score indicates better performance.
The third is Log-scale CRPS, a CRPS variant in which both predictions and observations are log-transformed prior to scoring (as for log WIS), combining the distributional breadth of CRPS with greater sensitivity to relative errors at low counts. See Methods (Equations \eqref{eq:crps}--\eqref{eq:log_score}) for more detailed explanations of these scoring functions.

We used \ourmethod to search for influenza models with the same data splits as described in Fig.~\ref{fig:fig1}, but varied the optimization metric used for in-search evaluation. For each metric, we ran five independent replicas of 2,500-node searches. 
Supplementary Figure~\ref{fig:metrics_in_search} shows how the search evolved under each metric by tracking the cumulative best score---the in-search validation score of the best model found so far at each point in the search. These curves reveal differences across metrics in both search efficiency and out-of-sample generalization.
The continuous gradients provided by distance-based metrics (CRPS and Log CRPS) allow the search to identify performant architectures rapidly, with cumulative best test scores generally reaching their minimum between 500 and 750 nodes (\supfig{fig:metrics_in_search}, left and middle columns).
However, searches under these metrics are also clearly susceptible to overfitting, visible in later stages of the search as rising test curves (orange) concurrent with falling validation curves (purple).
Because of this late-stage overfitting, the final validation-selected models optimized for one metric do not always achieve the best test performance on that metric: notably, optimizing for standard CRPS yields better late-search test performance on Log CRPS (panel d) than optimizing for Log CRPS directly (panel e).
Because WIS and log WIS approximate CRPS and Log CRPS, respectively, this result partially explains our strong log WIS performance in the previous section, even though those searches used WIS as the optimization metric.
This late-stage overfitting also reinforces the need to select best \ourmethod nodes by selection score (see Methods, Equation~\eqref{eq:selection_score}), which includes components for both validation and test scores (with a larger weighting on test score), rather than directly by validation score.

\begin{figure}[htbp]
    \centering
    \includegraphics[width=.75\textwidth]{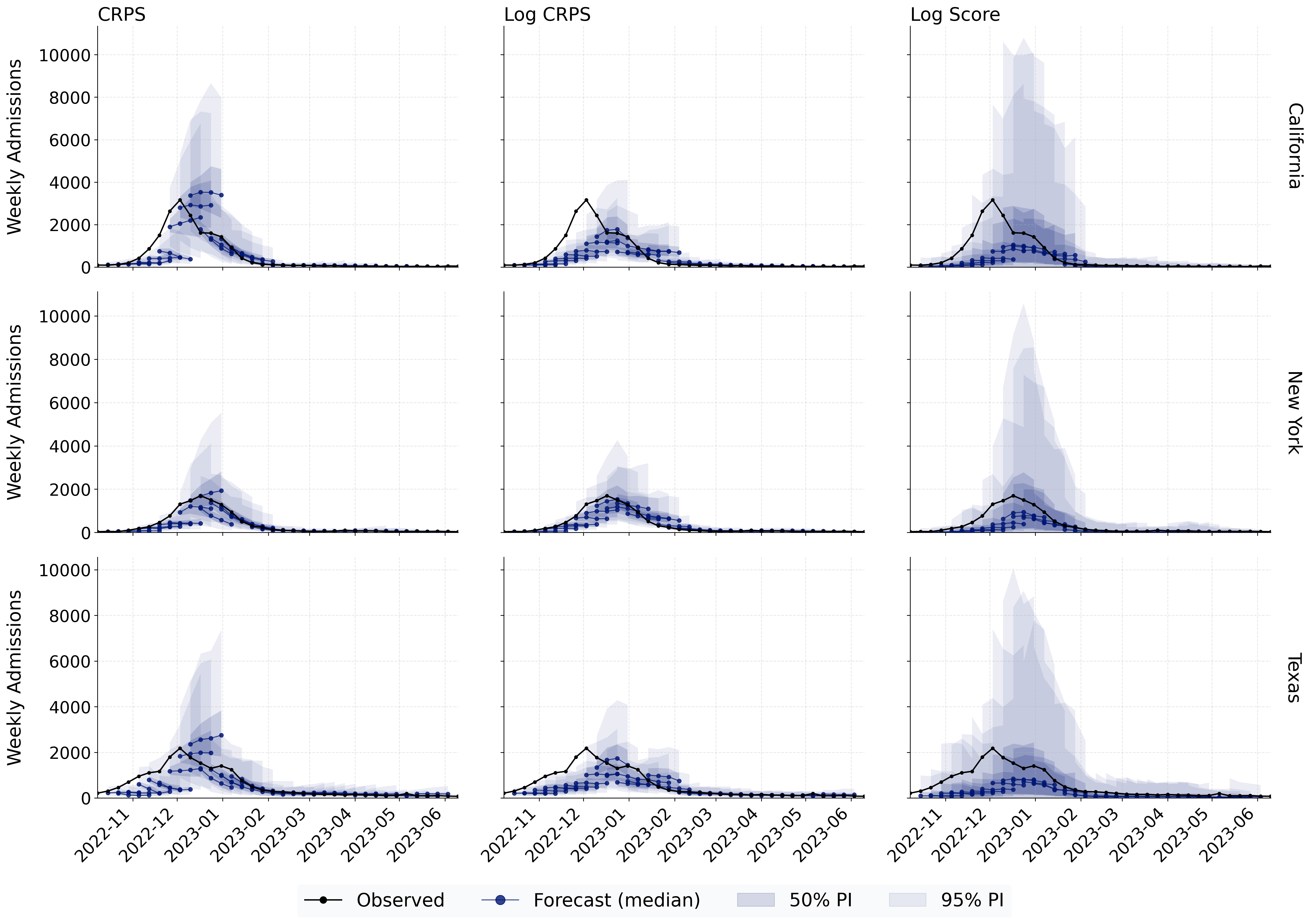}
    \caption{\textbf{Forecast behavior differs by \ourmethod optimization metric.}
Test-period forecasts of weekly influenza hospital admissions in three selected states (rows: California, New York, and Texas). Forecasts are from the best node (by selection score, Equation~\eqref{eq:selection_score}) of the first of five replicas of \ourmethod searches optimized for CRPS (left), Log CRPS (middle), or Log Score (right).
Black lines and points show observed admissions.
Each connected blue line is the median forecast issued on a single reference date across its forecast horizons, and darker and lighter shading show the corresponding 50\% and 95\% prediction intervals (PIs); shading overlaps across consecutive forecasts.}
\label{fig:metric_forecasts}
    \label{fig:fig5}
\end{figure}

We therefore selected a single best node from each replica using this score and evaluated its test-period forecasts across all three metrics.
Fig.~\ref{fig:metric_forecasts} illustrates the resulting test-period forecasts from the best node of one replica for each metric in three selected states\footnote{The selected states are broadly representative of overall observed trends}.
Although all three models slightly lag the true epidemic curve, underpredicting during the growth phase and overpredicting during the decline, the size of these errors and the shape of the predictive distributions differ qualitatively by metric.
The CRPS-optimized model tracks the observed curve most closely---its medians come nearest to the observed peaks, particularly in New York, though it briefly overshoots immediately after the peak in California and Texas---and its prediction intervals contract tightly once admissions return to low levels.
In contrast, the Log CRPS- and Log Score-optimized models substantially underpredict the peak in all three states.
The Log Score-optimized model is further distinguished by intervals that remain wider relative to observed admissions throughout the low-incidence end of the season (March--June) in all three states, whereas CRPS intervals (and Log CRPS intervals in California and New York) collapse onto the observations.
These differences are consistent with how the metrics weight errors: standard CRPS penalizes absolute errors and therefore rewards accurately capturing the high-count surge, whereas Log CRPS penalizes multiplicative (relative) errors, so underpredicting the peak by a given factor costs roughly the same as a comparable relative error at low counts.
Finally, the Log Score has a very harsh penalty ($-\infty$) when a model assigns no probability to the eventual outcome, so this incentivizes a wide spreading out of probability across a large range of possible outcomes.

\suptab{tab:forecast_metrics_extended_si} summarizes test-period forecast quality and behavior across all replicas and jurisdictions. Because post-search selection via Equation~\eqref{eq:selection_score} avoids the late-stage overfitting seen in \supfig{fig:metrics_in_search}, the selected models for each optimization metric achieve, on average, the best (lowest) test score on that metric. We additionally report mean bias (the average difference between predicted median and observed admissions), the proportions of median under- and overprediction, and the mean width of the 50\% prediction interval.
The two distance-based metrics (CRPS and Log CRPS) yield similar aggregate bias and interval widths. Standard CRPS achieves the smallest mean bias, most balanced under- and overprediction, and narrowest 50\% prediction intervals (bolded values). Models optimized for Log Score diverge most sharply, exhibiting the strongest underprediction bias, widest prediction intervals, and highest variance across replicas. This is because Log Score assigns an effectively infinite penalty when the observed value falls outside the predicted distribution, which favors dispersed forecasts.

Together, these results show that the choice of optimization metric in \ourmethod measurably shapes the behavior of final forecasts, and that \ourmethod makes it practical to systematically experiment with objective functions and problem design at scale. In our exploratory analysis, standard CRPS provided the most stable search dynamics and the best-performing forecasts overall.

\subsubsection*{Instruction-following fidelity varies across methods and search configurations.}

\noindent In the prospective evaluation, most \ourmethod-generated models implemented the requested methodology successfully, but unsuccessful or partially successful implementations tended to substitute standard ML regressors for more specialized statistical or mechanistic approaches. This limitation is important because the strength of collaborative forecasting ensembles depends partly on methodological diversity across mechanistic, time-series, statistical, and ML models~\cite{cramer_evaluation_2022, mcgowan_collaborative_2019}. 
Reproducing that diversity with an automated system therefore requires not only instruction guidance but also a mechanism to enforce method compliance.
We therefore investigated how instruction-following fidelity varied with the requested methodology and with underlying LLM, and whether fidelity could be improved by evaluating methodological compliance during the search.

We compared four \ourmethod{} configurations: Gemini 2.5 Flash with an in-loop judge, and Gemini 2.5 Flash, Gemini 3 Flash, and Gemini 3 Pro without a judge. We applied each configuration to four established FluSight-submitting methods representing distinct modeling strategies: \texttt{UMass-gbqr}\cite{ray2024flusion}, \texttt{UGA\_flucast-INFLAenza}\cite{fox_uga_flucast_inflaenza}, \texttt{Cornell\_JHU-hierarchSIR}\cite{alleman2026data}, and \texttt{NU-PGF\_FLUH}\cite{stonge_pgf_fluh}. Each configuration–method combination was evaluated in three independent runs. An external post-hoc judge, using method-specific rubrics calibrated against expert epidemiologist assessments, classified each final implementation as a `Match', `Partial Match’, or `No Match’. Predictive performance was assessed using the selection score \eqref{eq:selection_score}, with lower values indicating better forecasts (Methods).

Instruction-following fidelity varied markedly across the requested methods (\suptab{tab:llm_summary_updated}). 
While ten of 12 \texttt{UMass-gbqr} runs achieved a full match, none of the 12 \texttt{NU-PGF\_FLUH} runs did, indicating that fidelity depended strongly on the methodology being reproduced. Higher-tier Gemini models yielded selected models with lower mean selection score for all four methods: averaging the method-level means equally, selection score decreased from 255.7 with Gemini 2.5 Flash to 164.0 with Gemini 3 Flash and 143.4 with Gemini 3 Pro. Improvements in predictive performance did not, however, consistently correspond to greater fidelity. For example, Gemini 3 Pro achieved full matches in all three \texttt{Cornell\_JHU-hierarchSIR} runs but only partial matches in all three \texttt{UGA\_flucast-INFLAenza} runs.
The effect of the in-loop judge was similarly method dependent. Adding an in-loop judge to Gemini 2.5 Flash increased the total number of full matches from four of 12 to seven of 12 and lowered mean selection score for three of the four methods. The clearest improvement occurred for \texttt{UGA\_flucast-INFLAenza}, for which full matches increased from one of three runs to three of three and mean selection score decreased from 316.16 to 196.51. Conversely, for \texttt{Cornell\_JHU-hierarchSIR}, the \ourmethod search with an in-loop judge produced one full match but increased mean selection score from 179.09 to 411.97. 

Together, these results show that higher-tier LLMs consistently improved predictive performance, whereas instruction fidelity and the benefits of in-loop judging varied across methodologies. Because we evaluated only one implementation of the in-loop judge, these results do not establish whether alternative judge designs or calibration strategies would produce more consistent improvements. 
Full run-level results are reported in \suptabs{tab:judge_comparison} for the in-loop judge and \ref{tab:llm_comparison} for Gemini tier.

%% file: Sections/3-discussion.tex

This work demonstrates that an LLM-guided tree search system can generate epidemiological forecasting models that compete with the state of the art across three respiratory pathogens in a fully prospective, leak-proof evaluation setting. The key finding is not that any single \ourmethod-generated model outperforms all alternatives, but rather that the system can rapidly produce a large, methodologically diverse pool of competitive models whose ensemble matches or exceeds the gold-standard CDC hub ensembles---themselves aggregations of forecasts from dozens of expert teams~\cite{mcgowan_collaborative_2019, reich_accuracy_2019, 
cramer_evaluation_2022}. Crucially, every model in this pool is a complete, executable Python program whose forecasting logic and timestamped outputs can be read, audited, and modified by any domain scientist. Model source code and weekly forecast submissions are publicly available under the Apache~2.0 open-source license\footnote{\url{https://github.com/google-research/google-research/tree/master/epi_forecasts}}---a property that distinguishes \ourmethod from black-box AutoML\cite{he2021automl} systems and aligns with the transparency increasingly emphasized in public health forecasting.

The practical implications of this automation extend beyond the specific pathogens studied here. Currently, collaborative forecasting hubs rely on sustained, labor-intensive participation of expert teams running and maintaining bespoke models week after week. This manual paradigm scales poorly to highly granular geographic resolutions (e.g., US counties), new geographic regions, or novel pathogens~\cite{viboud_rapidd_2018,kerr_software_2026}. By automating model ideation, implementation, and evaluation, systems like \ourmethod can democratize access to expert-level forecasting, providing a concrete blueprint for deploying ensemble-grade capacity at larger scales. Additionally, \ourmethod  could substantially lower the barrier to entry for hub participation, enabling rapid deployment of competitive forecasting capacity in settings that currently lack the specialized modeling workforce required. This potentially allows expanding the forecast `hub' concept\cite{kerr_software_2026} to jurisdictions and granularities where it is harder to attract expert attention.

Notably, the RSV results presented here provide a direct test of a `cold start' scenario. With minimal historical data, no mature modeling literature to draw from, and a hub in its first season, \ourmethod's best-performing models came from unconstrained architectural searches in which the system discovered effective approaches without domain-specific guidance. While the resulting models were necessarily simpler than those generated for influenza (both were gradient-boosted pipelines fed with cross-pathogen auxiliary inputs), their top-ranked performance among the other RSVHub submissions, ensemble and baseline models demonstrates that automated search can provide useful forecasting capacity even in data-scarce regimes. 

One limitation of the \ourmethod-generated forecasts presented and evaluated in this work is that, while they achieve high accuracy scores relative to existing models, they suffer from similar systematic biases observed in other models across multiple seasons of these forecasting efforts. 
Forecasts of seasonal influenza and COVID-19 have struggled to capture dynamics of rapid increase and decrease during seasonal epidemics: models tend to underestimate the rate of increase as the epidemic is rising and the rate of decrease as it falls\cite{lopez_challenges_2024,cramer_evaluation_2022,mathis_evaluation_2024}.
Forecasts generated by \ourmethod are no exception.
However, this reflects not so much a limitation of the \ourmethod model generation pipeline as it does a structural challenge with developing a set of scorable tasks for epidemic forecasting that incentivize forecasts which can be used for improving situational awareness during seasonal outbreaks or for decision-support.
Developing evaluation rubrics that are aligned with public health decision-making is an active area of research\cite{bilinski_adaptive_2023,gerding_evaluating_2025,mills_metric_2026}, and systems like \ourmethod highlight possibilities for training models that have been optimized for specific contexts.
Indeed, our retrospective comparison of optimization metrics serves as a first step in this direction. By demonstrating that the \ourmethod search is responsive to the mathematical properties of its reward signal, we highlight how forecasters can use this pipeline to rapidly discover novel architectures tailored to specific settings or requirements.

A second limitation concerns the ensemble aggregation strategy itself. In this study, the submitted ensemble forecast was computed as a simple, equally weighted median across component models—an approach that, while robust and well-established in forecasting practice~\cite{reich_accuracy_2019}, makes no attempt to learn from the relative strengths and weaknesses of its components. Yet the forecasting task is inherently high-dimensional: individual models may excel in different jurisdictions, at different forecast horizons, or under different epidemic conditions such as rapid growth versus post-peak decline. The optimal combination strategy is therefore not a single set of static weights but a context-dependent mapping—itself a challenging optimization problem~\cite{claeskens_forecast_2016,ray_comparing_2023}. Framing ensemble construction as a scorable task and applying \ourmethod's tree search to find effective adaptive weighting schemes or selection logic is a natural direction for future work.

A further limitation of the prospective evaluation is that not all models are assessed on identical sets of forecasting tasks. Models join hubs at different points in the season, occasionally skip submission weeks, or cover different subsets of jurisdictions, meaning that each pairwise comparison is computed over a different intersection of shared tasks.
While the pairwise relative scoring framework we adopt (following established practice~\cite{Bracher2021}) mitigates this by computing score ratios only over mutually completed tasks and then averaging across all opponents, it does not fully eliminate the issue: a model that begins submitting after a particularly difficult or easy phase of the epidemic is effectively evaluated on a different forecasting problem than one that was active throughout. We partially address this by imposing an 80\% task-coverage eligibility threshold, but acknowledge that strict like-for-like comparison across all models remains infeasible in any real-time collaborative forecasting setting.
The prospective evaluation is also restricted to subnational forecasts: we did not generate or submit forecasts for the national US target accepted by the CDC Hubs, and all evaluations therefore concern state- and territory-level jurisdictions.

While traditional algorithms are expected to be deterministic from end to end, \ourmethod automates the exploratory model search process typically performed by human scientists, a process that is itself inherently non-reproducible. Because \ourmethod operates as a stochastic search algorithm subject to multiple sources of randomness, independent searches seeded with the same prompt will naturally yield distinct outputs. This variability is compounded by the rapid development cycle of foundation models; for instance, the majority of the models composing the \texttt{Google-SAI} ensembles were generated using Gemini 2.5 Flash, a model scheduled for deprecation in October 2026\cite{google_gemini_2026}. However, this stochasticity in the discovery phase does not compromise the reproducibility of our results. The \ourmethod-produced model codes (and their corresponding forecasts) used in this study are entirely deterministic, auditable, and reproducible. Rather than guaranteeing identical code generation across different search runs, \ourmethod's contribution lies in discovering transparent, executable Python code that scientists can reliably run, inspect, and maintain to reproduce the forecasts and analysis reported here.

More broadly, the computational footprint of our prospective campaign---142 tree searches of up to 2,500 nodes each---should be viewed as a snapshot of foundation model capabilities in the Fall of 2025, when candidate code had to be frozen prior to the start of the forecasting season. 
\ourmethod's extensive tree search over thousands of nodes was originally designed to compensate for the higher coding and reasoning error rates of earlier-generation LLMs. As shown in our retrospective comparisons across Gemini tiers, newer frontier models generate performant, compliant forecasting code in substantially fewer attempts; consequently, research teams deploying agentic forecasting pipelines today would likely require far smaller search budgets than the 2,500-node limit used here.

The retrospective experiments presented in Results reveal the capabilities and structural limitations of the current system. The bias toward gradient-boosted implementations---observed even when the search was explicitly prompted with mechanistic or statistical instructions---likely reflects a fundamental asymmetry in the training distribution across the Google Gemini model tiers tested in this study: standard software patterns are more reliably generated than highly specialized scientific frameworks. 
We showed that two complementary interventions can partially mitigate this bias: using more capable foundation models (Gemini~3 Pro achieved full fidelity on some of the more complex methods where Gemini Flash models could not), and deploying an LLM judge-in-the-loop to gate non-compliant implementations. 
However, for the most specialized methods (e.g., \texttt{NU-PGF\_FLUH}) neither intervention was sufficient to achieve full methodological fidelity.
We emphasize that these instruction-following and in-loop judge experiments were designed as small-scale exploratory analyses rather than an exhaustive benchmark, and could not be expanded further on the prospective baseline model due to the scheduled deprecation of Gemini 2.5 Flash

Our results suggest that expert human judgment still plays a meaningful role in guiding model development, even in a highly automated forecasting pipeline like \ourmethod.
The Google team that drove this project included several scientists, analysts, and engineers (without explicit infectious disease or epidemiological training) and one expert consultant from academia who has participated in epidemiological forecasting challenges for over ten years.
While the tree search experiments were orchestrated, compiled and analyzed by the Google engineers, the expert provided input on \ourmethod prompt framing and which existing models to prioritize exploring, and gave judgments on how faithful the \ourmethod-generated re-implementations of models were. 
Additionally, the epidemiology expert identified and suggested two models that were highly performant on FluSight Forecast Hub several years ago but were no longer currently submitting: \texttt{LANL-DBM} and \texttt{LANL-Inferno}. 
\ourmethod implemented a recombination of these models, and this adaptation was one of the top-scoring models from GoogleHub (Figure~\ref{fig:fig4}).
Our \texttt{Google-SAI} ensembles were also generated using a direct average of \textit{human-selected} models from the \ourmethod-generated pool. Although we could have automated model selection using validation and retrospective test scores, this would not have considered model diversity.  
Therefore, we suggest there is some anecdotal evidence that even in the presence of a highly automated LLM-driven model-generation architecture, there remains a place for human expertise to highlight and drive opportunities for model exploration.


Our prospective results demonstrate that autonomous LLM-guided tree search can operate reliably under the rigors of real-time public health surveillance, producing transparent, auditable forecasting software that outperforms state-of-the-art human-curated ensembles across multiple pathogens. 
Rather than replacing domain scientists with black-box predictions, systems like \ourmethod shift the role of human expertise from labor-intensive software implementation to high-level scientific direction, including defining epidemiological hypotheses, curating data, and designing evaluation rubrics that align with public health needs. 
By dramatically lowering the engineering barrier required to create and maintain diverse model ensembles, autonomous code discovery offers a scalable blueprint for rapid outbreak response across new pathogens, granular geographies, and resource-limited public health settings.

%% file: Sections/4-methods.tex


\subsection{Data and Forecasting Task}
\label{sec:data_forecasting_task}

This study focuses on forecasting hospital admissions resulting from three major seasonal respiratory pathogens in the US: influenza, COVID-19, and RSV. Unlike retrospective `hindcasting,' this work implemented a real-time prospective study, requiring models to generate predictions before the ground-truth observations were recorded. This approach subjects the forecasting process to real-time surveillance conditions, including reporting delays and data revisions typical of live national healthcare systems.

The scale of the forecasting task is high-dimensional, requiring full geographic coverage of the US. Models provide predictions for 52 jurisdictions, including all 50 states, the District of Columbia, and Puerto Rico. Although the CDC hubs also accept optional submissions for the national-level US aggregate (a 53rd jurisdiction), we excluded the national target from our pipeline: national admission counts are an order of magnitude larger than individual state counts (which can disproportionately dominate training loss if pooled), and predictive quantiles cannot be directly summed across states without explicitly modeling cross-jurisdiction covariance. Because the CDC evaluates state-level performance separately from national-level forecasts using pairwise comparisons over shared tasks, omitting the national aggregate does not incur any penalty in hub evaluations. For each jurisdiction, forecasts must cover the four time horizons required by the CDC Forecast Hubs, from zero to three weeks ahead (Fig.~\ref{fig:fig1}c).
Additionally, to characterize the uncertainty of future disease trajectories, forecasts are produced in a probabilistic format rather than as single-point estimates. As illustrated in Fig.~\ref{fig:fig1}b, for each target week, models generate a predictive distribution represented by 23 specific quantiles (ranging from 0.01 to 0.99). This allows submissions to represent the full range of potential outcomes, from the expected median trajectory to the low-probability tails of the distribution that represent extreme surge events. Together, these requirements result in the generation of 4,784 unique prediction points per infection type (calculated as 52 jurisdictions $\times$ 4 horizons $\times$ 23 quantiles) every week.

The primary data source for inputs and targets was the CDC's Weekly Hospital Respiratory Data (HRD) from the National Healthcare Safety Network (NHSN)~\cite{cdc_nhsn_hrd_2025}. As shown in Fig.~\ref{fig:fig1}a, the study follows a rolling-origin submission scheme where new data are released weekly, encouraging a corresponding weekly forecast submission. During our study period, this resulted in a total of 28 weekly submissions for influenza, 25 for COVID-19, and 24 for RSV, of which 23 were included in the analysis (the first submission was excluded due to an operational error in the submission pipeline).

In our pipeline, preprocessing consisted of extracting jurisdiction-level weekly admission counts, cleaning non-numeric entries, and linearly interpolating any missing weeks to produce a complete time series grid across all 52 jurisdictions. The specific input features available to each pathogen's \ourmethod setup differed based on data availability and relevance. Following the practice of other successful forecasting models\cite{ray2024flusion}, ERA-generated influenza models received weekly incident influenza hospitalizations and were supplemented with a longer history of outpatient syndromic surveillance data from the CDC's ILINet system\cite{cdc_ilinet_2026}(which predates the current target series by more than two decades, offering a longer record of seasonal trends for influenza-like illnesses). COVID-19 models received weekly incident COVID-19 hospitalizations. RSV models received weekly hospitalizations for all three pathogens along with ILINet data to supplement the short history of the target (see Methods 5.1). Beyond this standardized preprocessing, each generated model was free to implement its own additional feature engineering within the \ourmethod sandbox.

The temporal data for each pathogen were divided into three non-overlapping splits to ensure rigorous evaluation: Validation, Retrospective Test, and Prospective Held-out (Fig.~\ref{fig:fig1}d, \suptab{tab:dates}). The Validation block was used during search for optimization; the Retrospective Test block was used post-search for node selection (Methods~4.2); and the Prospective Held-out period represents the live evaluation phase, where model performance was recorded in real-time against data that did not exist during the development phase. For influenza, we chose to not use the most recent season as the Retrospective Test season and instead use the 2022/2023 season. This was because the 2024/2025 season was in many locations unusually high, and we thought that it might penalize model formulations too harshly in the test phase.

The above procedure was applied for the prospective and retrospective studies reported in Results. While our prospective forecasts were generated in real-time using preliminary, less complete data snapshots and evaluated using the finalized data, the retrospective analyses used the updated (revised and back-filled) versions of the data. However, we note that this distinction is not a threat to the study’s validity because retrospective models are compared solely against each other in internal benchmarks rather than being measured against historical real-time hub submissions.

\subsection{Evaluation Metrics}
\label{sec:evaluation_metrics}

This study evaluates forecasts using variants of the Weighted Interval Score (WIS). WIS is a proper scoring rule that penalizes both lack of precision (wide intervals) and poor calibration (ground truth falling outside the predicted quantiles). Small WIS values imply predictions are both sharp and well-calibrated.

Formally, let $F$ denote a predictive distribution of the quantity of interest $Y$ with median $m$ and central prediction intervals at levels $\alpha_1, \ldots, \alpha_K$, where $l_{\alpha_k}$ and $u_{\alpha_k}$ are the $\alpha_k/2$ and $1 - \alpha_k/2$ quantiles of $F$. 
The WIS for an observed value $y$ is defined as~\cite{Bracher2021}
\begin{equation}
\label{eq:wis}
    \text{WIS}(F, y) = \frac{1}{K + 0.5} \left( w_0 \, |y - m| \;+\; \sum_{k=1}^{K} w_k \cdot \text{IS}_{\alpha_k}(F, y) \right),
\end{equation}
where $w_0 = 0.5$, $w_k = \alpha_k / 2$, and the interval score for level $\alpha_k$ is
\begin{equation}
\label{eq:interval_score}
\text{IS}_{\alpha_k}(F, y) = (u_{\alpha_k} - l_{\alpha_k}) + \frac{2}{\alpha_k}(l_{\alpha_k} - y)\,\mathbb{1}(y < l_{\alpha_k}) + \frac{2}{\alpha_k}(y - u_{\alpha_k})\,\mathbb{1}(y > u_{\alpha_k}).
\end{equation}
In this study, forecasts are represented by 23 quantiles corresponding to $K = 11$ central prediction intervals plus the median, yielding $\alpha_k \in \{0.02, 0.05, \ldots, 0.80, 0.90\}$. 

The log WIS variant used for official CDC leaderboard rankings applies the same WIS formula~\eqref{eq:wis} to log-transformed predictions and observations, i.e., inputs are transformed following $\tilde y = \log(1+y)$, where $\log$ is the natural logarithm and the offset of one accommodates possible zero counts. Concretely, let $\tilde F$ denote the predictive distribution of $\log(1 + Y)$ induced by $F$, or, equivalently, the distribution with quantiles $\log(1 + l_{\alpha_k}), \log(1 + u_{\alpha_k})$, and median $\log(1 + m)$. Then, $\text{log\,WIS} = \text{WIS}(\tilde F, \tilde y)$.

Because not all models submit predictions for the same set of tasks---some join later in the season, omit certain reference dates, or cover fewer jurisdictions or prediction horizons---direct comparison of mean absolute scores can be misleading. We therefore report \textit{pairwise relative WIS}~\cite{cramer_evaluation_2022} as the primary model comparison metric in aggregated settings. For each pair of models $(i,j)$, the ratio of their mean scores is computed over the intersection of tasks both models submitted; the relative score for model $i$ is then the geometric mean of these pairwise ratios across all opponents $j \neq i$. We rescale scores so the relevant CDC hub ensemble reports a score of one, and scores below this indicate better-than-ensemble performance. We compute this metric for WIS and log WIS, and report both in~\suptabs{tab:flu_model_summary},~\ref{tab:covid_model_summary}, and~\ref{tab:rsv_model_summary}. Pairwise relative log WIS serves as the primary ranking metric throughout the prospective comparisons in this work, as it does in official CDC comparisons\cite{cdc2026flusight}.

As seen in Equation~\eqref{eq:interval_score}, the interval score strictly partitions into three distinct penalty components: dispersion $(u_{\alpha_k} - l_{\alpha_k})$, which measures the inherent width or uncertainty of the forecast; overprediction $\frac{2}{\alpha_k}(l_{\alpha_k} - y)\,\mathbf{1}(y < l_{\alpha_k})$, which penalizes the model when the true observation falls below the predicted interval; and underprediction $\frac{2}{\alpha_k}(y - u_{\alpha_k})\,\mathbf{1}(y > u_{\alpha_k})$, which applies when the observation falls above it. By summing these components separately across all $K$ intervals via Equation~\eqref{eq:wis}, we can distinguish models that are uninformative because of excessive hedging (dominated by dispersion) from those suffering from severe directional biases (dominated by over- or underprediction) that a single aggregate score would obscure. We plot WIS decomposition in \supfigs{fig:flu_wis_decomp}, \ref{fig:covid_wis_decomp}, \ref{fig:rsv_wis_decomp}. 

Additionally, to complement these aggregate scores with a measure of consistency, we report standardized ranks~\cite{cramer_evaluation_2022}. For each individual forecasting task (a unique combination of jurisdiction, horizon, and reference date), all submitting models are ranked by log WIS and the ranks linearly rescaled to $[0,1]$, where $1$ denotes the best-performing model on that task and $0$ the worst. The resulting distribution of standardized ranks across tasks summarizes how frequently a model performs near the top or bottom of the model set, capturing variability in performance that aggregate mean scores alone would obscure.

An additional qualitative evaluation assessed whether the \ourmethod-generated code followed the methodological instructions provided in each prompt (\suptab{tab:flu_model_summary}). One coauthor (N.G.R.) reviewed the instruction text and corresponding output code and classified each model as `Success', `Partial fail’, or `Fail'. Models classified as Success’ followed all instructions that were feasible within the connected sandbox, excluding requests to access unavailable data or libraries. `Partial fail' indicated that most major components were implemented but one or more substantive methodological requirements were omitted or approximated (for example, omitting a requested Kalman filter or implementing an approximation rather than an explicit compartmental model). `Fail’ indicated that the requested methodology was largely or entirely disregarded (for example, implementing gradient boosting when an SIR-type model was requested).

\subsection{The \ourmethod Framework}
\label{sec:era_framework}

Empirical Research Assistance (\ourmethod) is an autonomous system that automates development of empirical software for scorable scientific tasks\cite{aygun_ai_2025}.
It searches over the space of code implementations using a tree search algorithm: each node in the tree corresponds to a concrete implementation of the required task, including specific combinations of feature engineering, data preprocessing, and model architecture, and is scored against a predefined objective function.
Node selection for tree expansion balances exploitation of the highest-scoring candidates with exploration of less-visited ones via a Predictor + Upper Confidence bound applied to Trees (PUCT) algorithm.

The search is guided by an LLM—specifically, by one of the Gemini family of models—acting as an agent that generates, tests, and refines Python code. Starting from a natural language problem statement, the agent proposes initial architectures and iteratively improves them based on execution feedback and performance scores (Fig.~\ref{fig:fig2}). 
This branching structure allows the system to explore multiple modeling paradigms simultaneously (executing on ten parallel threads), learning from the successes and failures of previous nodes to discover increasingly effective solutions. To manage computational resources and ensure system stability, \ourmethod search trajectories are bounded by dual constraints: a maximum number of explored nodes and a cumulative execution time limit within the sandbox environment.

To maintain scientific rigor and prevent data leakage, \ourmethod operates within a secure, immutable evaluation harness. This harness contains relevant ground truth datasets and evaluation logic but is entirely hidden from the LLM agent. The agent can submit code to the harness and receive performance metrics and error logs, but it cannot modify the evaluation protocol or access ground-truth data from the held-out periods. This `blind' evaluation ensures discovered models are robust and prevents the system from `hacking' evaluation criteria.

\subsection{\ourmethod Configuration for Epidemiological Forecasting}
\label{sec:epi_configuration}

This section describes two task-specific design choices made to apply \ourmethod to multi-pathogen forecasting: a structured prompting strategy across multiple search modes, and a two-stage model selection scheme that separates search-time scoring from post-search node selection. Both are part of the pipeline used to generate models for prospective submission to the CDC hubs.

\subsubsection{Method Prompt Sources}
\label{sec:method_prompt_sources}

Each independent \ourmethod search used for this analysis ran for up to $2,500$ generated nodes or until a cumulative sandbox execution time limit of $2500$ hours over ten parallel threads was reached, whichever came first. 
Because candidate evaluations within each search tree were executed across ten parallel sandbox threads, the 2,500-hour cumulative compute cap corresponded to a maximum wall-clock runtime of approximately 10.4 days per search (which only a few compute-intensive searches reached; most terminated earlier upon reaching 2,500 nodes). 
Total computational cost per \ourmethod search depends on LLM token usage (which scales with prompt length, generated code length, and model tier) and sandbox CPU execution time. 
Because the prospective searches used Gemini 2.5 Flash on CPU sandboxes with a prompt scaffold similar to our prior study\cite{aygun_ai_2025}, per-node LLM token usage was comparable to the values reported there (approximately 10,000 request tokens and 3,000 response tokens per node).

Within these total node and time bounds, the LLM agent generated candidate code that was executed inside a sealed evaluation harness described in the previous section; only the resulting score and execution logs were returned to the agent, with all data and evaluation logic immutable from the agent's perspective.

For each pathogen and forecast origin, searches were initialized under one of three prompt regimes. `Scientist-inspired' prompts provide the agent with a description of an existing methodology---drawn from metadata of models submitting to one of the forecasting hubs (e.g., FluSight) or summarized from a published paper---and instruct it to implement and refine that method. `Recombination' prompts provide descriptions of two distinct existing methods and instruct the agent to combine their logic into a single hybrid model that encompasses the strengths of both components\cite{aygun_ai_2025}. `AI-led' searches receive either a method prompt generated by Gemini Deep Research or only the task description, leaving the search unconstrained over architectures.

Influenza benefited from the richest existing modeling ecosystem resulting in 56 unique prompts and modeling strategies: its  prompt set spanned active FluSight and COVIDHub submissions, historically high-performing methods identified by experts from the published literature, modeling suggestions generated by Gemini Deep Research, and a structured grid of mechanistic–statistical recombinations. For COVID-19, \ourmethod prompts primarily targeted active COVIDHub submissions, supplemented by a subset of high-performing models carried over from the prior retrospective study~\cite{aygun_ai_2025}. 
This was the inaugural season of RSVHub, an infection that is less modeled in published literature. We used 14 prompts targeting the small number of available RSVHub methods alongside Gemini Deep Research-suggested architectures; the limited pool of viable base models precluded systematic recombination.

The canonical base prompt used as a scaffold to seed all influenza \ourmethod runs is included in Supplementary
Section~\ref{sec:problem_statement_flu}, with infection-specific variants for COVID-19 and RSV in Supplementary
Section~\ref{sec:problem_statement_variations}. Additional specific modeling instructions provided to \ourmethod for each method can be found in the
GoogleHub repository\footnote{\url{https://github.com/google-research/google-research/tree/master/epi_forecasts}}.

\subsubsection{Two-stage Model Selection}
\label{sec:model_selection}

\textbf{Optimization metric and moving-window aggregation.} During search, candidates were scored using WIS on the validation block of the relevant pathogen-season split (Fig.~\ref{fig:fig1}d, purple). To avoid rewarding candidates that fit narrowly to a single forecast origin, each candidate's validation WIS was aggregated across multiple rolling origins within the validation block. This encourages the search to select models that perform consistently across epidemic phases in a season (growth, peak, and decline). This aggregated validation WIS drove node selection within the PUCT tree search. 


\noindent \textbf{Post-search node selection.} A single \ourmethod search produces a tree of candidates whose aggregated validation WIS scores are known by construction. 
To choose one representative model per search while protecting against overfitting to the validation block, we evaluated every node post-hoc on a held-out retrospective test block (Fig.~\ref{fig:fig1}d, orange)---a period never seen by the agent during the search---and selected the node minimizing
\begin{equation}
    \label{eq:selection_score}
    \text{Selection Score} = \text{Validation WIS} + 2 \times \text{Retrospective Test WIS}.
\end{equation}
The $2\times$ weighting on the retrospective test term places greater emphasis on out-of-search generalization than on in-search fit.

\subsection{Real-time Prospective Implementation}
\label{sec:prospective_implementation}

\subsubsection{Pathogen-specific Implementations}
\label{sec:pathogen_implementations}
The application of \ourmethod varied across the three target pathogens due to differences in data availability and hub maturity. Influenza benefited from a mature forecasting ecosystem, allowing for a large pool of candidate models (see Methods 4.1).
For COVID-19, in addition to new \ourmethod-generated models, five models were carried over from a retrospective study\cite{aygun_ai_2025} and fine-tuned on the larger validation dataset available for this prospective season.
RSV presented a qualitatively different challenge. Historical surveillance data for this pathogen are extremely limited, the published forecasting literature is sparse, and the short available hospitalization record constrained the size of meaningful validation and test splits. To compensate, we provided \ourmethod models with concurrent COVID-19 and influenza hospitalization time series as auxiliary inputs to harness for generalizable cross-pathogen patterns. This resulted in a comparatively smaller pool of models (see Supplementary Fig.~\ref{fig:selection_diagram}).

\subsubsection{Model Selection}
\label{sec:prospective_model_selection}

Initially, to construct a large library of reasonable forecasting candidates, we executed a wide array of independent \ourmethod search experiments prior to the start of the prospective season using the prompt sources outlined in Methods 4.1. 
Then, the optimal `best node' from each independent search tree was identified using the compound selection score \eqref{eq:selection_score}. From this pool, a subset was curated for inclusion in GoogleHub based on a combination of their performance scores and methodological diversity. Finally, a smaller, high-performance subset was selected to form the official \texttt{Google-SAI} submission ensemble for each pathogen. The model selection process is outlined in Supplementary Fig.~\ref{fig:selection_diagram}. 

For several modeling strategies, multiple independent \ourmethod search runs were conducted. When an initial run failed to reproduce the target methodology faithfully---as assessed by expert review---additional runs were launched with modified configurations to increase the likelihood of a faithful implementation. In some cases, a non-faithful adaptation nonetheless produced strong forecasting performance and was retained in the model pool alongside its more faithful counterparts. This accounts for the presence of multiple versions of certain models in GoogleHub and in \suptabs{tab:flu_model_summary}, \ref{tab:covid_model_summary}, \ref{tab:rsv_model_summary} (indicated by version $v$).

We note that, for influenza, a strategy shift occurred around January 24, 2026; upon learning that the CDC utilizes log WIS rather than WIS as the primary metric for official evaluations, we updated our selection process to prioritize models based on Log WIS performance. We did not rerun \ourmethod to search for models, rather, we calculated log WIS scores post-hoc and used these to select best models for inclusion in GoogleHub and in \gflumodel. We examine the sensitivity of downstream forecast behavior to this choice of optimization metric in Results. 

\subsubsection{Generating Prospective Forecasts and Ensembles}
\label{sec:prospective_forecasts}
The prospective evaluation phase consisted of approximately 25 weeks of live submissions to the three forecasting hubs. 
Selected models (described in the previous section) were frozen at the season start---\ourmethod was run once to generate each model, then the same model code was retrained each week using the real-time surveillance data released that week. Retraining models consisted solely of refitting their statistical parameters (e.g., autoregressive coefficients, regression weights) and tuning internal hyperparameters (e.g., lag lengths, window sizes) via rolling validation loops executed in the training and validation windows. Models then generated forecasts for each week's reference date, including predictions for all four horizons and all 52 jurisdictions, which were submitted to GoogleHub weekly. 

Final ensemble forecasts for all three pathogens were generated using a median ensemble approach, where the prediction for each jurisdiction and horizon was calculated as a direct quantile average of the 23 quantiles across all participating component models, still ensuring monotonicity of the outputs.

\subsection{Retrospective Experiment Design}
\label{sec:retrospective_experiments}

\subsubsection{Analysis of Optimization Metrics}
\label{sec:optimization_metrics}
We investigated how the choice of \ourmethod's internal optimization metric dictates stability of the search and the behavior of output forecasts. 
For each optimization metric, we conducted five independent \ourmethod searches, yielding 15 searches in total. All searches used the same influenza forecasting task, prompt, data, and computational budget, such that only the optimization metric differed between conditions. During search, candidate nodes were scored on the validation set using the assigned optimization metric. Following search completion, all nodes from each search tree were evaluated post-hoc on the retrospective test set using the same metric, and one final node per search was selected using Equation~\eqref{eq:selection_score}, with the validation and retrospective test WIS terms replaced by the corresponding optimization metric. The resulting five selected models for each condition were then evaluated using all three metrics, allowing us to assess both performance on the metric targeted during search and performance under the alternative scoring rules.

We run \ourmethod using three additional optimization metrics from the two principal families of proper scoring rules for probabilistic forecasting~\cite{Gneiting2007}. 
First, the Continuous Ranked Probability Score (CRPS) is a distance-based metric that measures the integrated squared difference between the cumulative distribution function (CDF) of the forecast $F$, and the empirical CDF of the observation $y$. It is defined as
\begin{equation}
    \label{eq:crps}
    \text{CRPS}(F, y) = \int_{-\infty}^{\infty} (F(x) - \mathbb{1}(x \ge y))^2 \, dx,
\end{equation}
where $\mathbb{1}(x \ge y)$ is an indicator function that equals one if $x \ge y$ and zero otherwise. CRPS penalizes both over- and underprediction continuously, scaling with the absolute magnitude of the forecast error. The WIS \eqref{eq:wis} serves as a discrete approximation of this exact CRPS integral, evaluating the predictive distribution across a finite set of quantiles.

Second, Log-scale CRPS adapts this distance metric for epidemiological targets that span multiple orders of magnitude, applying greater sensitivity to relative rather than absolute errors. It is computed by applying a natural logarithm transformation to both the predictive distribution and the ground truth observation prior to scoring
\begin{equation}
    \label{eq:log_crps}
    \text{Log-CRPS}(F_{\log}, y) = \int_{-\infty}^{\infty} (F_{\log}(x) - \mathbb{1}(x \ge \log(y)))^2 \, dx,
\end{equation}
where $F_{\log}$ represents the CDF of the log-transformed predictions. In practice, to accommodate zero-counts in incidence data, a $+1$ offset is added to both predictions and observations prior to transformation. This is the continuous analogue of log WIS. 

The Logarithmic Score (Log Score) is a local proper scoring rule based on the predictive probability density or mass function, $f$. Because the \ourmethod evaluation harness requires a metric to be minimized (a loss function), we evaluate the negative log-probability assigned to the true observation $y$
\begin{equation}
    \label{eq:log_score}
    \text{Log Score}(f, y) = -\log(f(y)).
\end{equation}
Unlike distance-based metrics, the Log Score evaluates only the probability mass placed exactly on the observed outcome. Consequently, if a model's predictive distribution assigns zero probability to the ground truth outcomes, it incurs an effectively infinite penalty (as $-\log(0) \to \infty$), regardless of how close the bulk of the distribution falls to the observed value.

To enable the calculation of Log Scores during these optimization experiments, we modified the predictive output format. Rather than generating the standard 23 quantiles, models in this subset were required to output 1,000 predictive samples per target. This sample-based format allowed for the use of the \texttt{scoringrules}\cite{zanetta_scoringrules_2024} library, which provides robust implementations for Log Score, CRPS, and Log-CRPS calculations. To guarantee data integrity, all retrospective experiments were conducted using an immutable evaluation harness that isolated the LLM agent from the ground-truth data (Methods 3).

\subsubsection{Evaluation of \ourmethod Instruction Following}
\label{sec:instruction_following}
\noindent We conducted a retrospective sensitivity analysis to assess how instruction-following fidelity varied with the LLM used for code generation and with the inclusion of an in-loop fidelity judge. We compared four search configurations: (1) Gemini 2.5 Flash without an in-loop judge, (2) Gemini 2.5 Flash with an in-loop judge, (3) Gemini 3 Flash without a judge, and (4) Gemini 3 Pro without a judge. The in-loop judge (itself a call to Gemini 2.5 Flash) was evaluated only with Gemini 2.5 Flash and was not crossed with the two Gemini 3 configurations.

We applied each configuration to four established FluSight methods spanning distinct modeling strategies: (1) a pure ML gradient-boosting model \texttt{UMass-gbqr}\cite{ray2024flusion}, which uses feature engineering and gradient boosting; (2) a package-dependent spatial time-series model \texttt{UGA\_flucast-INFLAenza}\cite{fox_uga_flucast_inflaenza}; (3) a hierarchical compartmental model \texttt{Cornell\_JHU-hierarchSIR}\cite{alleman2026data}; and (4) a probability-generating-function-based model \texttt{NU-PGF\_FLUH}\cite{stonge_pgf_fluh}. Each method was used as the target for each of the four \ourmethod configurations described above. Three independent search trajectories were conducted for each configuration–method combination under matched computational budgets, giving 48 searches in total.

All searches were subject to the same stopping criteria. A search terminated after either 2,500 explored nodes or 2,500 cumulative hours of parallel sandbox runtime, whichever occurred first. These dual criteria bounded both the breadth of the search and the computational time available to models with different runtime requirements. Within each trajectory, the final model was selected according to the selection score defined in Equation~\eqref{eq:selection_score}. 

For the configuration incorporating an in-loop judge, Gemini 2.5 Flash was used both to generate code and, through a separate and independent API call, to assess fidelity. After code had been generated at each node, the judge received the method-specific seed instruction together with the candidate model code (Supplementary Section~\ref{SIsec:judge_prompts}). It then assigned an instruction-fidelity score from one to ten. Candidate implementations meeting the prespecified threshold of seven proceeded to full validation and received a validation-based score. Candidates below the threshold were not validated and were instead assigned a penalty value of 1,000 when calculating the in-search validation WIS. The judge therefore acted as a binary gate within the otherwise single-objective search, preventing candidates assessed as non-compliant from advancing on the basis of predictive performance alone.

After search completion, the selected model from every trajectory was evaluated by the same external post-search judge. Unlike the generic in-loop judge, the external judge used a method-specific rubric developed by an expert epidemiologist (N.G.R.). Each rubric identified the principal features required for the generated code to constitute a faithful implementation of the corresponding reference method. On the basis of the generated code and this rubric, the external judge classified each implementation as a `Match,' `Partial Match,' or `No Match.' This post-search assessment did not affect model selection and was used only to evaluate the methodological fidelity of the selected models.

We assessed concordance between the external judge and independent classifications made by N.G.R. on a subset of ten retrospectively generated models. 
The classifications agreed in all but one case. In the discordant case, the generated code nominally implemented the requested method in a \texttt{try} block but fell back to degenerate output via an \texttt{except} clause, which the judge assessed as compliant based on the code structure alone.
This case identified a limitation of code-only fidelity assessment when the implemented fallback behavior differs from the apparent primary implementation.

%% file: Sections/6-supp-figs.tex
\section{Prospective Model Generation}

\begin{figure}[H]
    \centering
    \includegraphics[width=\linewidth]{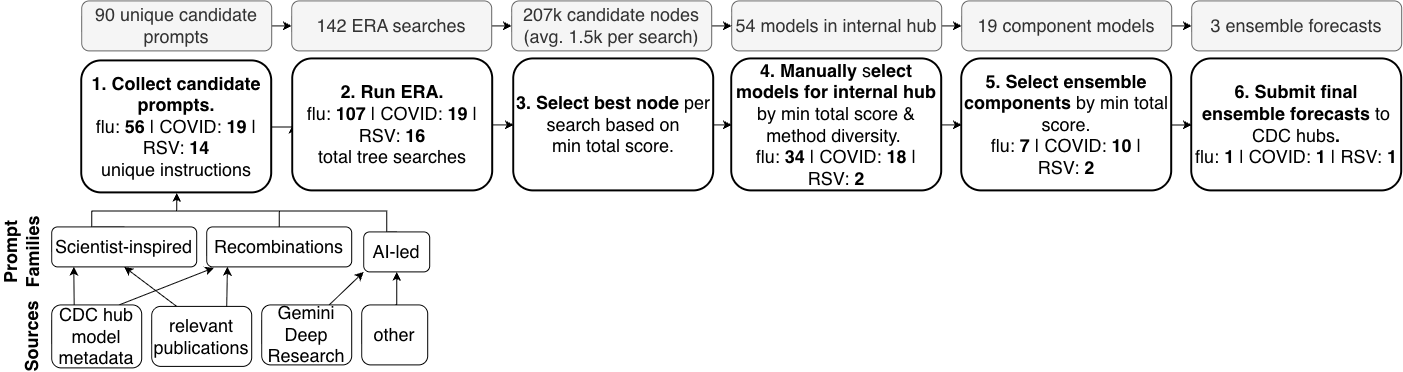}
    \caption{Flowchart showing the model selection process for the prospective study, from candidate prompt construction for \ourmethod through internal hub model selection and final ensemble inclusion and submission for influenza, COVID and RSV.}
    \label{fig:selection_diagram}
\end{figure}

\begin{table}[H]
\centering
\caption{Validation and test periods for each pathogen. The validation period was used during \ourmethod model search to calculate an evaluation score for selecting nodes in the tree for further exploration. The retrospective test period was entirely unseen during search and was used to calculate scores for best-node selection (see Equation~\eqref{eq:selection_score}). Validation and test periods for all three pathogens are illustrated in Fig.~\ref{fig:fig1}.}
\label{tab:dates}
\begin{tabular}{@{}lll@{}}
\toprule
\textbf{Pathogen} & \textbf{Validation Period} & \textbf{Retrospective Test Period} \\
\midrule
Influenza & 2023-10-13 to 2024-05-15 \& & 2022-10-15 to 2023-05-15 \\
          & 2024-10-15 to 2025-05-15 & \\
COVID-19& 2023-06-01 to 2024-06-01 & 2024-06-01 to 2025-09-01 \\
RSV& 2024-10-01 to 2025-07-01 & 2025-07-07 to 2025-11-07 \\
\bottomrule
\end{tabular}
\end{table}

\section{Prospective Model Performance Analysis}

This section contains additional analysis figures and tables for the models submitting prospectively to the three CDC hubs and for component models submitted to GoogleHub. 
All analyses are restricted to 52 jurisdictions (50 states, Washington D.C., and Puerto Rico; national-level forecasts are excluded) and to forecast horizons 0–3 weeks ahead (horizon -1 excluded). 

\subsection{Influenza} \label{SIsec:flu}

Here, we present individual analysis for influenza forecasts, both for our component models and for those models submitting to the FluSight Forecast Hub. Models submitting influenza forecasts are analyzed over reference dates from 22 November 2025 to 27 May 2026 (28 reference dates, covering a total of $5,824$ forecasting tasks). Analyses are restricted to models submitting for $\geq80\%$ of the required task space, i.e. $\geq4,659$ tasks (unless otherwise specified). Under this criterion, 43 hub-submitting models are eligible for analysis out of a total of 57 models which submitted at least one task prediction to the FluSight Forecast Hub this season.

\begingroup
\small 
\setlength{\tabcolsep}{4pt}
\setlength{\aboverulesep}{0pt}
\setlength{\belowrulesep}{0pt}
\renewcommand{\arraystretch}{1.4}
\begin{longtable}{>{\raggedright\arraybackslash}m{4.1cm}
                  >{\centering\arraybackslash}m{0.8cm}   
                  >{\centering\arraybackslash}m{1.0cm}   
                  >{\centering\arraybackslash}m{1.0cm}   
                  >{\centering\arraybackslash}m{1.0cm}   
                  >{\centering\arraybackslash}m{1.0cm}   
                  >{\centering\arraybackslash}m{1.0cm}   
                  >{\centering\arraybackslash}m{1.0cm}   
                  >{\centering\arraybackslash}m{1.8cm}   
                  >{\centering\arraybackslash}m{1.1cm}}  
    \caption{Summary performance of \ourmethod-generated component models and the top-ranked eligible models submitting to FluSight Forecast Hub. 
    Model names prefixed by \texttt{*} are included in \gflumodel; the \emph{Type} column labels each \ourmethod\ model as \emph{Scientist-inspired}, \emph{Recombination}, or \emph{AI-led}, while hub models are marked \emph{CDC}. 
    Columns report, per model: the number of scored tasks; pairwise relative log WIS and WIS (relative to the FluSight ensemble $=1$); mean log WIS and mean WIS; mean absolute error of the median forecast (MAE); and empirical coverage of the 95\% prediction interval (95\% Cov.).
    The absolute-error columns (Mean log WIS, Mean WIS, MAE) are shaded independently from green (lowest, best) to red (highest, worst) across models; the pairwise relative columns use a diverging scale centered at one, where values below one (green) indicate better and above one (red) worse performance than the ensemble; 95\% coverage is left unshaded, as its target is ${\approx}95\%$ rather than a minimum or maximum.
    Models are ranked by pairwise relative log WIS (ascending).
    For each pair of models, pairwise relative score ratios are calculated over the intersection of tasks both models submitted, then geometrically averaged across all opponents, so the metric is not affected by differences in the set of tasks each model chose to forecast~\cite{Bracher2021}. 
    Because pairwise relative WIS is computed on the task overlap between each pair of models, which varies across pairs, a model can have a lower mean log WIS than another yet a higher pairwise relative log WIS.
    The \emph{Replica judgment} column gives an expert assessment of whether each `scientist-inspired' \ourmethod replica's code followed the model's stated methodology: \ding{51}~Success, \LEFTcircle~Partial fail, \ding{55}~Fail.}
\label{tab:flu_model_summary} \\
    \toprule
        \textbf{Model} & \textbf{n tasks} & \textbf{Rel. log WIS} & \textbf{Mean log WIS} & \textbf{Rel. WIS} & \textbf{Mean WIS} & \textbf{MAE} & \textbf{95\% Cov.} & \textbf{Type} & \textbf{Replica judgment} \\
    \midrule
    \endfirsthead
    
    \toprule
        \textbf{Model} & \textbf{n tasks} & \textbf{Rel. log WIS} & \textbf{Mean log WIS} & \textbf{Rel. WIS} & \textbf{Mean WIS} & \textbf{MAE} & \textbf{95\% Cov.} & \textbf{Type} & \textbf{Replica judgment} \\
    \midrule
    \endhead
    
    \input{tables/internal_flu_crosshub_rel.tex}
\end{longtable}
\endgroup

\begin{figure}[H]
    \centering
    \includegraphics[width=.99\linewidth]{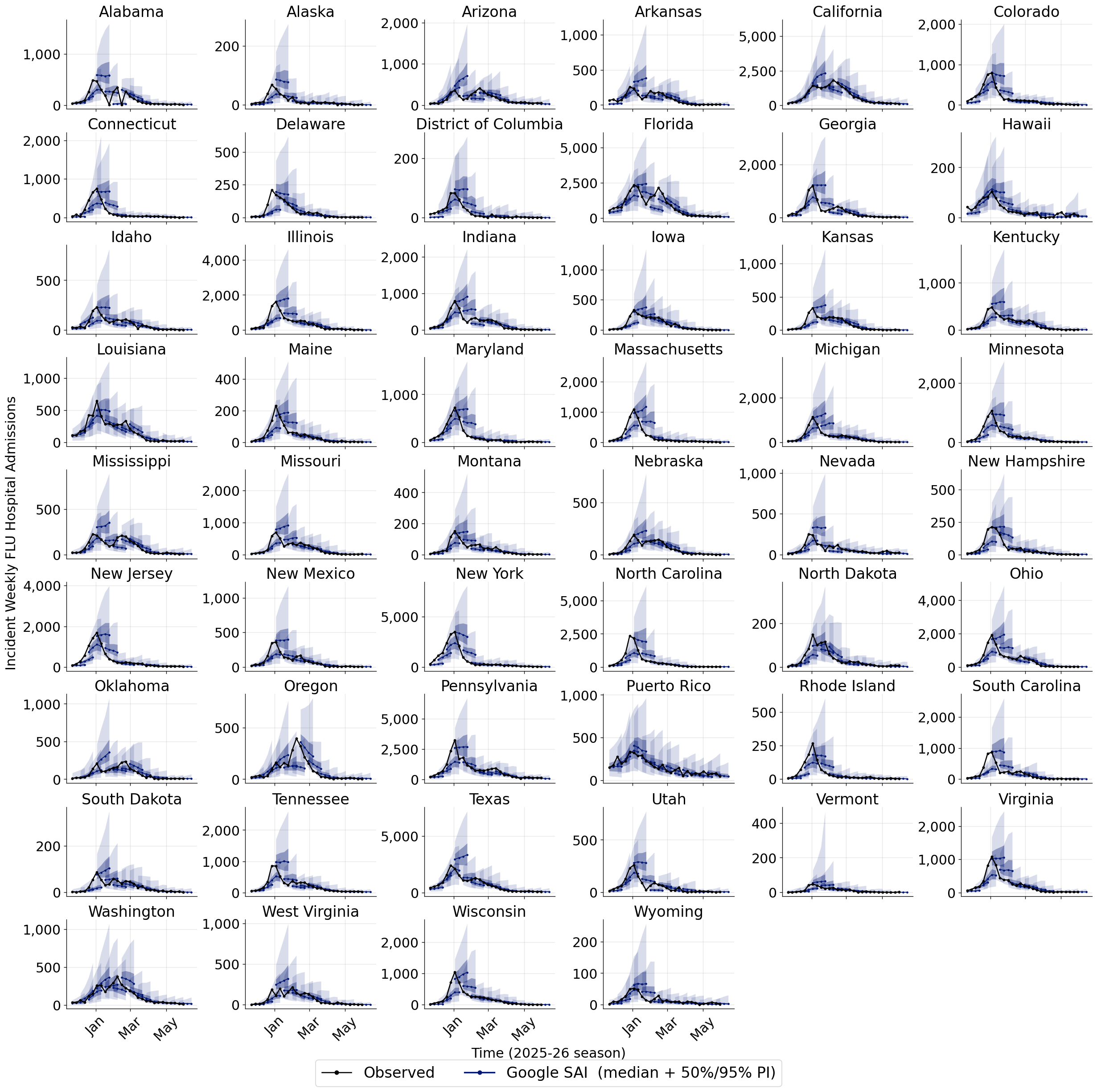}
    \caption{National weekly observed influenza hospitalizations (black) and \gflumodel~ forecast submissions by jurisdiction over the 2025–26 season with median (dark blue points) and corresponding 50\% and 95\% prediction intervals (blue shaded regions). Only every second set of submitted horizon forecasts is plotted for improved readability. See Fig.~\ref{fig:fig3}a for a larger example forecast plot for California, including all reference dates.}
    \label{fig:flu_forecasts}
\end{figure}

\begin{figure}[H]
    \centering
    \includegraphics[width=.9\linewidth]{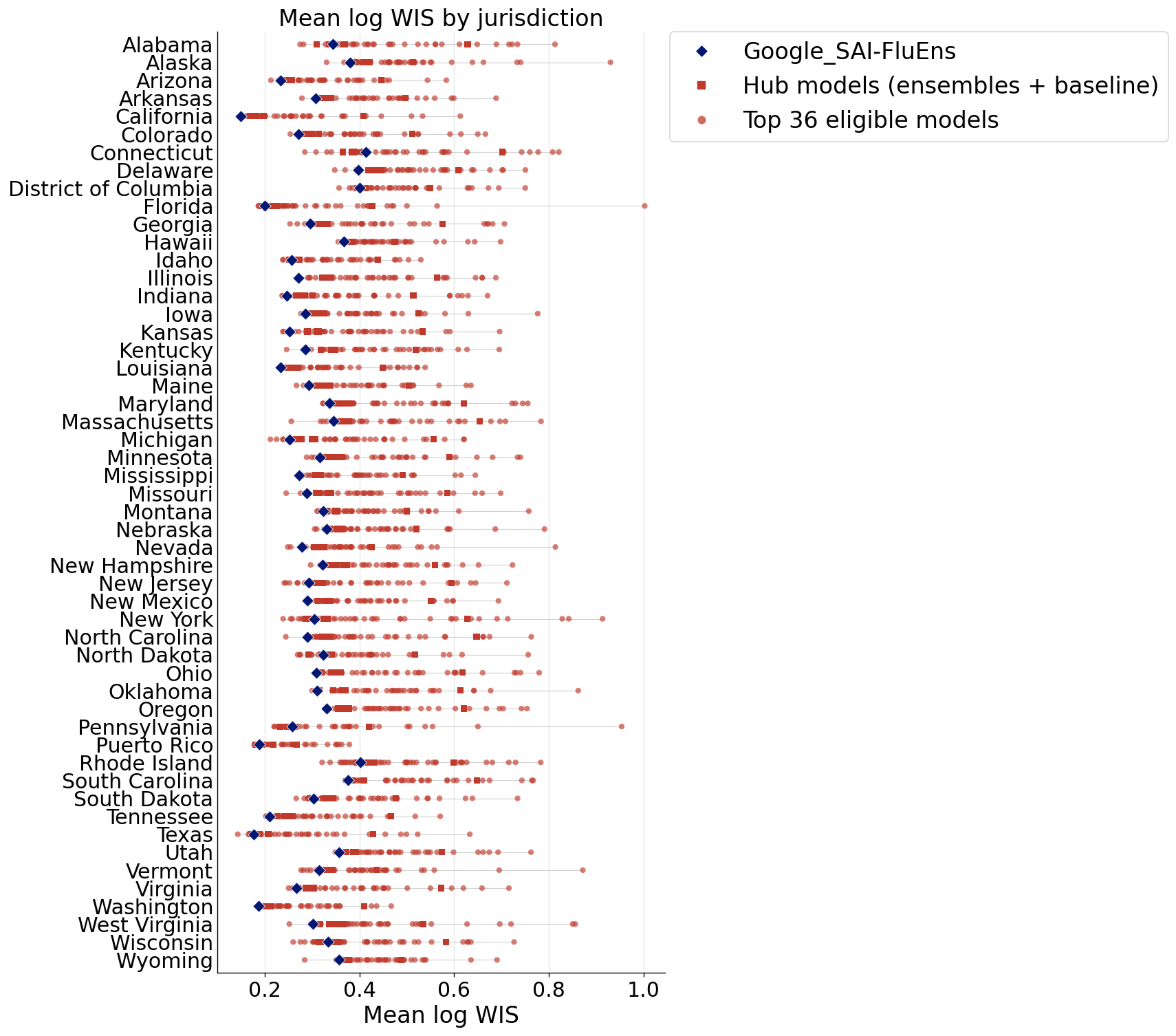}
    \caption{Season-average forecasting skill by jurisdiction for FluSight Forecast Hub-submitting models. Each row represents a US jurisdiction and each marker a model's mean log WIS for that jurisdiction (lower is better; jurisdictions ordered alphabetically). Red circles show eligible hub-submitting models---those meeting the inclusion criteria ($\geq 80\%$ task coverage); the CDC hub ensemble and baseline are shown as red squares and \gflumodel\ as blue diamonds.}
    \label{fig:flu_performace_by_location}
\end{figure}

\begingroup
\small 
\setlength{\tabcolsep}{4pt}
\setlength{\aboverulesep}{0pt}
\setlength{\belowrulesep}{0pt}
\renewcommand{\arraystretch}{1.4}
\begin{longtable}{>{\raggedright\arraybackslash}m{5.0cm}
                  >{\centering\arraybackslash}m{1.2cm}
                  >{\centering\arraybackslash}m{1.5cm}
                  >{\centering\arraybackslash}m{1.5cm}
                  >{\centering\arraybackslash}m{1.5cm}
                  >{\centering\arraybackslash}m{1.5cm}}
    \caption{Mean log WIS by model and prediction horizon for models submitting to FluSight Forecast Hub. Models are ranked by the sum of mean log WIS at horizons 0 and 1 (ascending). Within each horizon column, cells are shaded independently from green (lowest log WIS, best performance) to red (highest log WIS, worst performance); colors are not comparable across columns. The \textit{n tasks} column reports total number of scored forecast tasks submitted across all horizons. Two \texttt{CFA\_Pyrenew} models that do not submit forecasts for all four horizons are included; empty cells indicate horizons for which no forecast was submitted.}
    \label{tab:flu_logwis_by_horizon} \\
    \toprule
    \textbf{Model} & \textbf{n tasks} & \textbf{0} & \textbf{1} & \textbf{2} & \textbf{3} \\
    \midrule
    \endfirsthead
    
    \toprule
    \textbf{Model} & \textbf{n tasks} & \textbf{0} & \textbf{1} & \textbf{2} & \textbf{3} \\
    \midrule
    \endhead
    
    \input{tables/flu_logwis_by_horizon.tex}
\end{longtable}
\endgroup

\begin{figure}[H]
    \centering
    \includegraphics[width=\linewidth]{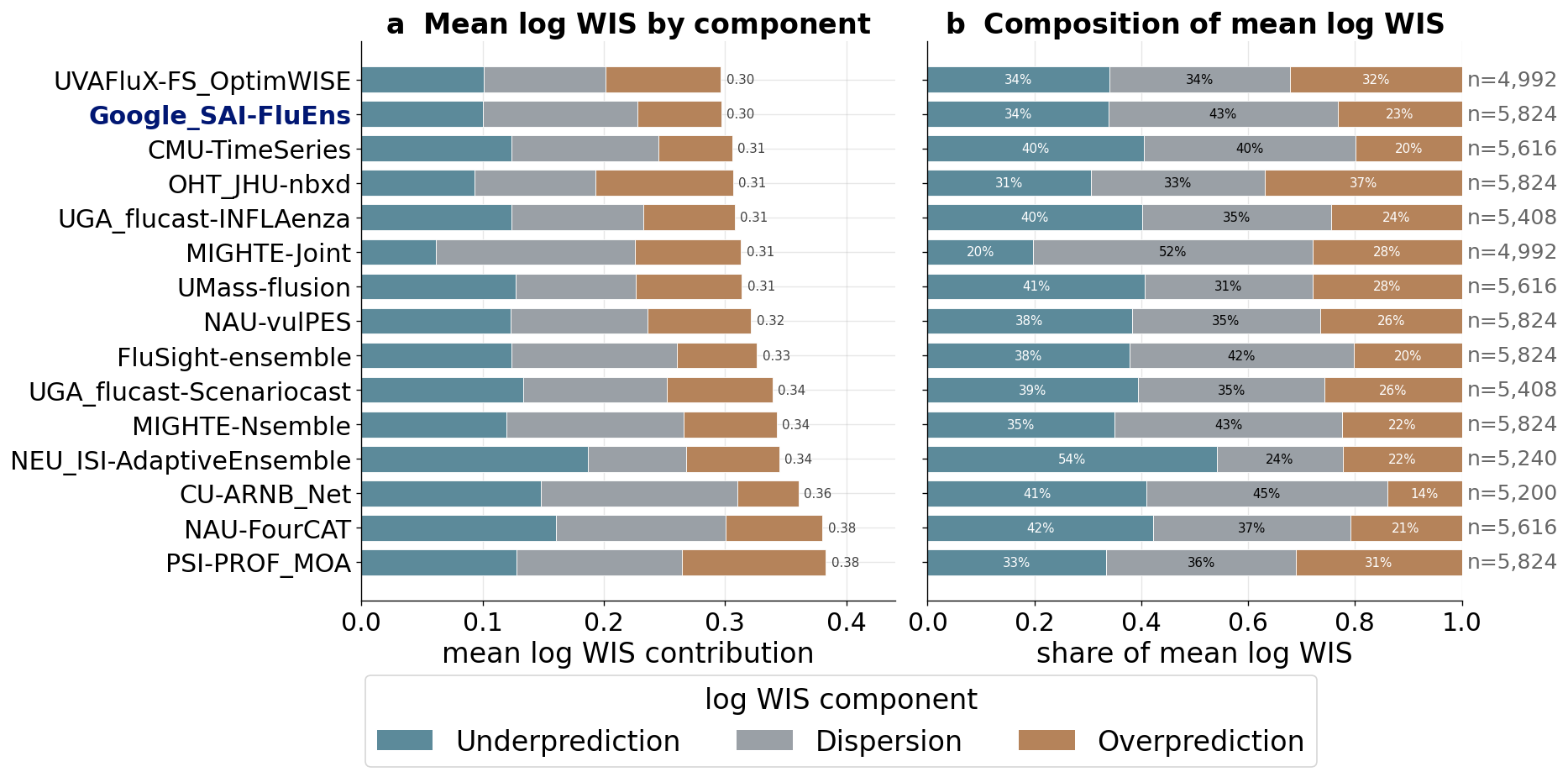}
    \caption{Log WIS decomposition into underprediction, dispersion, and overprediction for the top 15 ranked eligible models submitting scorable predictions for $\geq 80\%$ of the task space to the CDC's FluSight Forecast Hub for the 2025-26 season. Models listed in order of ascending absolute WIS (i.e., not considering number of submitted tasks). \textbf{a.} Absolute WIS contributions for each model. \textbf{b.} WIS contributions normalized to one, with number of tasks submitted by each model on the right.
}
    \label{fig:flu_wis_decomp}
\end{figure}

\begin{figure}[H]
    \centering
    \includegraphics[width=.75\linewidth]{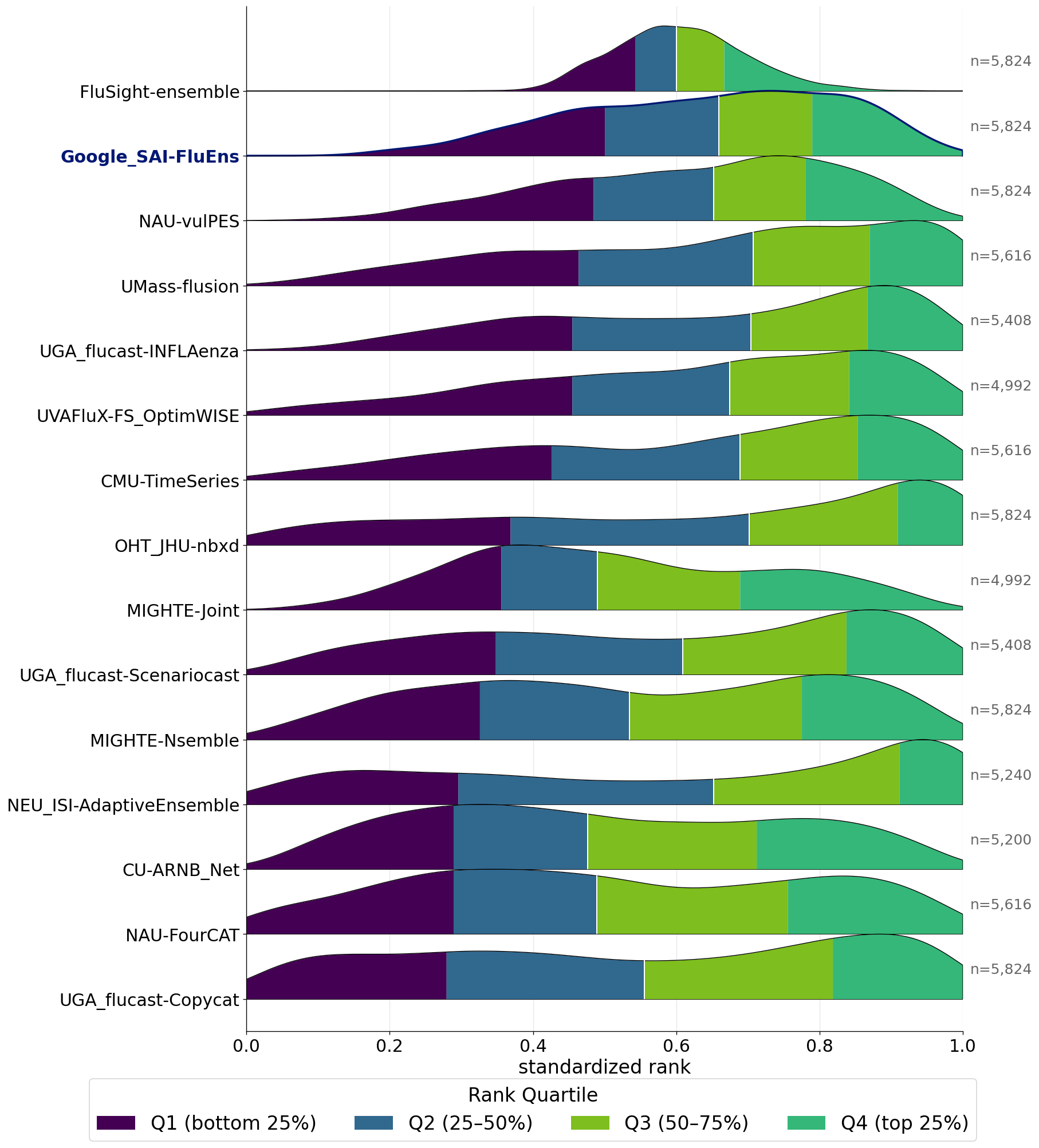}
    \caption{Standardized log WIS rank distributions of top 15 ranked eligible models submitting scorable predictions for $\geq 80\%$ of the task space to the CDC's FluSight Forecast Hub for the 2025-26 season. A standardized rank of one indicates that the model had the best log WIS for that particular task (location, target, and horizon for that reference date), and a value of zero indicates it had the worst log WIS of submitting models. Density plots show interpolated distributions of standardized ranks achieved by each model for every forecast. Quartiles are colored from purple (bottom quarter, i.e., worst model ranks), through blue and light green to bright green (top quarter, i.e., best model ranks). Medians are represented by vertical white lines. Models are ordered by lowest quartile, with models that rarely had a low rank near the top. \gflumodel is outlined in dark blue.}
    \label{fig:flu_rank_distribution}
\end{figure}



\subsection{COVID-19} \label{SIsec:covid}

This section includes analyses for COVID-19 forecast models, analyzed between 13 December 2025 and 27 May 2026 (25 reference dates, covering a total of 5,200 possible prediction tasks). Analyses are restricted to models submitting for $\geq80\%$ of the required task space, i.e., $\geq4,160$ tasks (unless otherwise specified). Under this criterion, 12 hub-submitting models are eligible for analysis out of a total of 17 models which submitted at least one task prediction to COVIDHub this season.

\begingroup
\small 
\setlength{\tabcolsep}{4pt}
\setlength{\aboverulesep}{0pt}
\setlength{\belowrulesep}{0pt}
\renewcommand{\arraystretch}{1.4}
\begin{longtable}{>{\raggedright\arraybackslash}m{4.2cm}
                  >{\centering\arraybackslash}m{0.8cm}   
                  >{\centering\arraybackslash}m{1.0cm}   
                  >{\centering\arraybackslash}m{1.0cm}   
                  >{\centering\arraybackslash}m{1.0cm}   
                  >{\centering\arraybackslash}m{1.0cm}   
                  >{\centering\arraybackslash}m{1.0cm}   
                  >{\centering\arraybackslash}m{1.0cm}   
                  >{\centering\arraybackslash}m{1.8cm}}  
    \caption{Summary performance of \ourmethod-generated component models and the top-ranked eligible models submitting to COVIDHub. 
    Model names prefixed by \texttt{*} are included in \gcovidmodel; the \emph{Type} column labels each \ourmethod\ model as \emph{Scientist-inspired}, \emph{Recombination}, or \emph{AI-led}, while hub models are marked \emph{CDC} (or \emph{CDC ensemble}). 
    Columns report, per model: the number of scored tasks; pairwise relative log WIS and WIS (relative to the COVIDHub ensemble $=1$); mean log WIS and mean WIS; mean absolute error of the median forecast (MAE); and empirical coverage of the 95\% prediction interval (95\% Cov.).
    The absolute-error columns (Mean log WIS, Mean WIS, MAE) are shaded independently from green (lowest, best) to red (highest, worst) across models; the pairwise relative columns use a diverging scale centered at one, where values below one (green) indicate better and above one (red) worse performance than the ensemble; 95\% coverage is left unshaded, as its target is ${\approx}95\%$ rather than a minimum or maximum.
    Models are ranked by pairwise relative log WIS (ascending).
    For each pair of models, pairwise relative score ratios are calculated over the intersection of tasks both models submitted, then geometrically averaged across all opponents, so the metric is not affected by differences in the set of tasks each model chose to forecast~\cite{Bracher2021}. 
    Because pairwise relative WIS is computed on the task overlap between each pair of models, which varies across pairs, a model can have a lower mean log WIS than another yet a higher pairwise relative log WIS.}
    \label{tab:covid_model_summary} \\
    \toprule
        \textbf{Model} & \textbf{n tasks} & \textbf{Rel. log WIS} & \textbf{Mean log WIS} & \textbf{Rel. WIS} & \textbf{Mean WIS} & \textbf{MAE} & \textbf{95\% Cov.} & \textbf{Type} \\
    \midrule
    \endfirsthead
    \toprule
        \textbf{Model} & \textbf{n tasks} & \textbf{Rel. log WIS} & \textbf{Mean log WIS} & \textbf{Rel. WIS} & \textbf{Mean WIS} & \textbf{MAE} & \textbf{95\% Cov.} & \textbf{Type} \\
    \midrule
    \endhead
    \input{tables/internal_covid_crosshub_rel.tex}
\end{longtable}
\endgroup

\begin{figure}[H]
    \centering
    \includegraphics[width=0.99\linewidth]{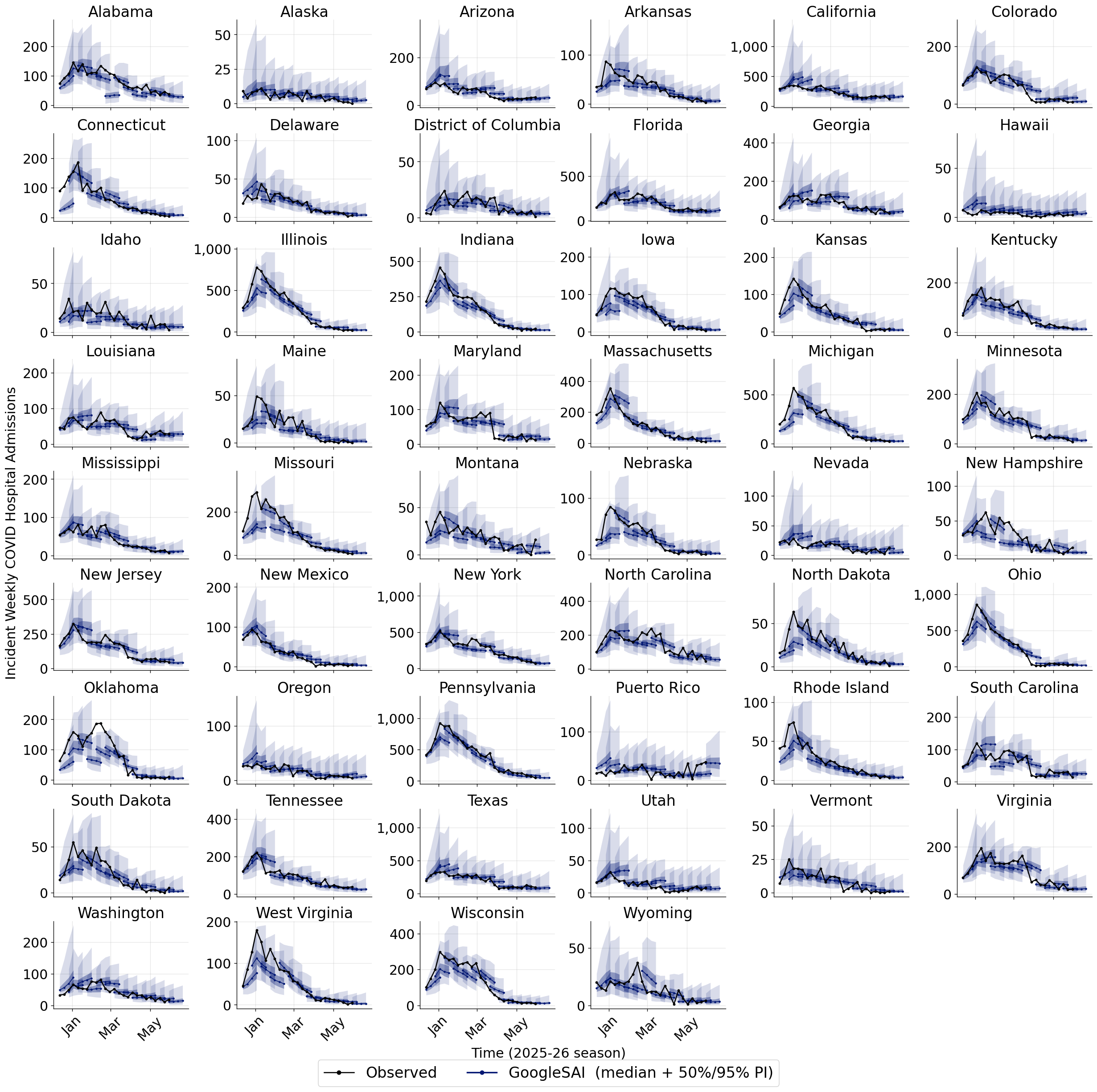}
    \caption{National weekly observed COVID-19 hospitalizations (black) and \gcovidmodel~forecast submissions by jurisdiction over the 2025–26 season with median (dark blue points) and corresponding 50\% and 95\% prediction intervals (blue shaded regions). Only every second set of submitted horizon forecasts is plotted for improved readability. See Fig.~\ref{fig:fig3}a for a larger example forecast plot for California, including all reference dates.}
    \label{fig:covid_forecasts}
\end{figure}

\begin{figure}[H]
    \centering
    \includegraphics[width=.9\linewidth]{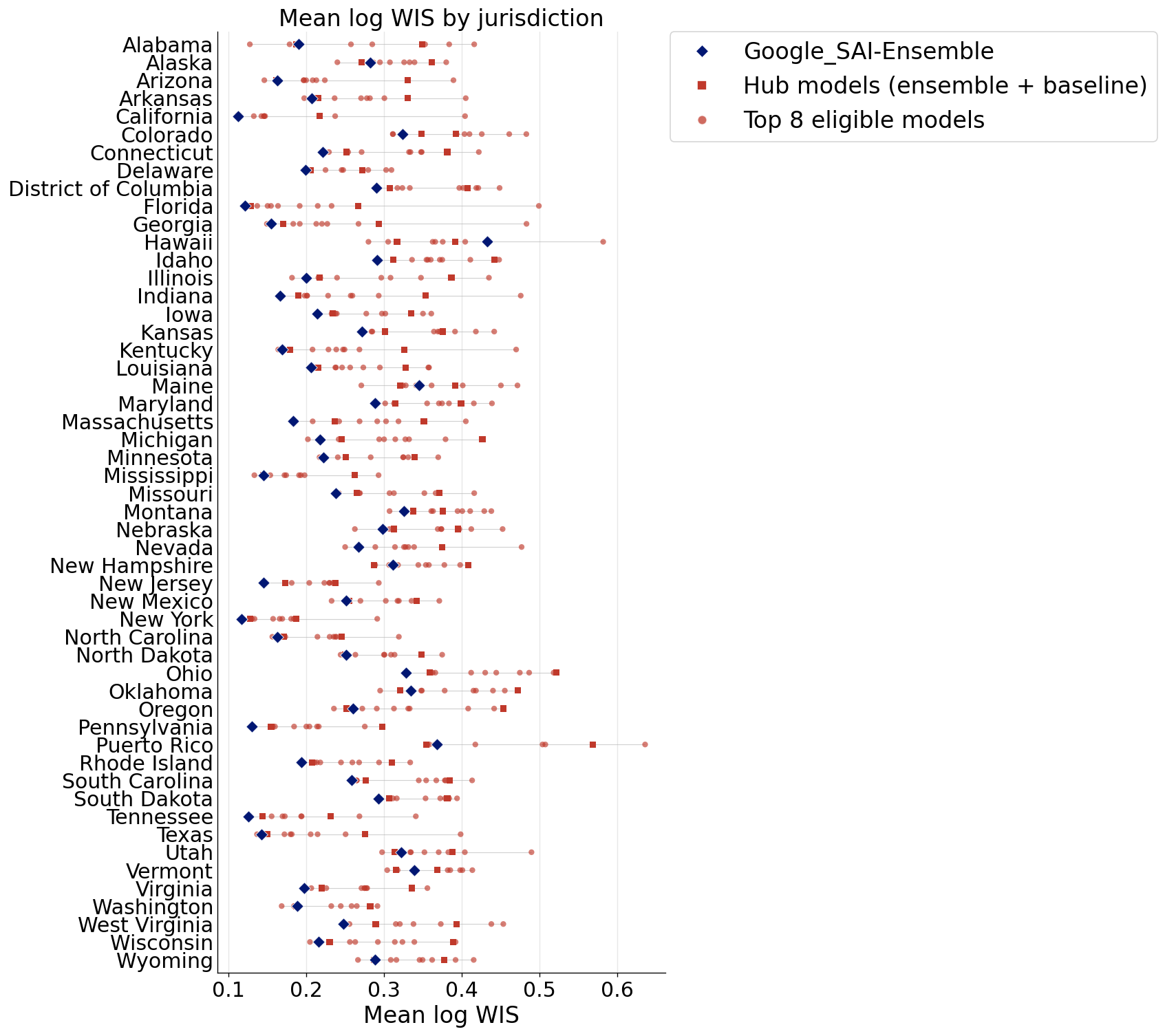}
    \caption{Season-average forecasting skill by jurisdiction for COVIDHub-submitting models. Each row represents a US jurisdiction and each marker a model's mean log WIS for that jurisdiction (lower is better; jurisdictions ordered alphabetically). Red circles show eligible hub-submitting models---those meeting the inclusion criteria ($\geq 80\%$ task coverage); the CDC hub ensemble and baseline are shown as red squares and \gcovidmodel\ as blue diamonds.}
    \label{fig:covid_performace_by_location}
\end{figure}

\begingroup
\small 
\setlength{\tabcolsep}{4pt}
\setlength{\aboverulesep}{0pt}
\setlength{\belowrulesep}{0pt}
\renewcommand{\arraystretch}{1.4}
\begin{longtable}{>{\raggedright\arraybackslash}m{5.0cm}
                  >{\centering\arraybackslash}m{1.2cm}
                  >{\centering\arraybackslash}m{1.5cm}
                  >{\centering\arraybackslash}m{1.5cm}
                  >{\centering\arraybackslash}m{1.5cm}
                  >{\centering\arraybackslash}m{1.5cm}}
    \caption{Mean log WIS by model and prediction horizon for models submitting to the CDC's COVIDHub. Models are ranked by the sum of mean log WIS at horizons 0 and 1 (ascending). 
    Within each horizon column, cells are shaded independently from green (lowest log WIS, best performance) to red (highest log WIS, worst performance); colors are not comparable across columns. 
    The \textit{n tasks} column reports total number of scored forecast tasks submitted across all horizons. Two CFA\_Pyrenew models that do not submit forecasts for all four horizons are included; empty cells indicate horizons for which no forecast was submitted.}
    \label{tab:covid_logwis_by_horizon} \\
    \toprule
    \textbf{Model} & \textbf{n tasks} & \textbf{0} & \textbf{1} & \textbf{2} & \textbf{3} \\
    \midrule
    \endfirsthead
    
    \toprule
    \textbf{Model} & \textbf{n tasks} & \textbf{0} & \textbf{1} & \textbf{2} & \textbf{3} \\
    \midrule
    \endhead
    
    \input{tables/covid_logwis_by_horizon.tex}
\end{longtable}
\endgroup

\begin{figure}[H]
    \centering
    \includegraphics[width=\linewidth]{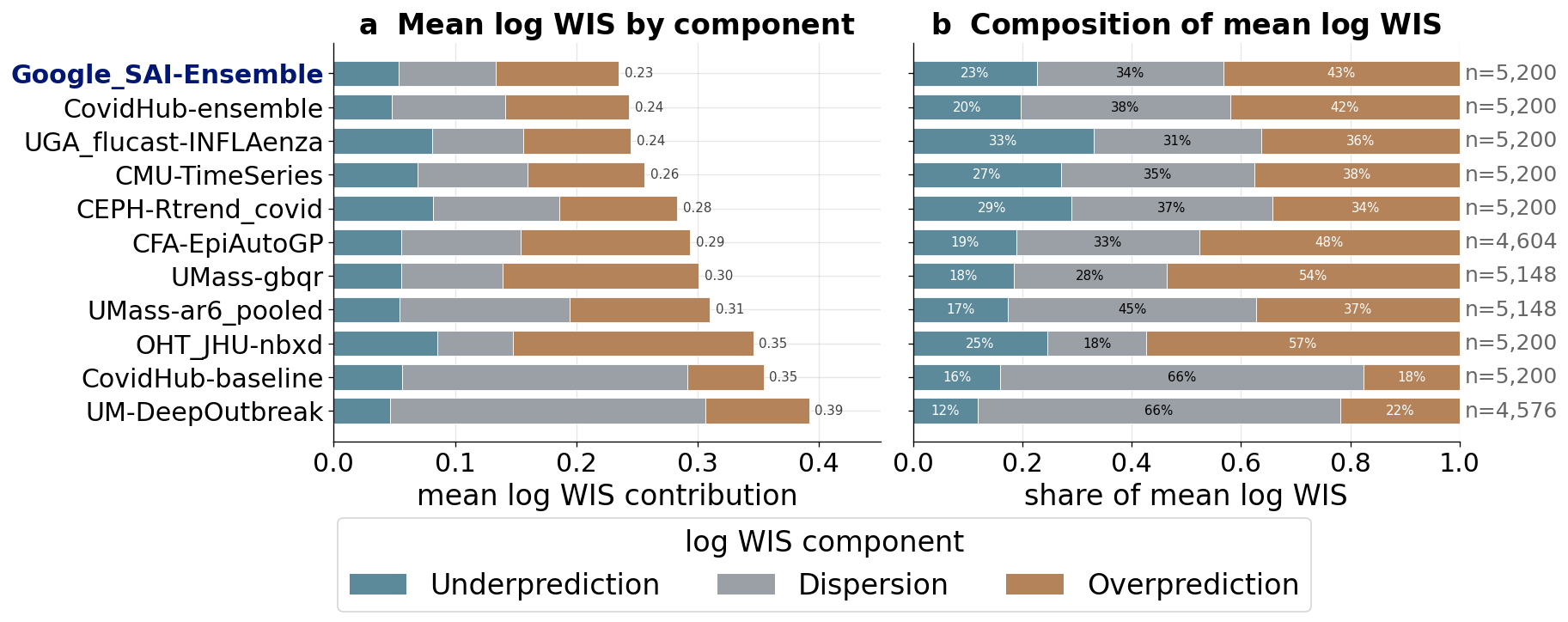}
    \caption{
    Log WIS decomposition into underprediction, dispersion, and overprediction for the top 15 ranked eligible models submitting scorable predictions for $\geq 80\%$ of the task space to the CDC's COVIDHub for the 2025-26 season. Models listed in order of ascending absolute WIS (i.e., not considering number of submitted tasks). \textbf{a.} Absolute WIS contributions for each model. \textbf{b.} WIS contributions normalized to one, with number of tasks submitted by each model on the right.
}
    \label{fig:covid_wis_decomp}
\end{figure}

\begin{figure}[H]
    \centering
    \includegraphics[width=.75\linewidth]{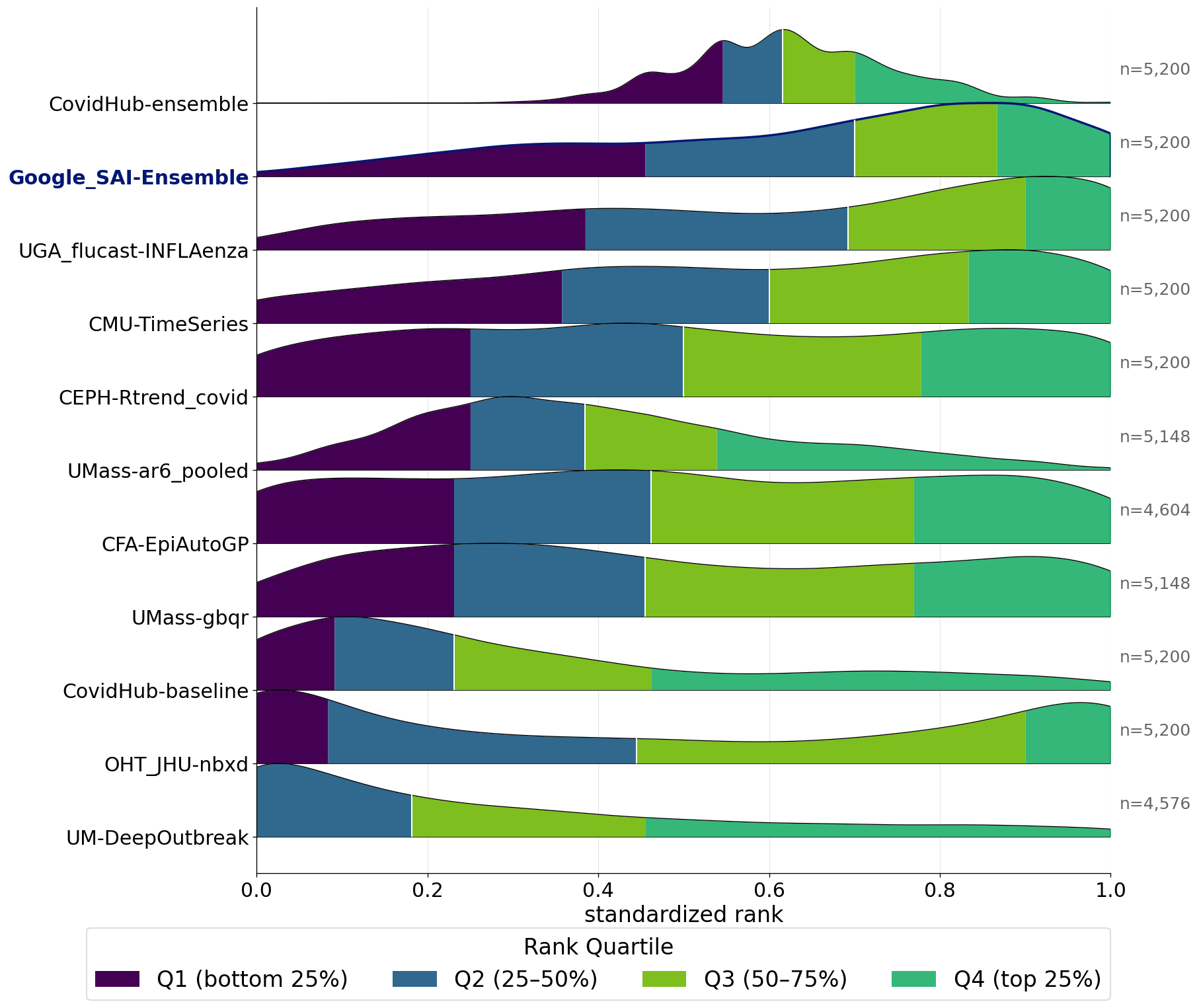}
    \caption{Standardized log WIS rank distributions of top 12 ranked eligible models submitting scorable predictions for $\geq 80\%$ of the task space to the CDC's COVIDHub for the 2025-26 season. A standardized rank of one indicates that the model had the best log WIS for that particular task (location, target, and horizon for that reference date), and a value of zero indicates it had the worst log WIS of submitting models. Density plots show interpolated distributions of standardized ranks achieved by each model for every forecast. 
    Quartiles are colored from purple (bottom quarter, i.e., worst model ranks), through blue and light green to bright green (top quarter, i.e., best model ranks). Medians are represented by vertical white lines. Models are ordered by lowest quartile, with models that rarely had a low rank near the top. \gcovidmodel~is outlined in dark blue.
    }
    \label{fig:covid_rank_distribution}
\end{figure}


\subsection{RSV} \label{SIsec:rsv}

Here we provide analyses for RSV forecast models, analyzed between 3 January and 27 May 2026 (22 reference dates, covering a total of $4,576$ maximum forecast tasks per submitting model). Analyses are restricted to models submitting for $\geq80\%$ of the required task space, i.e., $\geq3,660$ tasks (unless otherwise specified). Under this criterion, four hub-submitting models are eligible for analysis out of a total of seven models which submitted at least one task prediction to RSVHub this season.

\begingroup
\small 
\setlength{\tabcolsep}{4pt}
\setlength{\aboverulesep}{0pt}
\setlength{\belowrulesep}{0pt}
\renewcommand{\arraystretch}{1.4}
\begin{longtable}{>{\raggedright\arraybackslash}m{4.2cm}
                  >{\centering\arraybackslash}m{0.8cm}   
                  >{\centering\arraybackslash}m{1.0cm}   
                  >{\centering\arraybackslash}m{1.0cm}   
                  >{\centering\arraybackslash}m{1.0cm}   
                  >{\centering\arraybackslash}m{1.0cm}   
                  >{\centering\arraybackslash}m{1.0cm}   
                  >{\centering\arraybackslash}m{1.0cm}   
                  >{\centering\arraybackslash}m{1.8cm}}  
    \caption{Summary performance of \ourmethod-generated component models and the top-ranked eligible models submitting to RSVHub. 
    Model names prefixed by \texttt{*} are included in \grsvmodel; the \emph{Type} column labels each \ourmethod\ model as \emph{Scientist-inspired}, \emph{Recombination}, or \emph{AI-led}, while hub models are marked \emph{CDC} (or \emph{CDC ensemble}). 
    Columns report, per model: the number of scored tasks; pairwise relative log WIS and WIS (relative to the RSVHub ensemble $=1$); mean log WIS and mean WIS; mean absolute error of the median forecast (MAE); and empirical coverage of the 95\% prediction interval (95\% Cov.).
    The absolute-error columns (Mean log WIS, Mean WIS, MAE) are shaded independently from green (lowest, best) to red (highest, worst) across models; the pairwise relative columns use a diverging scale centered at one, where values below one (green) indicate better and above one (red) worse performance than the ensemble; 95\% coverage is left unshaded, as its target is ${\approx}95\%$ rather than a minimum or maximum.
    Models are ranked by pairwise relative log WIS (ascending).
    For each pair of models, pairwise relative score ratios are calculated over the intersection of tasks both models submitted, then geometrically averaged across all opponents, so the metric is not affected by differences in the set of tasks each model chose to forecast~\cite{Bracher2021}. 
    Because pairwise relative WIS is computed on the task overlap between each pair of models, which varies across pairs, a model can have a lower mean log WIS than another yet a higher pairwise relative log WIS.}
    \label{tab:rsv_model_summary} \\
    \toprule
        Model & n tasks & Rel. log WIS & Mean log WIS & Rel. WIS & Mean WIS & MAE & 95\% Cov. & Type \\
    \midrule
    \endfirsthead
    \toprule
        Model & n tasks & Rel. log WIS & Mean log WIS & Rel. WIS & Mean WIS & MAE & 95\% Cov. & Type \\
    \midrule
    \endhead
    \input{tables/internal_rsv_crosshub_rel.tex}
\end{longtable}
\endgroup

\begin{figure}[H]
    \centering
    \includegraphics[width=\linewidth]{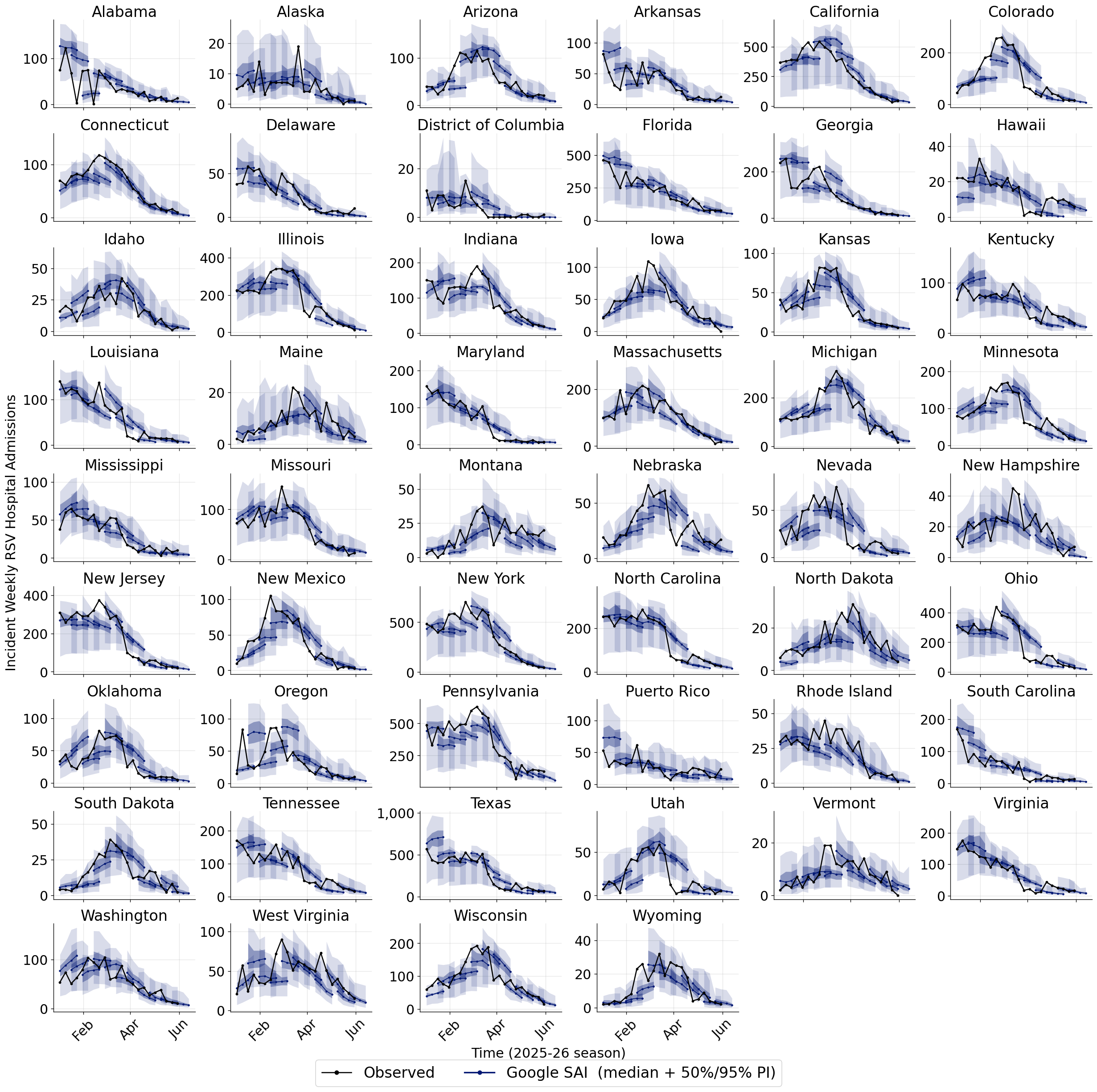}
    \caption{National weekly observed RSV hospitalizations (black) and Google\_SAI-RSVEns forecast submissions by jurisdiction over the 2025–26 season with median (dark blue points) and corresponding 50\% and 95\% prediction intervals (blue shaded regions). Only every second set of submitted horizon forecasts is plotted for improved readability. See Fig.~\ref{fig:fig3}a for a larger example forecast plot for California, including all reference dates.}
    \label{fig:rsv_forecasts}
\end{figure}

\begin{figure}[H]
    \centering
    \includegraphics[width=.9\linewidth]{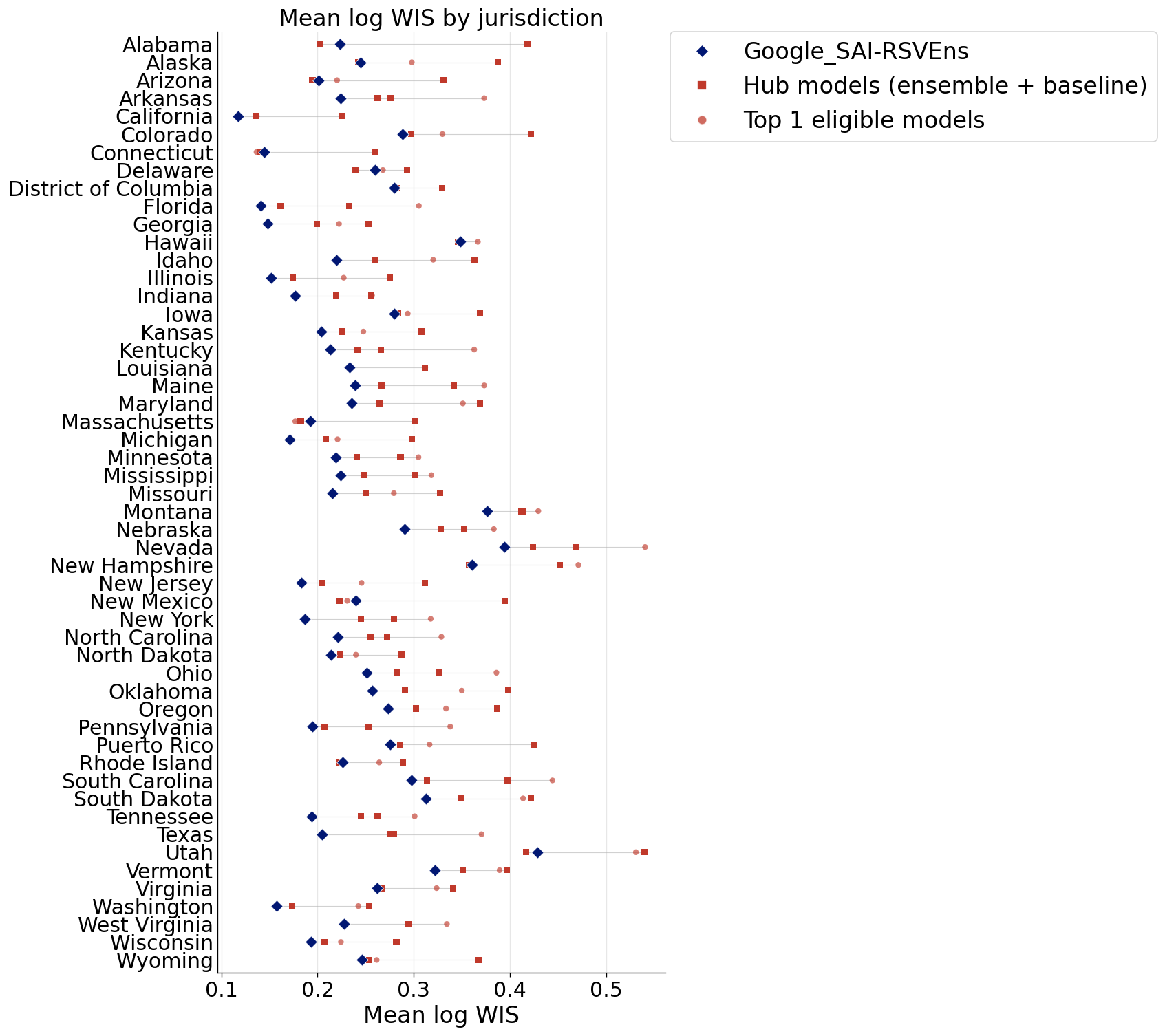}
    \caption{Season-average forecasting skill by jurisdiction for RSVHub-submitting models. Each row represents a US jurisdiction and each marker a model's mean log WIS for that jurisdiction (lower is better; jurisdictions ordered alphabetically). Red circles show eligible hub-submitting models---those meeting the inclusion criteria ($\geq 80\%$ task coverage); the CDC hub ensemble and baseline are shown as red squares and \grsvmodel\ as blue diamonds.}

    \label{fig:rsv_performace_by_location}
\end{figure}

\begingroup
\small 
\setlength{\tabcolsep}{4pt}
\setlength{\aboverulesep}{0pt}
\setlength{\belowrulesep}{0pt}
\renewcommand{\arraystretch}{1.4}
\begin{longtable}{>{\raggedright\arraybackslash}m{5.0cm}
                  >{\centering\arraybackslash}m{1.2cm}
                  >{\centering\arraybackslash}m{1.5cm}
                  >{\centering\arraybackslash}m{1.5cm}
                  >{\centering\arraybackslash}m{1.5cm}
                  >{\centering\arraybackslash}m{1.5cm}}
    \caption{Mean log WIS by model and prediction horizon for models submitting to the CDC’s RSVHub. Models are ranked by the sum of mean log WIS at horizons 0 and 1 (ascending). Within each horizon column, cells are shaded independently from green (lowest log WIS, best performance) to red (highest log WIS, worst performance); colors are not comparable across columns. The \textit{n tasks} column reports total number of scored forecast tasks submitted across all horizons. Two \texttt{CFA\_Pyrenew} models that do not submit forecasts for all four horizons are included; empty cells indicate horizons for which no forecast was submitted.}
    \label{tab:rsv_logwis_by_horizon} \\
    \toprule
    \textbf{Model} & \textbf{n tasks} & \textbf{0} & \textbf{1} & \textbf{2} & \textbf{3} \\
    \midrule
    \endfirsthead
    
    \toprule
    \textbf{Model} & \textbf{n tasks} & \textbf{0} & \textbf{1} & \textbf{2} & \textbf{3} \\
    \midrule
    \endhead
    
    \input{tables/rsv_logwis_by_horizon.tex}
\end{longtable}
\endgroup

\begin{figure}[H]
    \centering
    \includegraphics[width=\linewidth]{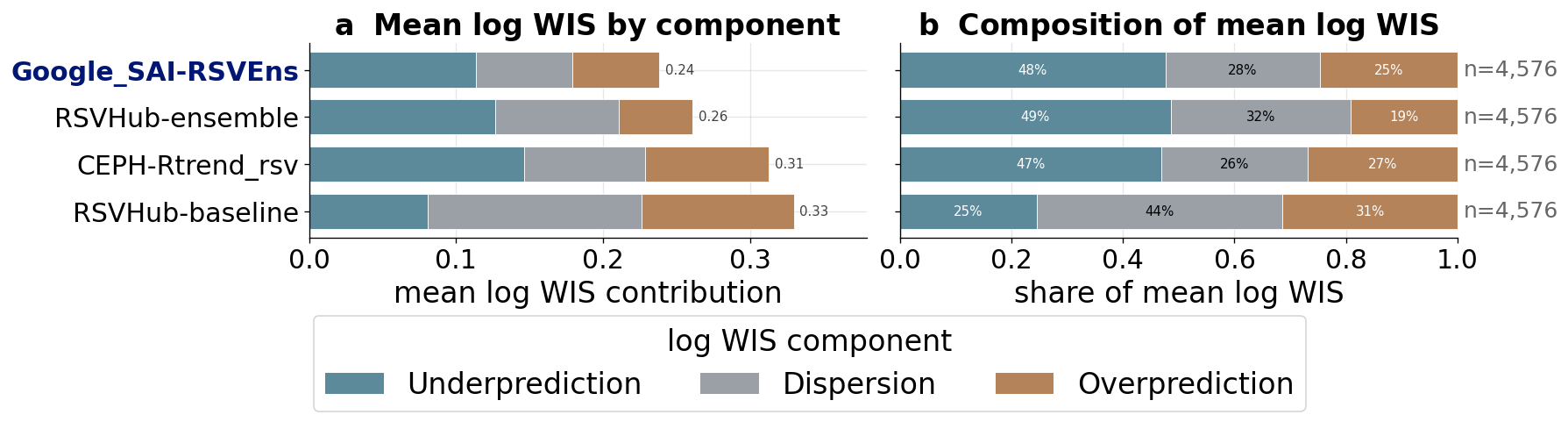}
    \caption{Log WIS decomposition into underprediction, dispersion, and overprediction for the top 15 ranked eligible models submitting scorable predictions for $\geq 80\%$ of the task space to the CDC's RSVHub for the 2025-26 season. Models listed in order of ascending absolute WIS (i.e., not considering number of submitted tasks). \textbf{a.} Absolute WIS contributions for each model. \textbf{b.} WIS contributions normalized to one, with number of tasks submitted by each model on the right.
}
    \label{fig:rsv_wis_decomp}
\end{figure}

\begin{figure}[H]
    \centering
    \includegraphics[width=.75\linewidth]{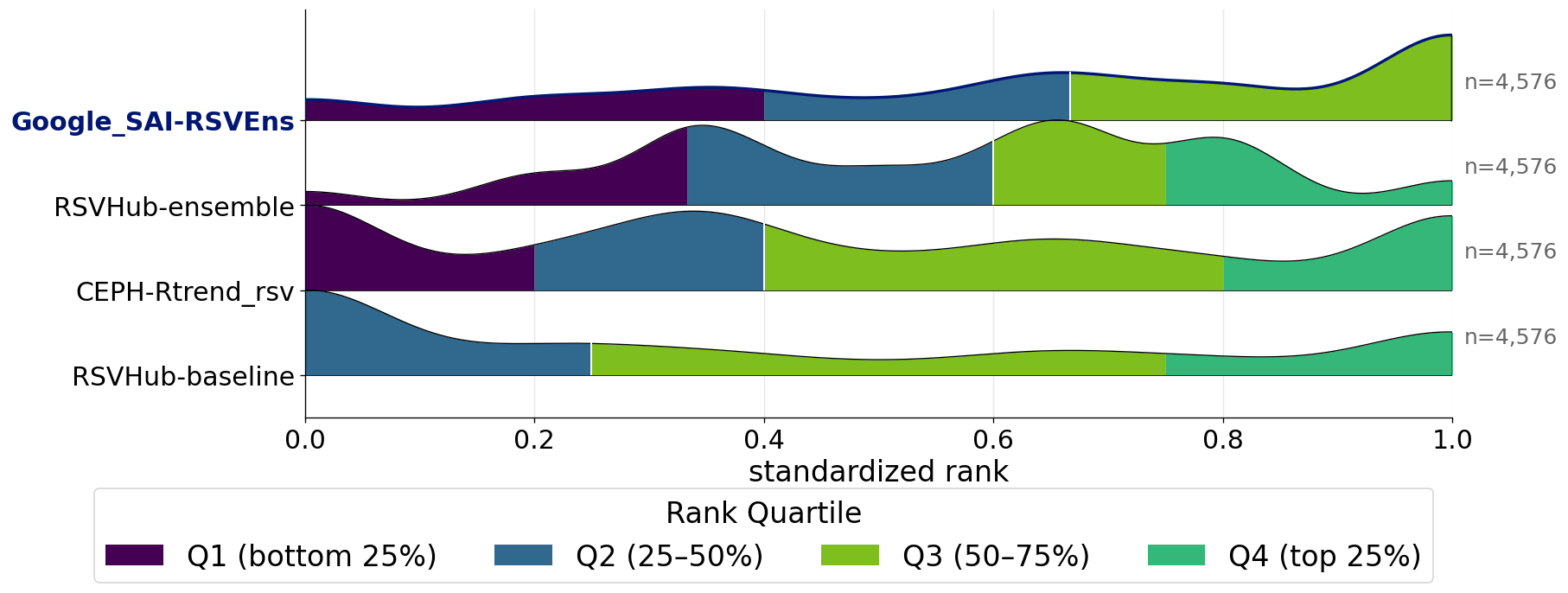}
    \caption{Standardized log WIS rank distributions of ranked eligible models submitting scorable predictions for $\geq80\%$ of the task space to the CDC's RSVHub for the 2025-26 season. A standardized rank of one indicates that the model had the best log WIS for that particular task (location, target, and horizon for that reference date), and a value of zero indicates it had the worst log WIS of submitting models. Density plots show interpolated distributions of standardized ranks achieved by each model for every forecast. Quartiles are colored from purple (bottom quarter, i.e., worst model ranks), through blue and light green to bright green (top quarter, i.e., best model ranks). Medians are represented by vertical white lines. Models are ordered by lowest quartile, with models that rarely had a low rank near the top. \grsvmodel~is outlined in dark blue.
    Note that, with only four models considered, an invisible Q4 (\grsvmodel) implies that in more than 25\% of cases, the model ranked first, while an invisible Q1 (\texttt{RSVHub-baseline}) implies that the model had zero rank in more than $25\%$ of considered cases.}
    \label{fig:rsv_rank_distribution}
\end{figure}


\newpage
\section{Retrospective Optimization Metric Exploration Extended Results}

\begin{figure}[htbp]
    \centering
    \includegraphics[width=.85\textwidth]{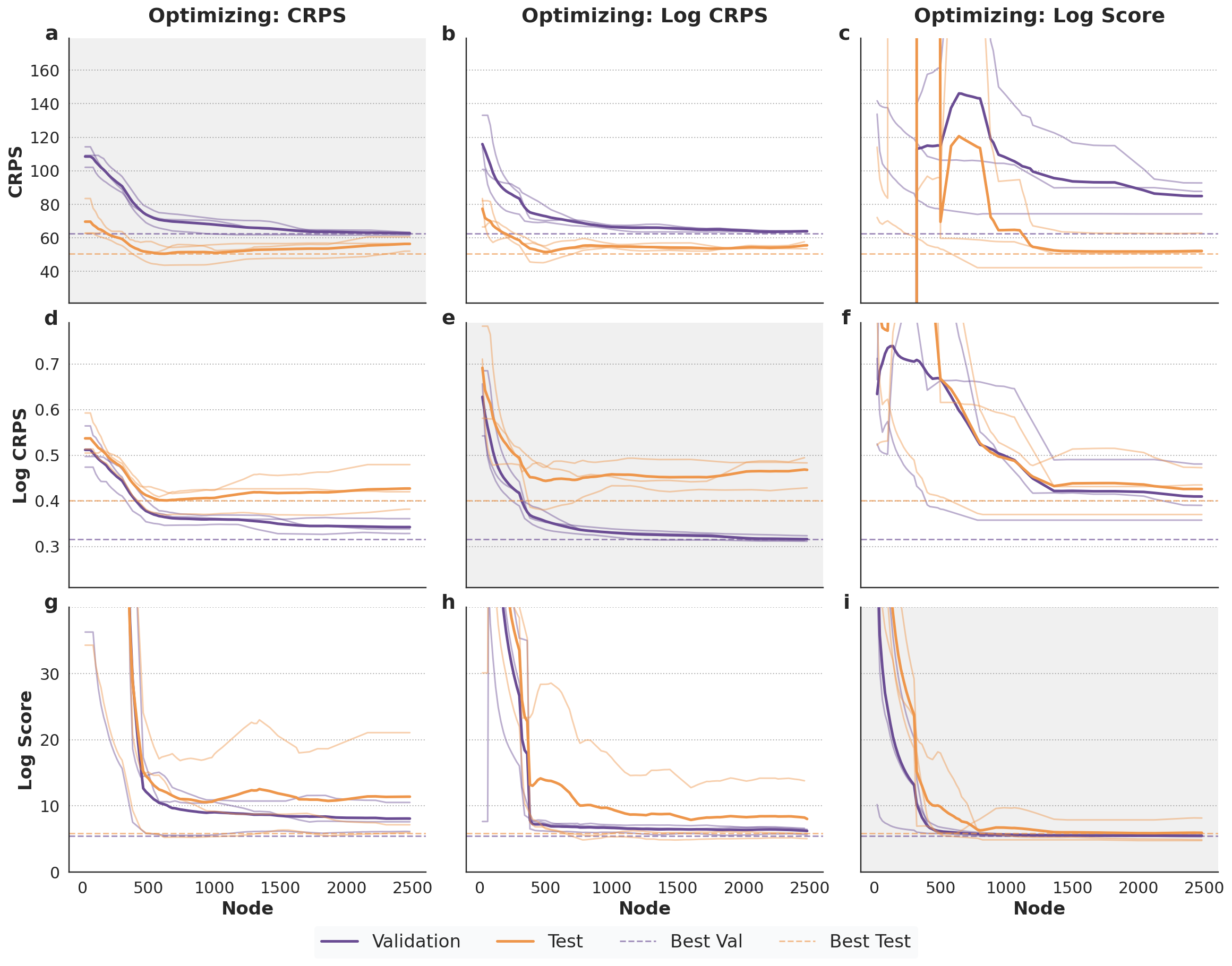}
    \caption{Influence of Hill-climbing Metrics on Search Stability and Generalization. Each column represents a set of \ourmethod experiments with the objective function one of three metrics: CRPS, log-scale CRPS and Log Score. Five experiments were conducted for each metric (faint lines) and, for each experiment, resulting forecasts were scored using each of the three metrics on both an in-sample validation set (purple lines) and an out-of-sample test set (orange lines). Each line tracks the cumulative best score of the \ourmethod search across 2,500 nodes ($x$-axis). The mean values across experiments are highlighted (bold lines). Dashed horizontal lines represent the best mean score across all optimization experiments for that specific metric. The mean validation score lines are monotonically decreasing when the optimized score and the evaluation metric agree because we track the cumulative best-scoring node. However, the test score of that same node need not decrease monotonically, and a rising test curve concurrent with a still-falling validation curve is a diagnostic signal of overfitting to the validation period. The persistent vertical gap between validation and test lines reflects a difference in prediction difficulty between the two seasons rather than necessarily a modeling failure.}
    \label{fig:metrics_in_search}
\end{figure}

\begin{table}[htbp]
  \small 
  \setlength{\tabcolsep}{3pt} 
  \centering
  \caption{Forecast quality statistics for models discovered under each optimization metric, evaluated on the retrospective influenza test set (Fig.~\ref{fig:fig1}) across all jurisdictions. Values are means $\pm$ standard deviations across five independent search replicas. \textbf{Mean Bias}: average difference between predicted median and observed admissions (negative $\implies$ underprediction).
  \textbf{Prop. Under/Over}: proportion of predictions where the median falls below or above the observed value. \textbf{Mean 50\% PI Width}: average width of the 50\% prediction interval.}
  \label{tab:forecast_metrics_extended_si}
  \begin{tabular}{
    l
    S[table-format=-2.1] @{\,$\pm$\,} S[table-format=2.1] 
    S[table-format=1.2] @{\,$\pm$\,} S[table-format=1.2] 
    S[table-format=1.2] @{\,$\pm$\,} S[table-format=1.2] 
    S[table-format=3.1] @{\,$\pm$\,} S[table-format=2.1] 
    S[table-format=2.1] @{\,$\pm$\,} S[table-format=2.1] 
    S[table-format=1.2] @{\,$\pm$\,} S[table-format=1.2] 
    S[table-format=2.1] @{\,$\pm$\,} S[table-format=1.1] 
  }
    \toprule
    {\thead{\textbf{Optimization} \\ \textbf{Metric}}} & 
    \multicolumn{2}{c}{\thead{\textbf{Mean} \\ \textbf{Bias}}} & 
    \multicolumn{2}{c}{\thead{\textbf{Prop.} \\ \textbf{Under}}} & 
    \multicolumn{2}{c}{\thead{\textbf{Prop.} \\ \textbf{Over}}} & 
    \multicolumn{2}{c}{\thead{\textbf{50\%~PI} \\ \textbf{Width}}} & 
    \multicolumn{2}{c}{\thead{\textbf{CRPS}}} & 
    \multicolumn{2}{c}{\thead{\textbf{Log} \\ \textbf{CRPS}}} & 
    \multicolumn{2}{c}{\thead{\textbf{Log} \\ \textbf{Score}}} \\
    \midrule
    \textbf{CRPS}           & \textbf{-38.0} &  8.6 & \textbf{0.55} & 0.02 & \textbf{0.43} & 0.03 & \textbf{136.8} & 31.5 & \textbf{45.8} &  2.9 & 0.41 & 0.03 & 16.4 & 8.9 \\
    \textbf{Log-scale CRPS} & -39.0 &  6.8 & 0.56 & 0.03 & 0.42 & 0.03 & 149.0 & 18.5 & 47.5 &  3.4 & \textbf{0.39} & 0.03 &  9.9 & 3.0 \\
    \textbf{Log score}      & -93.2 & 84.8 & 0.66 & 0.18 & 0.32 & 0.18 & 188.8 & 36.0 & 60.7 & 10.8 & 0.57 & 0.18 &  \textbf{5.1} & 0.4 \\
    \bottomrule
  \end{tabular}
\end{table}

\newpage
\section{Retrospective Fidelity Extended Results and Judge Prompts}

\subsection*{LLM Judge Instructions} \label{SIsec:judge_prompts}

\begin{tcolorbox}[
  colback=white,
  colframe=black!30,
  coltitle=black,
  colbacktitle=black!10,
  fonttitle=\ttfamily\small,
  title=In-loop LLM Judge Prompt,
  breakable,
  enhanced,
  left=6pt, right=6pt, top=4pt, bottom=4pt,
]
\small

You are evaluating whether a forecasting implementation follows the core principles of a specified method.

METHOD INSTRUCTION: \texttt{\{method\_instruction\}}

IMPLEMENTATION CODE: \texttt{\{code\_str\}}

Evaluate whether the implementation:
\begin{itemize}
    \item Follows the core methodology described in the instruction.
    \item Maintains the essential principles of the method (even if making improvements).
    \item Doesn't deviate into a completely different approach.
\end{itemize}
Consider that some practical adaptations may be necessary (e.g., using provided dataframes instead of external APIs), but the algorithmic approach and core principles should be preserved.

Respond with:

SCORE: A number from 0-10 (10 = perfectly follows method, 0 = completely different approach)

Format: SCORE: X
\end{tcolorbox}

\begin{table}[htbp]
  \centering
  \caption{Summary of search performance and instruction-following fidelity for models generated retrospectively. The \textit{Judge} column indicates whether the automated in-loop judge was active during the search. For each run, the final model was selected according to the selection score defined in Equation~\eqref{eq:selection_score}. Instruction-following outcomes are reported as counts of Match (M), Partial Match (PM), and No Match (NM) across three independent runs. Mean selection score (SS) is reported as mean $\pm$ std for the best node, selected by the same metric. Lower values indicate better performance on validation and test periods.}
  \label{tab:llm_summary_updated}
  \setlength{\tabcolsep}{4pt}
    \begin{tabular}{
      lllc
      S[table-format=3.2]
      @{\,${}\pm{}$\,}
      S[table-format=3.2]
    }
    \toprule
    \textbf{Prompted Method}
    & \textbf{Gemini Version}
    & \textbf{Judge}
    & \textbf{Outcomes (M/PM/NM)}
    & \multicolumn{2}{c}{\textbf{SS (mean $\pm$ std)}} \\
    \midrule

    \multirow{4}{*}{\texttt{Cornell\_JHU-hierarchSIR}}
    & 2.5 Flash & No  & 0 / 3 / 0 & 179.09 & 11.53  \\
    & 2.5 Flash & Yes & 1 / 1 / 1 & 411.97 & 207.36 \\
    & 3 Flash   & No  & 1 / 2 / 0 & 158.60 & 15.33  \\
    & 3 Pro     & No  & 3 / 0 / 0 & 140.62 & 11.93  \\
    \midrule
    
    \multirow{4}{*}{\texttt{NU-PGF\_FLUH}}
    & 2.5 Flash & No  & 0 / 3 / 0 & 236.06 & 83.69 \\
    & 2.5 Flash & Yes & 0 / 3 / 0 & 205.11 & 82.40 \\
    & 3 Flash   & No  & 0 / 3 / 0 & 153.97 & 5.22  \\
    & 3 Pro     & No  & 0 / 3 / 0 & 140.09 & 12.76 \\
    \midrule
    
    \multirow{4}{*}{\texttt{UMass-gbqr}}
    & 2.5 Flash & No  & 3 / 0 / 0 & 291.38 & 208.43 \\
    & 2.5 Flash & Yes & 3 / 0 / 0 & 167.62 & 17.72  \\
    & 3 Flash   & No  & 2 / 1 / 0 & 166.95 & 8.91   \\
    & 3 Pro     & No  & 2 / 1 / 0 & 155.08 & 5.47   \\
    \midrule
    
    \multirow{4}{*}{\texttt{UGA\_flucast-INFLAenza}}
    & 2.5 Flash & No  & 1 / 2 / 0 & 316.16 & 110.19 \\
    & 2.5 Flash & Yes & 3 / 0 / 0 & 196.51 & 37.95  \\
    & 3 Flash   & No  & 1 / 2 / 0 & 176.55 & 4.53   \\
    & 3 Pro     & No  & 0 / 3 / 0 & 137.78 & 18.58  \\
    \bottomrule
    \end{tabular}
\end{table}

\begin{table}[htbp]
\small
  \centering
  \caption{Comparison of Gemini 2.5 Flash performance with and without the in-loop judge. The \textit{Score} column reports the selection score~\eqref{eq:selection_score} over the validation and test periods. \textit{Judgment} refers to a post-hoc specialized LLM evaluation of whether the output code correctly follows instructions (`Match'), partially follows instructions (`Partial Match') or does not follow given instructions (`No Match').}
  \setlength{\tabcolsep}{3pt}

  \begin{tabular}{
    l |
    S[table-format=4.0]
    S[table-format=4.2]
    S[table-format=3.2]
    l |
    S[table-format=4.0]
    S[table-format=4.2]
    S[table-format=3.2]
    l
  }
    \toprule
    \multirow{2}{*}{\textbf{Prompted Method}}
    & \multicolumn{4}{c|}{\textbf{No Judge}}
    & \multicolumn{4}{c}{\textbf{In-loop Judge}} \\

    & {\textbf{Nodes}} & {\textbf{Runtime}} & {\textbf{Score}} & \textbf{Judgment}
    & {\textbf{Nodes}} & {\textbf{Runtime}} & {\textbf{Score}} & \textbf{Judgment} \\
    \midrule
    \midrule

    \multirow{3}{*}{\texttt{Cornell\_JHU-hierarchSIR}}
    & 927  & 2488.21 & 176.14 & Partial Match
    & 2031 & 2511.67 & 172.54 & No Match \\
    & 1017 & 2496.71 & 169.33 & Partial Match
    & 701  & 2499.10 & 531.69 & Match \\
    & 341  & 2495.12 & 191.81 & Partial Match
    & 357  & 2489.63 & 531.69 & Partial Match \\
    \midrule

    \multirow{3}{*}{\texttt{NU-PGF\_FLUH}}
    & 2500 & 219.94 & 332.31 & Partial Match
    & 1742 & 2499.13 & 166.94 & Partial Match \\
    & 2500 & 187.12 & 195.42 & Partial Match
    & 872  & 2497.58 & 148.72 & Partial Match \\
    & 2500 & 254.04 & 180.44 & Partial Match
    & 276  & 2494.35 & 299.68 & Partial Match \\
    \midrule

    \multirow{3}{*}{\texttt{UGA\_flucast-INFLAenza}}
    & 2498 & 120.64 & 207.47 & Partial Match
    & 2500 & 1020.40 & 185.45 & Match \\
    & 2500 & 105.51 & 427.79 & Partial Match
    & 2500 & 1172.44 & 238.76 & Match \\
    & 2500 & 79.68  & 313.21 & Match
    & 2500 & 989.20  & 165.32 & Match \\
    \midrule

    \multirow{3}{*}{\texttt{UMass-gbqr}}
    & 912  & 2492.3  & 183.02 & Match
    & 427  & 2491.09 & 147.17 & Match \\
    & 1121 & 2499.94 & 159.44 & Match
    & 467  & 2493.92 & 177.14 & Match \\
    & 2500 & 62.93   & 531.67 & Match
    & 632  & 2494.53 & 178.55 & Match \\

    \bottomrule
  \end{tabular}
  \label{tab:judge_comparison}
\end{table}

\begin{table}[htbp]
\small
  \centering
  \caption{Comparison of instruction-following capability and performance of `best-node' model codes when \ourmethod is used with different Gemini tiers.
  Selection score \textit{SS} \eqref{eq:selection_score} is reported for all runs, lower is better. \textit{Judgment} refers to a post-hoc specialized LLM evaluation of whether the output code correctly follows instructions (`Match'), partially follows instructions (`Partial Match') or does not follow given instructions (`No Match').}

  \begin{tabular}{
    l
    l
    S[table-format=4.0]
    S[table-format=4.2]
    S[table-format=3.2]
    l
  }
    \toprule
    \textbf{Prompted Method}
    & \textbf{Model}
    & {\textbf{Nodes}}
    & {\textbf{Runtime (hr)}}
    & {\textbf{SS}}
    & \textbf{Judgment} \\
    \midrule

    \multirow{9}{*}{\texttt{Cornell\_JHU-hierarchSIR}}
    & gemini-2.5-flash & 927 & 2488.21 & 176.14 & Partial Match \\
    & & 341 & 2495.12 & 191.81 & Partial Match \\
    & & 1017 & 2496.71 & 169.33 & Partial Match \\
    & gemini-3-flash-preview & 1267 & 2498.22 & 148.98 & Partial Match \\
    & & 747 & 2494.13 & 176.27 & Partial Match \\
    & & 1440 & 2499.55 & 150.54 & Match \\
    & gemini-3-pro-preview & 344 & 2497.31 & 138.28 & Match \\
    & & 394 & 2499.68 & 153.54 & Match \\
    & & 446 & 2490.9 & 130.03 & Match \\
    \midrule

    \multirow{9}{*}{\texttt{NU-PGF\_FLUH}}
    & gemini-2.5-flash & 2500 & 219.94 & 332.31 & Partial Match \\
    & & 2500 & 187.12 & 195.42 & Partial Match \\
    & & 2500 & 254.04 & 180.44 & Partial Match \\
    & gemini-3-flash-preview & 2500 & 106.35 & 150.59 & Partial Match \\
    & & 2500 & 125.49 & 159.98 & Partial Match \\
    & & 2500 & 120.35 & 151.35 & Partial Match \\
    & gemini-3-pro-preview & 2500 & 280.73 & 126.65 & Partial Match \\
    & & 2500 & 1930.41 & 141.59 & Partial Match \\
    & & 2500 & 1478.84 & 152.04 & Partial Match \\
    \midrule

    \multirow{9}{*}{\texttt{UGA\_flucast-INFLAenza}}
    & gemini-2.5-flash & 2498 & 120.64 & 207.47 & Partial Match \\
    & & 2500 & 105.51 & 427.79 & Partial Match \\
    & & 2500 & 79.68 & 313.21 & Match \\
    & gemini-3-flash-preview & 2458 & 89.8 & 181.22 & Partial Match \\
    & & 2370 & 73.82 & 172.17 & Partial Match \\
    & & 2386 & 114.83 & 176.26 & Match \\
    & gemini-3-pro-preview & 2500 & 156.67 & 145 & Partial Match \\
    & & 2500 & 115.5 & 151.67 & Partial Match \\
    & & 956 & 2493.22 & 116.67 & Partial Match \\
    \midrule

    \multirow{9}{*}{\texttt{UMass-gbqr}}
    & gemini-2.5-flash & 912 & 2492.3 & 183.02 & Match \\
    & & 1121 & 2499.94 & 159.44 & Match \\
    & & 2500 & 62.93 & 531.67 & Match \\
    & gemini-3-flash-preview & 851 & 2498.52 & 159.9 & Match \\
    & & 997 & 2497.47 & 176.96 & Match \\
    & & 790 & 2498.76 & 164 & Partial Match \\
    & gemini-3-pro-preview & 358 & 2496.17 & 149.91 & Partial Match \\
    & & 442 & 2497.78 & 160.8 & Match \\
    & & 353 & 2495.51 & 154.53 & Match \\
    \bottomrule
  \end{tabular}
  \label{tab:llm_comparison}
\end{table}

\newpage
\section{Epiforge Reporting Guidelines}

\begin{longtable}{
    >{\centering\arraybackslash}p{0.4cm}
    >{\raggedright\arraybackslash}p{8.0cm}
    >{\raggedright\arraybackslash}p{6.2cm}
}
\caption{\textbf{EPIFORGE 2020 reporting checklist.}
Recommended reporting items for epidemic forecasting and prediction research, following Pollett et al.~\cite{pollett_epiforge_2021}.}
\label{tab:epiforge} \\

\toprule
\textbf{\#} & \textbf{Checklist item} & \textbf{Location in manuscript} \\
\midrule
\endfirsthead

\toprule
\textbf{\#} & \textbf{Checklist item} & \textbf{Location in manuscript} \\
\midrule
\endhead

1 &
Describe the study as forecast or prediction research in at least the title or abstract. &
Title; Abstract. \\

2 &
Define purpose of study and forecasting targets. &
Introduction; Methods (Section~\ref{sec:data_forecasting_task}); Fig.~\ref{fig:fig1}a--c. \\

3 &
Fully document the methods. &
Methods (Sections~\ref{sec:data_forecasting_task}--\ref{sec:retrospective_experiments}); Figs.~\ref{fig:fig1}--\ref{fig:fig2}; Supplementary Fig.~\ref{fig:selection_diagram}. \\

4 &
Identify whether forecasts were performed prospectively, in real time, and/or retrospectively. &
Methods (Sections~\ref{sec:data_forecasting_task}, \ref{sec:prospective_implementation}, and~\ref{sec:retrospective_experiments}); Results. \\

5 &
Explicitly describe the origin of input source data, with references. &
Methods (Sections~\ref{sec:data_forecasting_task} and~\ref{sec:pathogen_implementations}); Data Availability. \\

6 &
Provide source data with publication, or document reasons why this was not possible. &
Data Availability. \\

7 &
Describe input data processing procedures. &
Methods (Sections~\ref{sec:data_forecasting_task} and~\ref{sec:pathogen_implementations}). \\

8 &
State and describe the model type, document model assumptions, including references. &
Methods (Sections~\ref{sec:method_prompt_sources} and~\ref{sec:pathogen_implementations}); Fig.~\ref{fig:fig4}; Supplementary Sections~\ref{SIsec:flu}, \ref{SIsec:covid}, and~\ref{SIsec:rsv}. \\

9 &
Make the model code available. &
Code Availability. \\

10 &
Describe model validation and justify the approach. &
Methods (Sections~\ref{sec:data_forecasting_task} and~\ref{sec:model_selection}); Fig.~\ref{fig:fig1}d; Supplementary Table~\ref{tab:dates}. \\

11 &
Describe the forecast accuracy evaluation method used, with justification. &
Methods (Section~\ref{sec:evaluation_metrics}); Eqs.~\eqref{eq:wis}--\eqref{eq:interval_score}. \\

12 &
Where possible, compare model results to a benchmark or other comparator model, with justification of comparator choice. &
Methods (Section~\ref{sec:evaluation_metrics}); Results; Figs.~\ref{fig:fig3}b and~\ref{fig:fig4}. \\

13 &
Describe the forecast horizon, with justification of its length. &
Methods (Section~\ref{sec:data_forecasting_task}); Fig.~\ref{fig:fig1}a--c. \\

14 &
Present and explain uncertainty of forecasting results. &
Methods (Sections~\ref{sec:data_forecasting_task} and~\ref{sec:evaluation_metrics}); Fig.~\ref{fig:fig3}a; Supplementary Figs.~\ref{fig:flu_forecasts}, \ref{fig:covid_forecasts}, and~\ref{fig:rsv_forecasts}. \\

15 &
Briefly summarize the results in nontechnical terms, including a nontechnical interpretation of forecast uncertainty. &
Abstract; Results; Discussion. \\

16 &
If results are published as a data object, encourage a time-stamped version number. &
Results; Code Availability. \\

17 &
Describe weaknesses of the forecast, including weaknesses specific to data quality and methods. &
Discussion; Methods (Sections~\ref{sec:data_forecasting_task} and~\ref{sec:evaluation_metrics}). \\

18 &
Comment on potential implications and impact for public health action and decision-making. &
Introduction; Discussion. \\

19 &
Comment on generalizability across populations. &
Discussion. \\

\bottomrule
\end{longtable}

%% file: tables/internal_flu_crosshub_rel.tex
\color[HTML]{011773} \bfseries \texttt{Google\_\allowbreak{}SAI-\allowbreak{}FluEns} & 5824 & {\cellcolor[HTML]{A8ECC5}} \color[HTML]{000000} 0.93 & {\cellcolor[HTML]{0B7D42}} \color[HTML]{F1F1F1} 0.297 & {\cellcolor[HTML]{BEF1D4}} \color[HTML]{000000} 0.94 & {\cellcolor[HTML]{B3DF72}} \color[HTML]{000000} 58.6 & {\cellcolor[HTML]{B7E075}} \color[HTML]{000000} 89.9 & 94\% & CDC &  \\
\texttt{UGA\_\allowbreak{}flucast-\allowbreak{}INFLAenza} & 5408 & {\cellcolor[HTML]{BEF1D4}} \color[HTML]{000000} 0.95 & {\cellcolor[HTML]{138C4A}} \color[HTML]{F1F1F1} 0.308 & {\cellcolor[HTML]{99E9BC}} \color[HTML]{000000} 0.92 & {\cellcolor[HTML]{8CCD67}} \color[HTML]{000000} 51.2 & {\cellcolor[HTML]{84CA66}} \color[HTML]{000000} 75.6 & 90\% & CDC &  \\
\texttt{FluSight-\allowbreak{}HJudge\_\allowbreak{}ensemble} & 5824 & {\cellcolor[HTML]{C9F3DC}} \color[HTML]{000000} 0.96 & {\cellcolor[HTML]{138C4A}} \color[HTML]{F1F1F1} 0.309 & {\cellcolor[HTML]{C9F3DC}} \color[HTML]{000000} 0.95 & {\cellcolor[HTML]{BBE278}} \color[HTML]{000000} 60.2 & {\cellcolor[HTML]{BBE278}} \color[HTML]{000000} 91.6 & 91\% & CDC &  \\
\texttt{CMU-\allowbreak{}TimeSeries} & 5616 & {\cellcolor[HTML]{CDF4DE}} \color[HTML]{000000} 0.96 & {\cellcolor[HTML]{118848}} \color[HTML]{F1F1F1} 0.306 & {\cellcolor[HTML]{65DD99}} \color[HTML]{000000} 0.88 & {\cellcolor[HTML]{AFDD70}} \color[HTML]{000000} 57.7 & {\cellcolor[HTML]{B3DF72}} \color[HTML]{000000} 89.0 & 94\% & CDC &  \\
\texttt{FluSight-\allowbreak{}trained\_\allowbreak{}mean} & 5824 & {\cellcolor[HTML]{D8F6E5}} \color[HTML]{000000} 0.97 & {\cellcolor[HTML]{15904C}} \color[HTML]{F1F1F1} 0.311 & {\cellcolor[HTML]{D8F6E5}} \color[HTML]{000000} 0.97 & {\cellcolor[HTML]{BFE47A}} \color[HTML]{000000} 61.0 & {\cellcolor[HTML]{BBE278}} \color[HTML]{000000} 91.7 & 90\% & CDC &  \\

%% file: tables/flu_logwis_by_horizon.tex
\texttt{CMU-\allowbreak{}TimeSeries} & 5,616 & {\cellcolor[HTML]{026C39}} \color[HTML]{F1F1F1} 0.2017 & {\cellcolor[HTML]{07753E}} \color[HTML]{F1F1F1} 0.2786 & {\cellcolor[HTML]{118848}} \color[HTML]{000000} 0.3429 & {\cellcolor[HTML]{3CA959}} \color[HTML]{000000} 0.4004 \\
\texttt{UGA\_\allowbreak{}flucast-\allowbreak{}INFLAenza} & 5,408 & {\cellcolor[HTML]{006837}} \color[HTML]{F1F1F1} 0.2007 & {\cellcolor[HTML]{097940}} \color[HTML]{F1F1F1} 0.2802 & {\cellcolor[HTML]{16914D}} \color[HTML]{000000} 0.3479 & {\cellcolor[HTML]{42AC5A}} \color[HTML]{000000} 0.4031 \\
\color[HTML]{011773} \bfseries \texttt{Google\_\allowbreak{}SAI-\allowbreak{}FluEns} & 5,824 & {\cellcolor[HTML]{15904C}} \color[HTML]{000000} 0.2088 & {\cellcolor[HTML]{006837}} \color[HTML]{F1F1F1} 0.2739 & {\cellcolor[HTML]{026C39}} \color[HTML]{F1F1F1} 0.3286 & {\cellcolor[HTML]{138C4A}} \color[HTML]{000000} 0.3756 \\
\texttt{FluSight-\allowbreak{}trained\_\allowbreak{}mean} & 5,824 & {\cellcolor[HTML]{006837}} \color[HTML]{F1F1F1} 0.2010 & {\cellcolor[HTML]{0E8245}} \color[HTML]{F1F1F1} 0.2831 & {\cellcolor[HTML]{18954F}} \color[HTML]{000000} 0.3502 & {\cellcolor[HTML]{4EB15D}} \color[HTML]{000000} 0.4085 \\
\texttt{FluSight-\allowbreak{}HJudge\_\allowbreak{}ensemble} & 5,824 & {\cellcolor[HTML]{0A7B41}} \color[HTML]{F1F1F1} 0.2047 & {\cellcolor[HTML]{118848}} \color[HTML]{000000} 0.2852 & {\cellcolor[HTML]{17934E}} \color[HTML]{000000} 0.3491 & {\cellcolor[HTML]{33A456}} \color[HTML]{000000} 0.3961 \\
\texttt{UMass-\allowbreak{}flusion} & 5,616 & {\cellcolor[HTML]{3CA959}} \color[HTML]{000000} 0.2149 & {\cellcolor[HTML]{108647}} \color[HTML]{F1F1F1} 0.2847 & {\cellcolor[HTML]{148E4B}} \color[HTML]{000000} 0.3464 & {\cellcolor[HTML]{51B35E}} \color[HTML]{000000} 0.4092 \\
\texttt{OHT\_\allowbreak{}JHU-\allowbreak{}nbxd} & 5,824 & {\cellcolor[HTML]{33A456}} \color[HTML]{000000} 0.2138 & {\cellcolor[HTML]{199750}} \color[HTML]{000000} 0.2902 & {\cellcolor[HTML]{118848}} \color[HTML]{000000} 0.3435 & {\cellcolor[HTML]{148E4B}} \color[HTML]{000000} 0.3780 \\
\texttt{FluSight-\allowbreak{}trained\_\allowbreak{}med} & 5,824 & {\cellcolor[HTML]{18954F}} \color[HTML]{000000} 0.2101 & {\cellcolor[HTML]{36A657}} \color[HTML]{000000} 0.2971 & {\cellcolor[HTML]{3FAA59}} \color[HTML]{000000} 0.3633 & {\cellcolor[HTML]{60BA62}} \color[HTML]{000000} 0.4159 \\
\texttt{UVAFluX-\allowbreak{}FS\_\allowbreak{}OptimWISE} & 4,992 & {\cellcolor[HTML]{7DC765}} \color[HTML]{000000} 0.2236 & {\cellcolor[HTML]{138C4A}} \color[HTML]{000000} 0.2865 & {\cellcolor[HTML]{006837}} \color[HTML]{F1F1F1} 0.3261 & {\cellcolor[HTML]{006837}} \color[HTML]{F1F1F1} 0.3501 \\
\texttt{FluSight-\allowbreak{}ensemble} & 5,824 & {\cellcolor[HTML]{1E9A51}} \color[HTML]{000000} 0.2112 & {\cellcolor[HTML]{42AC5A}} \color[HTML]{000000} 0.2995 & {\cellcolor[HTML]{54B45F}} \color[HTML]{000000} 0.3706 & {\cellcolor[HTML]{70C164}} \color[HTML]{000000} 0.4241 \\
\texttt{MIGHTE-\allowbreak{}Joint} & 4,992 & {\cellcolor[HTML]{30A356}} \color[HTML]{000000} 0.2134 & {\cellcolor[HTML]{39A758}} \color[HTML]{000000} 0.2978 & {\cellcolor[HTML]{199750}} \color[HTML]{000000} 0.3510 & {\cellcolor[HTML]{279F53}} \color[HTML]{000000} 0.3905 \\
\texttt{NAU-\allowbreak{}vulPES} & 5,824 & {\cellcolor[HTML]{2DA155}} \color[HTML]{000000} 0.2129 & {\cellcolor[HTML]{3FAA59}} \color[HTML]{000000} 0.2992 & {\cellcolor[HTML]{3CA959}} \color[HTML]{000000} 0.3630 & {\cellcolor[HTML]{54B45F}} \color[HTML]{000000} 0.4111 \\
\texttt{FluSight-\allowbreak{}lop\_\allowbreak{}norm} & 5,824 & {\cellcolor[HTML]{66BD63}} \color[HTML]{000000} 0.2203 & {\cellcolor[HTML]{5DB961}} \color[HTML]{000000} 0.3055 & {\cellcolor[HTML]{69BE63}} \color[HTML]{000000} 0.3769 & {\cellcolor[HTML]{78C565}} \color[HTML]{000000} 0.4282 \\
\texttt{CFA\_\allowbreak{}Pyrenew-\allowbreak{}Pyrenew\_\allowbreak{}HE\_\allowbreak{}Flu} & 2,398 & {\cellcolor[HTML]{51B35E}} \color[HTML]{000000} 0.2174 & {\cellcolor[HTML]{87CB67}} \color[HTML]{000000} 0.3158 & {\cellcolor[HTML]{D6D6D6}} \color[HTML]{000000}  & {\cellcolor[HTML]{D6D6D6}} \color[HTML]{000000}  \\
\texttt{UGA\_\allowbreak{}flucast-\allowbreak{}Scenariocast} & 5,408 & {\cellcolor[HTML]{89CC67}} \color[HTML]{000000} 0.2253 & {\cellcolor[HTML]{96D268}} \color[HTML]{000000} 0.3200 & {\cellcolor[HTML]{7DC765}} \color[HTML]{000000} 0.3847 & {\cellcolor[HTML]{75C465}} \color[HTML]{000000} 0.4264 \\
\texttt{MIGHTE-\allowbreak{}Nsemble} & 5,824 & {\cellcolor[HTML]{B9E176}} \color[HTML]{000000} 0.2334 & {\cellcolor[HTML]{8ECF67}} \color[HTML]{000000} 0.3178 & {\cellcolor[HTML]{6BBF64}} \color[HTML]{000000} 0.3780 & {\cellcolor[HTML]{91D068}} \color[HTML]{000000} 0.4410 \\
\texttt{UGA\_\allowbreak{}CEID-\allowbreak{}auto\_\allowbreak{}AVG\_\allowbreak{}LB} & 5,408 & {\cellcolor[HTML]{8CCD67}} \color[HTML]{000000} 0.2259 & {\cellcolor[HTML]{C9E881}} \color[HTML]{000000} 0.3353 & {\cellcolor[HTML]{FFFBB8}} \color[HTML]{000000} 0.4589 & {\cellcolor[HTML]{FEE593}} \color[HTML]{000000} 0.5705 \\
\texttt{NEU\_\allowbreak{}ISI-\allowbreak{}AdaptiveEnsemble} & 5,240 & {\cellcolor[HTML]{FFFDBC}} \color[HTML]{000000} 0.2506 & {\cellcolor[HTML]{B1DE71}} \color[HTML]{000000} 0.3273 & {\cellcolor[HTML]{82C966}} \color[HTML]{000000} 0.3872 & {\cellcolor[HTML]{6BBF64}} \color[HTML]{000000} 0.4208 \\
\texttt{PSI-\allowbreak{}PROF\_\allowbreak{}MOA} & 5,824 & {\cellcolor[HTML]{CDEA83}} \color[HTML]{000000} 0.2372 & {\cellcolor[HTML]{EFF8AA}} \color[HTML]{000000} 0.3503 & {\cellcolor[HTML]{E9F6A1}} \color[HTML]{000000} 0.4367 & {\cellcolor[HTML]{F1F9AC}} \color[HTML]{000000} 0.5082 \\
\texttt{UGA\_\allowbreak{}flucast-\allowbreak{}Copycat} & 5,824 & {\cellcolor[HTML]{DFF293}} \color[HTML]{000000} 0.2410 & {\cellcolor[HTML]{FFFEBE}} \color[HTML]{000000} 0.3571 & {\cellcolor[HTML]{FBFDBA}} \color[HTML]{000000} 0.4482 & {\cellcolor[HTML]{FFFEBE}} \color[HTML]{000000} 0.5202 \\
\texttt{CU-\allowbreak{}ARNB\_\allowbreak{}Net} & 5,200 & {\cellcolor[HTML]{FFFEBE}} \color[HTML]{000000} 0.2504 & {\cellcolor[HTML]{F1F9AC}} \color[HTML]{000000} 0.3512 & {\cellcolor[HTML]{B9E176}} \color[HTML]{000000} 0.4108 & {\cellcolor[HTML]{7AC665}} \color[HTML]{000000} 0.4296 \\
\texttt{PSI-\allowbreak{}PROF} & 5,824 & {\cellcolor[HTML]{FAFDB8}} \color[HTML]{000000} 0.2477 & {\cellcolor[HTML]{FDFEBC}} \color[HTML]{000000} 0.3562 & {\cellcolor[HTML]{F2FAAE}} \color[HTML]{000000} 0.4428 & {\cellcolor[HTML]{F4FAB0}} \color[HTML]{000000} 0.5101 \\
\texttt{NAU-\allowbreak{}FourCAT} & 5,616 & {\cellcolor[HTML]{FFF8B4}} \color[HTML]{000000} 0.2574 & {\cellcolor[HTML]{F5FBB2}} \color[HTML]{000000} 0.3530 & {\cellcolor[HTML]{E2F397}} \color[HTML]{000000} 0.4323 & {\cellcolor[HTML]{CDEA83}} \color[HTML]{000000} 0.4783 \\
\texttt{Gatech-\allowbreak{}ensemble\_\allowbreak{}prob} & 5,660 & {\cellcolor[HTML]{FAFDB8}} \color[HTML]{000000} 0.2476 & {\cellcolor[HTML]{FFF6B0}} \color[HTML]{000000} 0.3700 & {\cellcolor[HTML]{FFF5AE}} \color[HTML]{000000} 0.4685 & {\cellcolor[HTML]{FFF3AC}} \color[HTML]{000000} 0.5425 \\
\texttt{CEPH-\allowbreak{}Rtrend\_\allowbreak{}fluH} & 5,824 & {\cellcolor[HTML]{FFF7B2}} \color[HTML]{000000} 0.2583 & {\cellcolor[HTML]{FFFDBC}} \color[HTML]{000000} 0.3600 & {\cellcolor[HTML]{FEEDA1}} \color[HTML]{000000} 0.4815 & {\cellcolor[HTML]{FDC171}} \color[HTML]{000000} 0.6176 \\
\texttt{MOBS-\allowbreak{}EpyStrain\_\allowbreak{}Flu} & 5,209 & {\cellcolor[HTML]{FFF7B2}} \color[HTML]{000000} 0.2590 & {\cellcolor[HTML]{FFF8B4}} \color[HTML]{000000} 0.3659 & {\cellcolor[HTML]{FFF5AE}} \color[HTML]{000000} 0.4680 & {\cellcolor[HTML]{FEEC9F}} \color[HTML]{000000} 0.5559 \\
\texttt{Gatech-\allowbreak{}ensemble\_\allowbreak{}stat} & 5,824 & {\cellcolor[HTML]{DFF293}} \color[HTML]{000000} 0.2410 & {\cellcolor[HTML]{FEEB9D}} \color[HTML]{000000} 0.3855 & {\cellcolor[HTML]{FEE28F}} \color[HTML]{000000} 0.4998 & {\cellcolor[HTML]{FED07E}} \color[HTML]{000000} 0.6007 \\
\texttt{NEU\_\allowbreak{}ISI-\allowbreak{}FluBcast} & 5,088 & {\cellcolor[HTML]{FFFEBE}} \color[HTML]{000000} 0.2490 & {\cellcolor[HTML]{FEE999}} \color[HTML]{000000} 0.3892 & {\cellcolor[HTML]{FEE797}} \color[HTML]{000000} 0.4912 & {\cellcolor[HTML]{FEE695}} \color[HTML]{000000} 0.5682 \\
\texttt{UVAFluX-\allowbreak{}Ensemble} & 5,772 & {\cellcolor[HTML]{FEEFA3}} \color[HTML]{000000} 0.2691 & {\cellcolor[HTML]{FEEFA3}} \color[HTML]{000000} 0.3816 & {\cellcolor[HTML]{FEEC9F}} \color[HTML]{000000} 0.4836 & {\cellcolor[HTML]{FFF1A8}} \color[HTML]{000000} 0.5480 \\
\texttt{MOBS-\allowbreak{}GLEAM\_\allowbreak{}RL\_\allowbreak{}FLUH} & 5,696 & {\cellcolor[HTML]{FEEDA1}} \color[HTML]{000000} 0.2701 & {\cellcolor[HTML]{FEEB9D}} \color[HTML]{000000} 0.3861 & {\cellcolor[HTML]{FEEFA3}} \color[HTML]{000000} 0.4790 & {\cellcolor[HTML]{FEEC9F}} \color[HTML]{000000} 0.5574 \\
\texttt{NAU-\allowbreak{}epymorph} & 5,824 & {\cellcolor[HTML]{FEE18D}} \color[HTML]{000000} 0.2841 & {\cellcolor[HTML]{FFF0A6}} \color[HTML]{000000} 0.3795 & {\cellcolor[HTML]{FFFEBE}} \color[HTML]{000000} 0.4509 & {\cellcolor[HTML]{E5F49B}} \color[HTML]{000000} 0.4963 \\
\texttt{CFA\_\allowbreak{}Pyrenew-\allowbreak{}Pyrenew\_\allowbreak{}H\_\allowbreak{}Flu} & 2,814 & {\cellcolor[HTML]{FFF0A6}} \color[HTML]{000000} 0.2678 & {\cellcolor[HTML]{FECE7C}} \color[HTML]{000000} 0.4174 & {\cellcolor[HTML]{D6D6D6}} \color[HTML]{000000}  & {\cellcolor[HTML]{D6D6D6}} \color[HTML]{000000}  \\
\texttt{UMass-\allowbreak{}AR2} & 5,772 & {\cellcolor[HTML]{FED884}} \color[HTML]{000000} 0.2926 & {\cellcolor[HTML]{FECA79}} \color[HTML]{000000} 0.4210 & {\cellcolor[HTML]{FEC877}} \color[HTML]{000000} 0.5282 & {\cellcolor[HTML]{FDBB6C}} \color[HTML]{000000} 0.6260 \\
\texttt{Cornell\_\allowbreak{}JHU-\allowbreak{}hierarchSIR} & 5,824 & {\cellcolor[HTML]{FBA35C}} \color[HTML]{000000} 0.3298 & {\cellcolor[HTML]{FECE7C}} \color[HTML]{000000} 0.4174 & {\cellcolor[HTML]{FEE695}} \color[HTML]{000000} 0.4944 & {\cellcolor[HTML]{FEEC9F}} \color[HTML]{000000} 0.5567 \\
\texttt{UNC\_\allowbreak{}IDD-\allowbreak{}InfluPaint} & 5,712 & {\cellcolor[HTML]{FDB567}} \color[HTML]{000000} 0.3187 & {\cellcolor[HTML]{FDBF6F}} \color[HTML]{000000} 0.4314 & {\cellcolor[HTML]{FECE7C}} \color[HTML]{000000} 0.5232 & {\cellcolor[HTML]{FEE18D}} \color[HTML]{000000} 0.5789 \\
\texttt{FluSight-\allowbreak{}baseline} & 5,824 & {\cellcolor[HTML]{FDBF6F}} \color[HTML]{000000} 0.3103 & {\cellcolor[HTML]{F88C51}} \color[HTML]{000000} 0.4694 & {\cellcolor[HTML]{F26841}} \color[HTML]{000000} 0.6139 & {\cellcolor[HTML]{E75337}} \color[HTML]{000000} 0.7279 \\
\texttt{VTSanghani-\allowbreak{}PRIME} & 4,953 & {\cellcolor[HTML]{FDAD60}} \color[HTML]{000000} 0.3239 & {\cellcolor[HTML]{FA9B58}} \color[HTML]{000000} 0.4592 & {\cellcolor[HTML]{FDAF62}} \color[HTML]{000000} 0.5560 & {\cellcolor[HTML]{FDB163}} \color[HTML]{000000} 0.6385 \\
\texttt{CU-\allowbreak{}ensemble} & 5,824 & {\cellcolor[HTML]{F67C4A}} \color[HTML]{000000} 0.3520 & {\cellcolor[HTML]{FDBD6D}} \color[HTML]{000000} 0.4342 & {\cellcolor[HTML]{FEE491}} \color[HTML]{000000} 0.4971 & {\cellcolor[HTML]{FFF1A8}} \color[HTML]{000000} 0.5469 \\
\texttt{NIH-\allowbreak{}Flu\_\allowbreak{}ARIMA} & 4,876 & {\cellcolor[HTML]{F98E52}} \color[HTML]{000000} 0.3417 & {\cellcolor[HTML]{FBA05B}} \color[HTML]{000000} 0.4562 & {\cellcolor[HTML]{FED07E}} \color[HTML]{000000} 0.5203 & {\cellcolor[HTML]{FEE695}} \color[HTML]{000000} 0.5699 \\
\texttt{LosAlamos-\allowbreak{}DoSiDo} & 5,616 & {\cellcolor[HTML]{F99153}} \color[HTML]{000000} 0.3392 & {\cellcolor[HTML]{E34933}} \color[HTML]{000000} 0.5181 & {\cellcolor[HTML]{D62F27}} \color[HTML]{F1F1F1} 0.6628 & {\cellcolor[HTML]{C82227}} \color[HTML]{F1F1F1} 0.7812 \\
\texttt{SigSci-\allowbreak{}TSENS} & 4,752 & {\cellcolor[HTML]{F57245}} \color[HTML]{000000} 0.3579 & {\cellcolor[HTML]{E95538}} \color[HTML]{000000} 0.5088 & {\cellcolor[HTML]{F57547}} \color[HTML]{000000} 0.6031 & {\cellcolor[HTML]{F99355}} \color[HTML]{000000} 0.6668 \\
\texttt{UM-\allowbreak{}DeepOutbreak} & 5,200 & {\cellcolor[HTML]{F88950}} \color[HTML]{000000} 0.3444 & {\cellcolor[HTML]{D02927}} \color[HTML]{F1F1F1} 0.5413 & {\cellcolor[HTML]{DC3B2C}} \color[HTML]{000000} 0.6536 & {\cellcolor[HTML]{EE613E}} \color[HTML]{000000} 0.7131 \\
\texttt{LosAlamos-\allowbreak{}ThinMint} & 5,616 & {\cellcolor[HTML]{C82227}} \color[HTML]{F1F1F1} 0.4082 & {\cellcolor[HTML]{D83128}} \color[HTML]{F1F1F1} 0.5349 & {\cellcolor[HTML]{DE402E}} \color[HTML]{000000} 0.6483 & {\cellcolor[HTML]{E44C34}} \color[HTML]{000000} 0.7349 \\
\texttt{LosAlamos\_\allowbreak{}NAU-\allowbreak{}CModel\_\allowbreak{}Flu} & 5,820 & {\cellcolor[HTML]{A90426}} \color[HTML]{F1F1F1} 0.4315 & {\cellcolor[HTML]{A70226}} \color[HTML]{F1F1F1} 0.5798 & {\cellcolor[HTML]{A50026}} \color[HTML]{F1F1F1} 0.7161 & {\cellcolor[HTML]{A50026}} \color[HTML]{F1F1F1} 0.8241 \\
\texttt{UGA\_\allowbreak{}CEID-\allowbreak{}Walk} & 5,356 & {\cellcolor[HTML]{A50026}} \color[HTML]{F1F1F1} 0.4351 & {\cellcolor[HTML]{A50026}} \color[HTML]{F1F1F1} 0.5816 & {\cellcolor[HTML]{B10B26}} \color[HTML]{F1F1F1} 0.7034 & {\cellcolor[HTML]{B71126}} \color[HTML]{F1F1F1} 0.8004 \\

%% file: tables/internal_covid_crosshub_rel.tex
\color[HTML]{011773} \bfseries \texttt{Google\_\allowbreak{}SAI-\allowbreak{}Ensemble} & 5200 & {\cellcolor[HTML]{D8F6E5}} \color[HTML]{000000} 0.97 & {\cellcolor[HTML]{006837}} \color[HTML]{F1F1F1} 0.235 & {\cellcolor[HTML]{3DD47F}} \color[HTML]{000000} 0.88 & {\cellcolor[HTML]{60BA62}} \color[HTML]{F1F1F1} 12.6 & {\cellcolor[HTML]{7FC866}} \color[HTML]{000000} 19.5 & 92\% & CDC \\
*\texttt{G-\allowbreak{}CMU\_\allowbreak{}TimeSeries-\allowbreak{}UMass\_\allowbreak{}gbqr} & 5200 & {\cellcolor[HTML]{F6FDF9}} \color[HTML]{000000} 0.99 & {\cellcolor[HTML]{097940}} \color[HTML]{F1F1F1} 0.241 & {\cellcolor[HTML]{B3EECD}} \color[HTML]{000000} 0.95 & {\cellcolor[HTML]{93D168}} \color[HTML]{000000} 13.6 & {\cellcolor[HTML]{98D368}} \color[HTML]{000000} 20.2 & 86\% & Recombination \\
*\texttt{G-\allowbreak{}CADPH-\allowbreak{}CovidCAT\_\allowbreak{}Ensemble} & 3536 & {\cellcolor[HTML]{F9FEFB}} \color[HTML]{000000} 0.99 & {\cellcolor[HTML]{57B65F}} \color[HTML]{F1F1F1} 0.263 & {\cellcolor[HTML]{70E0A1}} \color[HTML]{000000} 0.91 & {\cellcolor[HTML]{006837}} \color[HTML]{F1F1F1} 10.2 & {\cellcolor[HTML]{006837}} \color[HTML]{F1F1F1} 15.0 & 94\% & Scientist-inspired \\
\texttt{CovidHub-\allowbreak{}ensemble} & 5200 & {\cellcolor[HTML]{FFFFFE}} \color[HTML]{000000} 1.00 & {\cellcolor[HTML]{0D8044}} \color[HTML]{F1F1F1} 0.243 & {\cellcolor[HTML]{FFFFFE}} \color[HTML]{000000} 1.00 & {\cellcolor[HTML]{BBE278}} \color[HTML]{000000} 14.5 & {\cellcolor[HTML]{DCF08F}} \color[HTML]{000000} 22.6 & 93\% & CDC ensemble \\
\texttt{UGA\_\allowbreak{}flucast-\allowbreak{}INFLAenza} & 5200 & {\cellcolor[HTML]{FFFFFE}} \color[HTML]{000000} 1.00 & {\cellcolor[HTML]{108647}} \color[HTML]{F1F1F1} 0.245 & {\cellcolor[HTML]{C6F2D9}} \color[HTML]{000000} 0.96 & {\cellcolor[HTML]{AFDD70}} \color[HTML]{000000} 14.2 & {\cellcolor[HTML]{BFE47A}} \color[HTML]{000000} 21.5 & 88\% & CDC \\
*\texttt{G-\allowbreak{}DeepResearch\_\allowbreak{}CounterfactualSimulation} & 5200 & {\cellcolor[HTML]{FFF8E6}} \color[HTML]{000000} 1.03 & {\cellcolor[HTML]{138C4A}} \color[HTML]{F1F1F1} 0.247 & {\cellcolor[HTML]{5EDC95}} \color[HTML]{000000} 0.90 & {\cellcolor[HTML]{63BC62}} \color[HTML]{F1F1F1} 12.6 & {\cellcolor[HTML]{78C565}} \color[HTML]{000000} 19.2 & 86\% & AI-led \\
*\texttt{G-\allowbreak{}UM-\allowbreak{}DeepOutbreak} & 4368 & {\cellcolor[HTML]{FEF4D7}} \color[HTML]{000000} 1.04 & {\cellcolor[HTML]{39A758}} \color[HTML]{F1F1F1} 0.257 & {\cellcolor[HTML]{62DC97}} \color[HTML]{000000} 0.90 & {\cellcolor[HTML]{08773F}} \color[HTML]{F1F1F1} 10.6 & {\cellcolor[HTML]{0D8044}} \color[HTML]{F1F1F1} 15.9 & 87\% & Scientist-inspired \\
*\texttt{G-\allowbreak{}UMass-\allowbreak{}gbqr} & 4368 & {\cellcolor[HTML]{FEF1CF}} \color[HTML]{000000} 1.05 & {\cellcolor[HTML]{45AD5B}} \color[HTML]{F1F1F1} 0.260 & {\cellcolor[HTML]{CDF4DE}} \color[HTML]{000000} 0.97 & {\cellcolor[HTML]{17934E}} \color[HTML]{F1F1F1} 11.3 & {\cellcolor[HTML]{16914D}} \color[HTML]{F1F1F1} 16.6 & 85\% & Scientist-inspired \\
\texttt{CMU-\allowbreak{}TimeSeries} & 5200 & {\cellcolor[HTML]{FEF0CA}} \color[HTML]{000000} 1.06 & {\cellcolor[HTML]{33A456}} \color[HTML]{F1F1F1} 0.256 & {\cellcolor[HTML]{FDE7AA}} \color[HTML]{000000} 1.08 & {\cellcolor[HTML]{DDF191}} \color[HTML]{000000} 15.4 & {\cellcolor[HTML]{F1F9AC}} \color[HTML]{000000} 23.6 & 93\% & CDC \\
\texttt{G-\allowbreak{}JHU\_\allowbreak{}CSSE-\allowbreak{}CSSE\_\allowbreak{}Ensemble} & 4160 & {\cellcolor[HTML]{FEEEC4}} \color[HTML]{000000} 1.06 & {\cellcolor[HTML]{66BD63}} \color[HTML]{F1F1F1} 0.267 & {\cellcolor[HTML]{FEF3D4}} \color[HTML]{000000} 1.04 & {\cellcolor[HTML]{219C52}} \color[HTML]{F1F1F1} 11.6 & {\cellcolor[HTML]{42AC5A}} \color[HTML]{F1F1F1} 17.8 & 95\% & Scientist-inspired \\
*\texttt{G-\allowbreak{}DeepResearch\_\allowbreak{}RegimeSwitchingDetection} & 5200 & {\cellcolor[HTML]{FDE7AA}} \color[HTML]{000000} 1.09 & {\cellcolor[HTML]{54B45F}} \color[HTML]{F1F1F1} 0.263 & {\cellcolor[HTML]{D4F6E3}} \color[HTML]{000000} 0.97 & {\cellcolor[HTML]{A2D76A}} \color[HTML]{000000} 13.9 & {\cellcolor[HTML]{96D268}} \color[HTML]{000000} 20.1 & 77\% & AI-led \\
*\texttt{G-\allowbreak{}MOBS-\allowbreak{}GLEAM\_\allowbreak{}COVID} & 4368 & {\cellcolor[HTML]{FDE39B}} \color[HTML]{000000} 1.11 & {\cellcolor[HTML]{84CA66}} \color[HTML]{000000} 0.274 & {\cellcolor[HTML]{FFFCF3}} \color[HTML]{000000} 1.01 & {\cellcolor[HTML]{30A356}} \color[HTML]{F1F1F1} 11.8 & {\cellcolor[HTML]{51B35E}} \color[HTML]{F1F1F1} 18.2 & 94\% & Scientist-inspired \\
*\texttt{G-\allowbreak{}CMU\_\allowbreak{}climate\_\allowbreak{}baseline-\allowbreak{}UMass\_\allowbreak{}ar6\_\allowbreak{}pooled} & 5200 & {\cellcolor[HTML]{FDDD86}} \color[HTML]{000000} 1.14 & {\cellcolor[HTML]{87CB67}} \color[HTML]{000000} 0.274 & {\cellcolor[HTML]{FDDD86}} \color[HTML]{000000} 1.12 & {\cellcolor[HTML]{F1F9AC}} \color[HTML]{000000} 16.0 & {\cellcolor[HTML]{F4FAB0}} \color[HTML]{000000} 23.8 & 92\% & Recombination \\
\texttt{G-\allowbreak{}UGA\_\allowbreak{}flucast-\allowbreak{}INFLAenza} & 4160 & {\cellcolor[HTML]{FDDD86}} \color[HTML]{000000} 1.14 & {\cellcolor[HTML]{ADDC6F}} \color[HTML]{000000} 0.284 & {\cellcolor[HTML]{FDE093}} \color[HTML]{000000} 1.11 & {\cellcolor[HTML]{4BB05C}} \color[HTML]{F1F1F1} 12.3 & {\cellcolor[HTML]{5AB760}} \color[HTML]{F1F1F1} 18.4 & 88\% & Scientist-inspired \\
\texttt{CEPH-\allowbreak{}Rtrend\_\allowbreak{}covid} & 5200 & {\cellcolor[HTML]{FCD66E}} \color[HTML]{000000} 1.16 & {\cellcolor[HTML]{A9DA6C}} \color[HTML]{000000} 0.283 & {\cellcolor[HTML]{FCD05A}} \color[HTML]{000000} 1.17 & {\cellcolor[HTML]{FFF1A8}} \color[HTML]{000000} 17.1 & {\cellcolor[HTML]{FEE08B}} \color[HTML]{000000} 26.2 & 88\% & CDC \\
\texttt{G-\allowbreak{}UMass-\allowbreak{}ar6\_\allowbreak{}pooled} & 4160 & {\cellcolor[HTML]{FCD467}} \color[HTML]{000000} 1.18 & {\cellcolor[HTML]{CDEA83}} \color[HTML]{000000} 0.294 & {\cellcolor[HTML]{FDE6A8}} \color[HTML]{000000} 1.09 & {\cellcolor[HTML]{42AC5A}} \color[HTML]{F1F1F1} 12.1 & {\cellcolor[HTML]{6EC064}} \color[HTML]{000000} 18.9 & 94\% & Scientist-inspired \\
\texttt{G-\allowbreak{}CMU-\allowbreak{}TimeSeries} & 4160 & {\cellcolor[HTML]{FCD364}} \color[HTML]{000000} 1.18 & {\cellcolor[HTML]{CFEB85}} \color[HTML]{000000} 0.295 & {\cellcolor[HTML]{FDD979}} \color[HTML]{000000} 1.14 & {\cellcolor[HTML]{5DB961}} \color[HTML]{F1F1F1} 12.6 & {\cellcolor[HTML]{7FC866}} \color[HTML]{000000} 19.5 & 97\% & Scientist-inspired \\
\texttt{G-\allowbreak{}Metaculus-\allowbreak{}cp} & 4160 & {\cellcolor[HTML]{FAC756}} \color[HTML]{000000} 1.20 & {\cellcolor[HTML]{DFF293}} \color[HTML]{000000} 0.300 & {\cellcolor[HTML]{FCD364}} \color[HTML]{000000} 1.16 & {\cellcolor[HTML]{6EC064}} \color[HTML]{000000} 12.8 & {\cellcolor[HTML]{84CA66}} \color[HTML]{000000} 19.6 & 87\% & Scientist-inspired \\
\texttt{CFA-\allowbreak{}EpiAutoGP} & 4604 & {\cellcolor[HTML]{F8BD53}} \color[HTML]{000000} 1.22 & {\cellcolor[HTML]{CDEA83}} \color[HTML]{000000} 0.294 & {\cellcolor[HTML]{DD492F}} \color[HTML]{F1F1F1} 1.33 & {\cellcolor[HTML]{FECE7C}} \color[HTML]{000000} 18.2 & {\cellcolor[HTML]{FCAA5F}} \color[HTML]{000000} 28.2 & 89\% & CDC \\
*\texttt{G-\allowbreak{}CEPH\_\allowbreak{}Rtrend\_\allowbreak{}covid-\allowbreak{}CMU\_\allowbreak{}climate\_\allowbreak{}baseline} & 5200 & {\cellcolor[HTML]{F6B851}} \color[HTML]{000000} 1.22 & {\cellcolor[HTML]{C9E881}} \color[HTML]{000000} 0.293 & {\cellcolor[HTML]{FDDB7E}} \color[HTML]{000000} 1.13 & {\cellcolor[HTML]{F4FAB0}} \color[HTML]{000000} 16.1 & {\cellcolor[HTML]{FFFEBE}} \color[HTML]{000000} 24.3 & 82\% & Recombination \\
\texttt{UMass-\allowbreak{}gbqr} & 5148 & {\cellcolor[HTML]{F4AD4E}} \color[HTML]{000000} 1.24 & {\cellcolor[HTML]{E0F295}} \color[HTML]{000000} 0.301 & {\cellcolor[HTML]{EC8D44}} \color[HTML]{F1F1F1} 1.25 & {\cellcolor[HTML]{FEDE89}} \color[HTML]{000000} 17.8 & {\cellcolor[HTML]{FEC877}} \color[HTML]{000000} 27.0 & 86\% & CDC \\
\texttt{UMass-\allowbreak{}ar6\_\allowbreak{}pooled} & 5148 & {\cellcolor[HTML]{EC8D44}} \color[HTML]{F1F1F1} 1.28 & {\cellcolor[HTML]{F7FCB4}} \color[HTML]{000000} 0.310 & {\cellcolor[HTML]{D73027}} \color[HTML]{F1F1F1} 1.38 & {\cellcolor[HTML]{FDAF62}} \color[HTML]{000000} 19.0 & {\cellcolor[HTML]{F98E52}} \color[HTML]{F1F1F1} 29.0 & 95\% & CDC \\
\texttt{G-\allowbreak{}CFA-\allowbreak{}EpiAutoGP} & 4160 & {\cellcolor[HTML]{EB8842}} \color[HTML]{F1F1F1} 1.29 & {\cellcolor[HTML]{FEEFA3}} \color[HTML]{000000} 0.321 & {\cellcolor[HTML]{EB8542}} \color[HTML]{F1F1F1} 1.26 & {\cellcolor[HTML]{A2D76A}} \color[HTML]{000000} 13.9 & {\cellcolor[HTML]{B1DE71}} \color[HTML]{000000} 21.0 & 97\% & Scientist-inspired \\
*\texttt{G-\allowbreak{}NEU\_\allowbreak{}ISI-\allowbreak{}AdaptiveEnsemble} & 4368 & {\cellcolor[HTML]{E8793E}} \color[HTML]{F1F1F1} 1.31 & {\cellcolor[HTML]{FEEFA3}} \color[HTML]{000000} 0.321 & {\cellcolor[HTML]{FEEFC7}} \color[HTML]{000000} 1.05 & {\cellcolor[HTML]{4EB15D}} \color[HTML]{F1F1F1} 12.3 & {\cellcolor[HTML]{69BE63}} \color[HTML]{F1F1F1} 18.8 & 90\% & Scientist-inspired \\
\texttt{G-\allowbreak{}CEPH-\allowbreak{}Rtrend\_\allowbreak{}covid} & 4160 & {\cellcolor[HTML]{D83328}} \color[HTML]{F1F1F1} 1.42 & {\cellcolor[HTML]{F99153}} \color[HTML]{000000} 0.352 & {\cellcolor[HTML]{D73027}} \color[HTML]{F1F1F1} 1.37 & {\cellcolor[HTML]{D1EC86}} \color[HTML]{000000} 15.0 & {\cellcolor[HTML]{E5F49B}} \color[HTML]{000000} 23.0 & 88\% & Scientist-inspired \\
\texttt{OHT\_\allowbreak{}JHU-\allowbreak{}nbxd} & 5200 & {\cellcolor[HTML]{D73027}} \color[HTML]{F1F1F1} 1.43 & {\cellcolor[HTML]{FCA85E}} \color[HTML]{000000} 0.346 & {\cellcolor[HTML]{D73027}} \color[HTML]{F1F1F1} 1.49 & {\cellcolor[HTML]{D22B27}} \color[HTML]{F1F1F1} 21.7 & {\cellcolor[HTML]{C41E27}} \color[HTML]{F1F1F1} 32.4 & 68\% & CDC \\
\texttt{CovidHub-\allowbreak{}baseline} & 5200 & {\cellcolor[HTML]{D73027}} \color[HTML]{F1F1F1} 1.46 & {\cellcolor[HTML]{F8864F}} \color[HTML]{F1F1F1} 0.354 & {\cellcolor[HTML]{E7743C}} \color[HTML]{F1F1F1} 1.28 & {\cellcolor[HTML]{FDC171}} \color[HTML]{000000} 18.6 & {\cellcolor[HTML]{FED683}} \color[HTML]{000000} 26.5 & 96\% & CDC baseline \\
\texttt{UM-\allowbreak{}DeepOutbreak} & 4576 & {\cellcolor[HTML]{D73027}} \color[HTML]{F1F1F1} 1.66 & {\cellcolor[HTML]{A50026}} \color[HTML]{F1F1F1} 0.392 & {\cellcolor[HTML]{D73027}} \color[HTML]{F1F1F1} 1.58 & {\cellcolor[HTML]{A50026}} \color[HTML]{F1F1F1} 22.8 & {\cellcolor[HTML]{A50026}} \color[HTML]{F1F1F1} 33.6 & 92\% & CDC \\

%% file: tables/covid_logwis_by_horizon.tex
\texttt{CFA\_\allowbreak{}Pyrenew-\allowbreak{}Pyrenew\_\allowbreak{}HE\_\allowbreak{}COVID} & 2,354 & {\cellcolor[HTML]{006837}} \color[HTML]{F1F1F1} 0.1768 & {\cellcolor[HTML]{0F8446}} \color[HTML]{F1F1F1} 0.2191 & {\cellcolor[HTML]{D6D6D6}} \color[HTML]{000000}  & {\cellcolor[HTML]{D6D6D6}} \color[HTML]{000000}  \\
\color[HTML]{011773} \bfseries \texttt{Google\_\allowbreak{}SAI-\allowbreak{}Ensemble} & 5,200 & {\cellcolor[HTML]{54B45F}} \color[HTML]{000000} 0.1897 & {\cellcolor[HTML]{006837}} \color[HTML]{F1F1F1} 0.2128 & {\cellcolor[HTML]{006837}} \color[HTML]{F1F1F1} 0.2501 & {\cellcolor[HTML]{006837}} \color[HTML]{F1F1F1} 0.2869 \\
\texttt{CovidHub-\allowbreak{}ensemble} & 5,200 & {\cellcolor[HTML]{30A356}} \color[HTML]{000000} 0.1864 & {\cellcolor[HTML]{0F8446}} \color[HTML]{F1F1F1} 0.2192 & {\cellcolor[HTML]{1E9A51}} \color[HTML]{000000} 0.2657 & {\cellcolor[HTML]{16914D}} \color[HTML]{000000} 0.3017 \\
\texttt{UGA\_\allowbreak{}flucast-\allowbreak{}INFLAenza} & 5,200 & {\cellcolor[HTML]{7DC765}} \color[HTML]{000000} 0.1940 & {\cellcolor[HTML]{39A758}} \color[HTML]{000000} 0.2278 & {\cellcolor[HTML]{199750}} \color[HTML]{000000} 0.2649 & {\cellcolor[HTML]{08773F}} \color[HTML]{F1F1F1} 0.2925 \\
\texttt{CMU-\allowbreak{}TimeSeries} & 5,200 & {\cellcolor[HTML]{D9EF8B}} \color[HTML]{000000} 0.2060 & {\cellcolor[HTML]{84CA66}} \color[HTML]{000000} 0.2389 & {\cellcolor[HTML]{51B35E}} \color[HTML]{000000} 0.2757 & {\cellcolor[HTML]{199750}} \color[HTML]{000000} 0.3037 \\
\texttt{CEPH-\allowbreak{}Rtrend\_\allowbreak{}covid} & 5,200 & {\cellcolor[HTML]{FFF3AC}} \color[HTML]{000000} 0.2183 & {\cellcolor[HTML]{DAF08D}} \color[HTML]{000000} 0.2556 & {\cellcolor[HTML]{D5ED88}} \color[HTML]{000000} 0.3076 & {\cellcolor[HTML]{CDEA83}} \color[HTML]{000000} 0.3510 \\
\texttt{UMass-\allowbreak{}gbqr} & 5,148 & {\cellcolor[HTML]{F1F9AC}} \color[HTML]{000000} 0.2108 & {\cellcolor[HTML]{FFFEBE}} \color[HTML]{000000} 0.2655 & {\cellcolor[HTML]{FFF0A6}} \color[HTML]{000000} 0.3344 & {\cellcolor[HTML]{FEE491}} \color[HTML]{000000} 0.3956 \\
\texttt{CFA-\allowbreak{}EpiAutoGP} & 4,604 & {\cellcolor[HTML]{FFFEBE}} \color[HTML]{000000} 0.2134 & {\cellcolor[HTML]{FFFEBE}} \color[HTML]{000000} 0.2658 & {\cellcolor[HTML]{FFFEBE}} \color[HTML]{000000} 0.3236 & {\cellcolor[HTML]{FFFEBE}} \color[HTML]{000000} 0.3722 \\
\texttt{UMass-\allowbreak{}ar6\_\allowbreak{}pooled} & 5,148 & {\cellcolor[HTML]{FEE593}} \color[HTML]{000000} 0.2244 & {\cellcolor[HTML]{FEEDA1}} \color[HTML]{000000} 0.2758 & {\cellcolor[HTML]{FEE999}} \color[HTML]{000000} 0.3396 & {\cellcolor[HTML]{FED683}} \color[HTML]{000000} 0.4032 \\
\texttt{CFA\_\allowbreak{}Pyrenew-\allowbreak{}Pyrenew\_\allowbreak{}H\_\allowbreak{}COVID} & 2,520 & {\cellcolor[HTML]{FECA79}} \color[HTML]{000000} 0.2315 & {\cellcolor[HTML]{F88950}} \color[HTML]{000000} 0.3116 & {\cellcolor[HTML]{D6D6D6}} \color[HTML]{000000}  & {\cellcolor[HTML]{D6D6D6}} \color[HTML]{000000}  \\
\texttt{CovidHub-\allowbreak{}baseline} & 5,200 & {\cellcolor[HTML]{E34933}} \color[HTML]{000000} 0.2591 & {\cellcolor[HTML]{ED5F3C}} \color[HTML]{000000} 0.3236 & {\cellcolor[HTML]{F26841}} \color[HTML]{000000} 0.3917 & {\cellcolor[HTML]{F7814C}} \color[HTML]{000000} 0.4423 \\
\texttt{OHT\_\allowbreak{}JHU-\allowbreak{}nbxd} & 5,200 & {\cellcolor[HTML]{A50026}} \color[HTML]{F1F1F1} 0.2774 & {\cellcolor[HTML]{EB5A3A}} \color[HTML]{000000} 0.3249 & {\cellcolor[HTML]{FCA55D}} \color[HTML]{000000} 0.3711 & {\cellcolor[HTML]{FECA79}} \color[HTML]{000000} 0.4100 \\
\texttt{UM-\allowbreak{}DeepOutbreak} & 4,576 & {\cellcolor[HTML]{B71126}} \color[HTML]{F1F1F1} 0.2728 & {\cellcolor[HTML]{A50026}} \color[HTML]{F1F1F1} 0.3559 & {\cellcolor[HTML]{A50026}} \color[HTML]{F1F1F1} 0.4345 & {\cellcolor[HTML]{A50026}} \color[HTML]{F1F1F1} 0.5035 \\

%% file: tables/internal_rsv_crosshub_rel.tex
\color[HTML]{011773} \bfseries \texttt{Google\_\allowbreak{}SAI-\allowbreak{}RSVEns} & 4576 & {\cellcolor[HTML]{32D277}} \color[HTML]{000000} 0.90 & {\cellcolor[HTML]{006837}} \color[HTML]{F1F1F1} 0.238 & {\cellcolor[HTML]{1BA256}} \color[HTML]{F1F1F1} 0.87 & {\cellcolor[HTML]{006837}} \color[HTML]{F1F1F1} 14.7 & {\cellcolor[HTML]{006837}} \color[HTML]{F1F1F1} 21.2 & 85\% & CDC \\
*\texttt{G-\allowbreak{}general\_\allowbreak{}instruction\_\allowbreak{}2} & 4576 & {\cellcolor[HTML]{69DE9C}} \color[HTML]{000000} 0.92 & {\cellcolor[HTML]{108647}} \color[HTML]{F1F1F1} 0.244 & {\cellcolor[HTML]{1AA456}} \color[HTML]{F1F1F1} 0.87 & {\cellcolor[HTML]{006837}} \color[HTML]{F1F1F1} 14.7 & {\cellcolor[HTML]{097940}} \color[HTML]{F1F1F1} 21.4 & 86\% & AI-led \\
*\texttt{G-\allowbreak{}general\_\allowbreak{}instruction\_\allowbreak{}lgbm} & 4576 & {\cellcolor[HTML]{D1F5E0}} \color[HTML]{000000} 0.98 & {\cellcolor[HTML]{5DB961}} \color[HTML]{F1F1F1} 0.256 & {\cellcolor[HTML]{E7FAEF}} \color[HTML]{000000} 0.99 & {\cellcolor[HTML]{AFDD70}} \color[HTML]{000000} 16.3 & {\cellcolor[HTML]{4EB15D}} \color[HTML]{F1F1F1} 22.4 & 80\% & AI-led \\
\texttt{RSVHub-\allowbreak{}ensemble} & 4576 & {\cellcolor[HTML]{FFFFFE}} \color[HTML]{000000} 1.00 & {\cellcolor[HTML]{84CA66}} \color[HTML]{000000} 0.261 & {\cellcolor[HTML]{FFFFFE}} \color[HTML]{000000} 1.00 & {\cellcolor[HTML]{BBE278}} \color[HTML]{000000} 16.4 & {\cellcolor[HTML]{FEEA9B}} \color[HTML]{000000} 25.3 & 87\% & CDC ensemble \\
\texttt{CEPH-\allowbreak{}Rtrend\_\allowbreak{}rsv} & 4576 & {\cellcolor[HTML]{E05834}} \color[HTML]{F1F1F1} 1.24 & {\cellcolor[HTML]{EF633F}} \color[HTML]{F1F1F1} 0.313 & {\cellcolor[HTML]{D73027}} \color[HTML]{F1F1F1} 1.25 & {\cellcolor[HTML]{A50026}} \color[HTML]{F1F1F1} 19.8 & {\cellcolor[HTML]{A50026}} \color[HTML]{F1F1F1} 28.4 & 78\% & CDC \\
\texttt{RSVHub-\allowbreak{}baseline} & 4576 & {\cellcolor[HTML]{D73027}} \color[HTML]{F1F1F1} 1.33 & {\cellcolor[HTML]{A50026}} \color[HTML]{F1F1F1} 0.330 & {\cellcolor[HTML]{EA8040}} \color[HTML]{F1F1F1} 1.14 & {\cellcolor[HTML]{FB9D59}} \color[HTML]{000000} 18.3 & {\cellcolor[HTML]{CE2827}} \color[HTML]{F1F1F1} 27.8 & 93\% & CDC baseline \\

%% file: tables/rsv_logwis_by_horizon.tex
\texttt{CFA\_\allowbreak{}Pyrenew-\allowbreak{}Pyrenew\_\allowbreak{}HE\_\allowbreak{}RSV} & 1,984 & {\cellcolor[HTML]{006837}} \color[HTML]{F1F1F1} 0.1768 & {\cellcolor[HTML]{006837}} \color[HTML]{F1F1F1} 0.2222 & {\cellcolor[HTML]{D6D6D6}} \color[HTML]{000000}  & {\cellcolor[HTML]{D6D6D6}} \color[HTML]{000000}  \\
\color[HTML]{011773} \bfseries \texttt{Google\_\allowbreak{}SAI-\allowbreak{}RSVEns} & 4,576 & {\cellcolor[HTML]{87CB67}} \color[HTML]{000000} 0.1888 & {\cellcolor[HTML]{006837}} \color[HTML]{F1F1F1} 0.2224 & {\cellcolor[HTML]{006837}} \color[HTML]{F1F1F1} 0.2530 & {\cellcolor[HTML]{006837}} \color[HTML]{F1F1F1} 0.2882 \\
\texttt{RSVHub-\allowbreak{}ensemble} & 4,576 & {\cellcolor[HTML]{66BD63}} \color[HTML]{000000} 0.1864 & {\cellcolor[HTML]{2DA155}} \color[HTML]{000000} 0.2309 & {\cellcolor[HTML]{A0D669}} \color[HTML]{000000} 0.2899 & {\cellcolor[HTML]{9BD469}} \color[HTML]{000000} 0.3358 \\
\texttt{CEPH-\allowbreak{}Rtrend\_\allowbreak{}rsv} & 4,576 & {\cellcolor[HTML]{F57547}} \color[HTML]{000000} 0.2182 & {\cellcolor[HTML]{E14430}} \color[HTML]{000000} 0.2821 & {\cellcolor[HTML]{F36B42}} \color[HTML]{000000} 0.3430 & {\cellcolor[HTML]{F57748}} \color[HTML]{000000} 0.4080 \\
\texttt{CFA\_\allowbreak{}Pyrenew-\allowbreak{}Pyrenew\_\allowbreak{}H\_\allowbreak{}RSV} & 2,234 & {\cellcolor[HTML]{FDB365}} \color[HTML]{000000} 0.2121 & {\cellcolor[HTML]{B71126}} \color[HTML]{F1F1F1} 0.2891 & {\cellcolor[HTML]{D6D6D6}} \color[HTML]{000000}  & {\cellcolor[HTML]{D6D6D6}} \color[HTML]{000000}  \\
\texttt{RSVHub-\allowbreak{}baseline} & 4,576 & {\cellcolor[HTML]{A50026}} \color[HTML]{F1F1F1} 0.2315 & {\cellcolor[HTML]{A50026}} \color[HTML]{F1F1F1} 0.2916 & {\cellcolor[HTML]{A50026}} \color[HTML]{F1F1F1} 0.3603 & {\cellcolor[HTML]{A50026}} \color[HTML]{F1F1F1} 0.4357 \\

%% file: Sections/7-prompt_tables.tex


\section{Problem Statement Prompts}
\label{sec:problem_statement}

Each \ourmethod experiment begins with a shared \emph{problem statement} that describes the forecasting task, target, available input data, evaluation procedure and metric; and specifies the required output format and structure.
The problem statement for all three pathogens follows a common template; only the target variable, auxiliary data sources, and data-augmentation guidance differ across pathogens.
Below we reproduce the canonical prompt for influenza in full, followed by a summary of the infection-specific variations for COVID-19 and RSV.

\subsection{Canonical Problem Statement (influenza)}
\label{sec:problem_statement_flu}

\begin{tcolorbox}[
  colback=white,
  colframe=black!30,
  coltitle=black,
  colbacktitle=black!10,
  fonttitle=\ttfamily\small,
  title=Problem Statement: influenza,
  breakable,
  enhanced,
  left=6pt, right=6pt, top=4pt, bottom=4pt
]
\small

\paragraph{Overview.}
Modelers at the CDC's FluSight Forecast Hub are tasked with producing unconditional probabilistic forecasts for weekly influenza hospitalizations across the United States.
This is a critical public health task, as these forecasts help inform resource allocation and policy decisions during flu season.
The challenge is to create forecasts that characterize uncertainty across all reasonable future scenarios, not just a limited set of conditions.

This task aims to develop a superhuman forecasting model for this problem.
The input data is a time series of historical flu hospitalizations and related signals.
The output must be a set of quantile forecasts for all 50 states, Washington DC, and Puerto Rico, across multiple time horizons.
The goal is to develop a model that is more accurate and better calibrated than those produced by leading human experts.

\paragraph{Problem Statement \& Deliverable.}
Your primary task is to create a forecasting model that predicts \textbf{probabilistic forecasts} of \textbf{Total Influenza Hospital Admissions} for every US state and jurisdiction.
The goal is to create a model that achieves the lowest possible \textbf{Weighted Interval Score (WIS)} over a rolling-window evaluation.

Your deliverable is a single Python function, \texttt{fit\_and\_predict\_fn}, that takes in training and test data and returns a pandas DataFrame containing the required quantile predictions.

\paragraph{Function Signature \& Output Requirements.}
The forecasting model must be encapsulated within a function named \texttt{fit\_and\_predict\_fn} with the following signature:
\begin{verbatim}
def fit_and_predict_fn(
    train_x: pd.DataFrame,
    train_y: pd.Series,
    test_x: pd.DataFrame,
) -> pd.DataFrame:
    return test_y_hat_quantiles
\end{verbatim}
The returned DataFrame must have its index match the input \texttt{test\_x}, columns named by quantile (e.g.\ \texttt{quantile\_0.01}, \texttt{quantile\_0.5}, \texttt{quantile\_0.975}), and the predicted quantiles for any given row must be \textbf{monotonically increasing}.

\paragraph{Dataset Description.}
The following data objects are available:
\begin{itemize}[nosep]
  \item \textbf{Primary Training Data:}
    \texttt{train\_x} (historical features), \texttt{train\_y} (historical target: Total Influenza Admissions).
  \item \textbf{Historical Augmentation Data:}
    \texttt{ilinet\_hhs}, \texttt{ilinet}, \texttt{ilinet\_state} --- DataFrames containing ${\sim}20$ years of historical Influenza-Like Illness (ILI) data, available only for dates before 2022-10-15.
  \item \textbf{Reference \& Example Data:}
    \texttt{locations} (geographic/population data),
    \texttt{sample\_submission\_df} (correct output format),
    \texttt{example\_train\_x}, \texttt{example\_train\_y}, \texttt{example\_test\_x} (small example DataFrames).
\end{itemize}

\paragraph{Feature Definitions.}
\texttt{target\_end\_date}: Saturday of the epiweek;
\texttt{location\_name}: full state name;
\texttt{location}: FIPS code;
\texttt{population}: total population;
\texttt{Total Influenza Admissions}: the target variable (available from late 2020 onward).

\paragraph{Augmenting Training Data with Historical ILINet Data.}
The core challenge is the limited history of the target variable. To overcome this, ${\sim}20$ seasons of historical ILINet data are provided. While not the same target, it is highly correlated and captures the essential seasonal dynamics of influenza. Two strategies are suggested:

\textbf{Strategy 1 --- Standardize and Combine:}
(1)~Apply a standardization method to both datasets independently to make the ``shape'' of the seasons comparable.
(2)~Treat the standardized historical ILINet data as additional, independent flu seasons and append them to the training data.
(3)~Train the model on this combined ``library'' of seasons.

\textbf{Strategy 2 --- Learn a Transformation:}
(1)~Identify the date range where both the target and the historical ILINet data overlap.
(2)~Use this overlapping period to learn a statistical transformation (e.g.\ linear regression, quantile mapping) that maps the ILINet data onto the same scale as Total Influenza Admissions.
(3)~Apply this transformation to the entire 20-year history to create a ``synthetic'' history for the target variable.
(4)~Train the final model on this augmented training set.

\paragraph{Key Considerations.}
\begin{itemize}[nosep]
  \item \textbf{Time Series Awareness:} Handle seasonality, trends, and lags appropriately.
  \item \textbf{Calibration:} Ensure predicted quantiles are well-calibrated.
  \item \textbf{Logging:} Configure models to be quiet during training (\texttt{verbose=0}). Do not suppress critical warnings or tracebacks.
\end{itemize}

\paragraph{Detailed Instructions.}
An expert has instructed the model to implement a specific method for this forecasting task. Minor improvements to the method are permitted, but the original core principles \textbf{must} be maintained. Before writing code, the model must explicitly list the 3--4 core principles of the prescribed method in a comment block, then implement accordingly.
\end{tcolorbox}
\vspace{6pt}

\subsection{Pathogen-specific Variations}
\label{sec:problem_statement_variations}

The COVID-19 and RSV problem statements share the same function signature, output format, evaluation metric (WIS), and key considerations as the influenza prompt above.
The differences are summarized below.

\begin{tcolorbox}[
  colback=white,
  colframe=black!30,
  coltitle=black,
  colbacktitle=black!10,
  fonttitle=\ttfamily\small,
  title=COVID-19 Variations,
  breakable,
  enhanced,
  left=6pt, right=6pt, top=4pt, bottom=4pt
]
\small
\begin{itemize}[nosep]
  \item \textbf{Target variable:} \texttt{Total COVID-19 Admissions} (replaces \texttt{Total Influenza Admissions}).
  \item \textbf{Forecast Hub:} CDC CovidHub Forecast Hub (replaces FluSight).
  \item \textbf{No auxiliary historical data:} The COVID-19 prompt does \emph{not} include ILINet augmentation data or data-augmentation strategies. The model relies solely on the primary training data (\texttt{train\_x}, \texttt{train\_y}) and the reference/example data.
  \item \textbf{Simpler dataset:} No \texttt{ilinet\_*} DataFrames are provided.
\end{itemize}
\end{tcolorbox}
\vspace{4pt}

\begin{tcolorbox}[
  colback=white,
  colframe=black!30,
  coltitle=black,
  colbacktitle=black!10,
  fonttitle=\ttfamily\small,
  title=RSV Variations,
  breakable,
  enhanced,
  left=6pt, right=6pt, top=4pt, bottom=4pt
]
\small
\begin{itemize}[nosep]
  \item \textbf{Target variable:} \texttt{Total RSV Admissions} (replaces \texttt{Total Influenza Admissions}).
  \item \textbf{Core challenge --- Extreme Data Sparsity:} For most states, the RSV target is only available from late 2024 onward, making this the most data-constrained of the three tasks. The prompt explicitly frames data sparsity as the central challenge and emphasizes ``transfer of knowledge'' from auxiliary datasets.
  \item \textbf{Richer auxiliary data:} In addition to the ${\sim}20$-year ILINet history, the RSV prompt provides:
    \begin{itemize}[nosep]
      \item \texttt{Total COVID-19 Admissions} and \texttt{Total Influenza Admissions} columns in \texttt{train\_x} (longer history than the RSV target).
      \item Weekly percentage of ED visits due to RSV from the NSSP dataset (available from 2022 onward, but \emph{not} for all locations, e.g.\ Puerto Rico and Missouri are missing).
    \end{itemize}
  \item \textbf{Data-augmentation strategies} (replacing the ILI-focused strategies in the flu prompt):
    \begin{itemize}[nosep]
      \item \textbf{Strategy 1 --- Transfer Learning from NSSP:} Learn a mapping between the NSSP signal and the RSV target in their overlap period, then apply it to create a synthetic history.
      \item \textbf{Strategy 2 --- Multi-Target Modeling:} Use the COVID-19, flu, and NSSP columns as features and learn cross-correlations to inform RSV forecasts.
    \end{itemize}
\end{itemize}
\end{tcolorbox}
\vspace{6pt}

%% file: Sections/8-model_inventories.tex
\section{Candidate Model Inventories}
\label{sec:model_inventories}

\begingroup\small
\setlength{\tabcolsep}{4pt}\setlength{\aboverulesep}{0pt}\setlength{\belowrulesep}{0pt}
\renewcommand{\arraystretch}{1.4}
\begin{longtable}{>{\raggedright\arraybackslash}m{5.5cm}
                  >{\centering\arraybackslash}m{2.5cm}>{\centering\arraybackslash}m{1.5cm}
                  >{\centering\arraybackslash}m{1.5cm}>{\centering\arraybackslash}m{1.8cm}
                  >{\centering\arraybackslash}m{1.8cm}}
    \caption{Inventory of all \ourmethod\ influenza candidate models evaluated during model selection for the 2025–26 season (one row per \ourmethod run). Validation and Test are \ourmethod’s internal held-out selection scores on the periods illustrated in Fig.~\ref{fig:fig1}; the selection score = validation $+\,2\times$ test (lower is better), and the three score columns are shaded independently green (best) to red (worst). Models are ranked by ascending selection score. Models prefixed by * were included in \gflumodel~ submitted to FluSight; \ding{51} in the GoogleHub column marks models recorded on the Google Research GitHub.}
    \label{tab:flu_inventory}\\
    \toprule
    \textbf{Model} & \textbf{Type} & \textbf{Validation} & \textbf{Test} & \textbf{Selection score} & \textbf{GoogleHub} \\
    \midrule \endfirsthead
    \toprule
    \textbf{Model} & \textbf{Type} & \textbf{Validation} & \textbf{Test} & \textbf{Selection score} & \textbf{GoogleHub} \\
    \midrule \endhead
    \bottomrule \endlastfoot
    \input{tables/internal_flu_inventory.tex}
\end{longtable}
\endgroup

\begingroup\small
\setlength{\tabcolsep}{4pt}\setlength{\aboverulesep}{0pt}\setlength{\belowrulesep}{0pt}
\renewcommand{\arraystretch}{1.4}
\begin{longtable}{>{\raggedright\arraybackslash}m{5.5cm}
                  >{\centering\arraybackslash}m{2.5cm}>{\centering\arraybackslash}m{1.5cm}
                  >{\centering\arraybackslash}m{1.5cm}>{\centering\arraybackslash}m{1.8cm}
                  >{\centering\arraybackslash}m{1.8cm}}
    \caption{Inventory of all \ourmethod\ COVID-19 candidate models evaluated during model selection for the 2025–26 season (one row per \ourmethod run). Validation and Test are \ourmethod’s internal held-out selection scores on the periods illustrated in Fig.~\ref{fig:fig1}; the the selection score = validation $+\,2\times$ test (lower is better), and the three score columns are shaded independently green (best) to red (worst). Models are ranked by ascending selection score. Models prefixed by * were included in \gcovidmodel~ submitted to COVIDHub; \ding{51} in the GoogleHub column marks models recorded on the Google Research GitHub.}
    \label{tab:covid_inventory}\\
    \toprule
    \textbf{Model} & \textbf{Type} & \textbf{Validation} & \textbf{Test} & \textbf{Selection score} & \textbf{GoogleHub} \\
    \midrule \endfirsthead
    \toprule
    \textbf{Model} & \textbf{Type} & \textbf{Validation} & \textbf{Test} & \textbf{Selection score} & \textbf{GoogleHub} \\
    \midrule \endhead
    \bottomrule \endlastfoot
    \input{tables/internal_covid_inventory.tex}
\end{longtable}
\endgroup

\begingroup\small
\setlength{\tabcolsep}{4pt}\setlength{\aboverulesep}{0pt}\setlength{\belowrulesep}{0pt}
\renewcommand{\arraystretch}{1.4}
\begin{longtable}{>{\raggedright\arraybackslash}m{5.5cm}
                  >{\centering\arraybackslash}m{2.5cm}>{\centering\arraybackslash}m{1.5cm}
                  >{\centering\arraybackslash}m{1.5cm}>{\centering\arraybackslash}m{1.8cm}
                  >{\centering\arraybackslash}m{1.8cm}}
    \caption{Inventory of all \ourmethod RSV candidate models evaluated during model selection for the 2025–26 season (one row per \ourmethod run). 
    Validation and Test are \ourmethod’s internal held-out selection scores on the periods illustrated in Fig.~\ref{fig:fig1}; the the selection score = validation $+\,2\times$ test (lower is better), and the three score columns are shaded independently green (best) to red (worst). Models are ranked by ascending selection score. Models prefixed by * were included in \grsvmodel~ submitted to RSVHub; \ding{51} in the GoogleHub column marks models recorded on the Google Research GitHub.}
    \label{tab:rsv_inventory}\\
    \toprule
    \textbf{Model} & \textbf{Type} & \textbf{Validation} & \textbf{Test} & \textbf{Selection score} & \textbf{GoogleHub} \\
    \midrule \endfirsthead
    \toprule
    \textbf{Model} & \textbf{Type} & \textbf{Validation} & \textbf{Test} & \textbf{Selection score} & \textbf{GoogleHub} \\
    \midrule \endhead
    \bottomrule \endlastfoot
    \input{tables/internal_rsv_inventory.tex}
\end{longtable}
\endgroup

%% file: tables/internal_flu_inventory.tex
*\texttt{G-\allowbreak{}LANL\_\allowbreak{}DBM x UMass\_\allowbreak{}Flusion} & Recombination & {\cellcolor[HTML]{C7E77F}} \color[HTML]{000000} 59.3 & {\cellcolor[HTML]{08773F}} \color[HTML]{F1F1F1} 42.1 & {\cellcolor[HTML]{006837}} \color[HTML]{F1F1F1} 143.5 & \color[HTML]{17B248} \bfseries \ding{51} \\
*\texttt{G-\allowbreak{}LANL\_\allowbreak{}DBM x LANL\_\allowbreak{}Inferno} & Recombination & {\cellcolor[HTML]{F1F9AC}} \color[HTML]{000000} 63.1 & {\cellcolor[HTML]{097940}} \color[HTML]{F1F1F1} 42.2 & {\cellcolor[HTML]{0D8044}} \color[HTML]{F1F1F1} 147.5 & \color[HTML]{17B248} \bfseries \ding{51} \\
*\texttt{G-\allowbreak{}CU\_\allowbreak{}SIRS x UVA\_\allowbreak{}Gaussian\_\allowbreak{}processes} & Recombination & {\cellcolor[HTML]{CFEB85}} \color[HTML]{000000} 59.9 & {\cellcolor[HTML]{30A356}} \color[HTML]{000000} 45.4 & {\cellcolor[HTML]{17934E}} \color[HTML]{000000} 150.7 & \color[HTML]{17B248} \bfseries \ding{51} \\
\texttt{G-\allowbreak{}LANL\_\allowbreak{}DBM x UGuelph\_\allowbreak{}CompositeCurve} & Recombination & {\cellcolor[HTML]{D3EC87}} \color[HTML]{000000} 60.2 & {\cellcolor[HTML]{3FAA59}} \color[HTML]{000000} 46.0 & {\cellcolor[HTML]{219C52}} \color[HTML]{000000} 152.2 &  \\
*\texttt{G-\allowbreak{}NU-\allowbreak{}PGF\_\allowbreak{}FLUH [G3Pro pkg n2027]} & Scientist-inspired & {\cellcolor[HTML]{DDF191}} \color[HTML]{000000} 61.1 & {\cellcolor[HTML]{39A758}} \color[HTML]{000000} 45.8 & {\cellcolor[HTML]{279F53}} \color[HTML]{000000} 152.7 & \color[HTML]{17B248} \bfseries \ding{51} \\
*\texttt{G-\allowbreak{}NU-\allowbreak{}PGF\_\allowbreak{}FLUH [G3 es n838]} & Scientist-inspired & {\cellcolor[HTML]{FFFEBE}} \color[HTML]{000000} 64.8 & {\cellcolor[HTML]{219C52}} \color[HTML]{000000} 44.7 & {\cellcolor[HTML]{36A657}} \color[HTML]{000000} 154.2 & \color[HTML]{17B248} \bfseries \ding{51} \\

%% file: tables/internal_covid_inventory.tex
*\texttt{G-\allowbreak{}UM-\allowbreak{}DeepOutbreak} & Scientist-inspired & {\cellcolor[HTML]{006837}} \color[HTML]{F1F1F1} 34.5 & {\cellcolor[HTML]{006837}} \color[HTML]{F1F1F1} 26.2 & {\cellcolor[HTML]{006837}} \color[HTML]{F1F1F1} 86.9 & \color[HTML]{17B248} \bfseries \ding{51} \\
*\texttt{G-\allowbreak{}MOBS-\allowbreak{}GLEAM\_\allowbreak{}COVID} & Scientist-inspired & {\cellcolor[HTML]{0B7D42}} \color[HTML]{F1F1F1} 35.1 & {\cellcolor[HTML]{87CB67}} \color[HTML]{000000} 27.8 & {\cellcolor[HTML]{51B35E}} \color[HTML]{000000} 90.7 & \color[HTML]{17B248} \bfseries \ding{51} \\
*\texttt{G-\allowbreak{}DeepResearch\_\allowbreak{}CounterfactualSimulation} & AI-led & {\cellcolor[HTML]{1E9A51}} \color[HTML]{000000} 35.9 & {\cellcolor[HTML]{8ECF67}} \color[HTML]{000000} 27.9 & {\cellcolor[HTML]{73C264}} \color[HTML]{000000} 91.7 & \color[HTML]{17B248} \bfseries \ding{51} \\
*\texttt{G-\allowbreak{}NEU\_\allowbreak{}ISI-\allowbreak{}AdaptiveEnsemble} & Scientist-inspired & {\cellcolor[HTML]{BFE47A}} \color[HTML]{000000} 39.1 & {\cellcolor[HTML]{148E4B}} \color[HTML]{000000} 26.7 & {\cellcolor[HTML]{89CC67}} \color[HTML]{000000} 92.5 & \color[HTML]{17B248} \bfseries \ding{51} \\
*\texttt{G-\allowbreak{}UMass-\allowbreak{}gbqr} & Scientist-inspired & {\cellcolor[HTML]{33A456}} \color[HTML]{000000} 36.3 & {\cellcolor[HTML]{B5DF74}} \color[HTML]{000000} 28.3 & {\cellcolor[HTML]{96D268}} \color[HTML]{000000} 92.9 & \color[HTML]{17B248} \bfseries \ding{51} \\
\texttt{G-\allowbreak{}OHT\_\allowbreak{}JHU-\allowbreak{}nbxd} & Scientist-inspired & {\cellcolor[HTML]{199750}} \color[HTML]{000000} 35.8 & {\cellcolor[HTML]{DCF08F}} \color[HTML]{000000} 28.8 & {\cellcolor[HTML]{A5D86A}} \color[HTML]{000000} 93.4 &  \\

%% file: tables/internal_rsv_inventory.tex
*\texttt{G-\allowbreak{}general\_\allowbreak{}instruction\_\allowbreak{}2} & AI-led & {\cellcolor[HTML]{006837}} \color[HTML]{F1F1F1} 21.7 & {\cellcolor[HTML]{17934E}} \color[HTML]{000000} 3.2 & {\cellcolor[HTML]{006837}} \color[HTML]{F1F1F1} 28.1 & \color[HTML]{17B248} \bfseries \ding{51} \\
*\texttt{G-\allowbreak{}general\_\allowbreak{}instruction\_\allowbreak{}lgbm} & AI-led & {\cellcolor[HTML]{279F53}} \color[HTML]{000000} 23.2 & {\cellcolor[HTML]{006837}} \color[HTML]{F1F1F1} 2.5 & {\cellcolor[HTML]{016A38}} \color[HTML]{F1F1F1} 28.2 & \color[HTML]{17B248} \bfseries \ding{51} \\
\texttt{G-\allowbreak{}updated general instruction} & AI-led & {\cellcolor[HTML]{87CB67}} \color[HTML]{000000} 24.9 & {\cellcolor[HTML]{148E4B}} \color[HTML]{000000} 3.1 & {\cellcolor[HTML]{3FAA59}} \color[HTML]{000000} 31.1 &  \\
\texttt{G-\allowbreak{}adversarial\_\allowbreak{}domain\_\allowbreak{}adapt} & AI-led & {\cellcolor[HTML]{1B9950}} \color[HTML]{000000} 23.0 & {\cellcolor[HTML]{96D268}} \color[HTML]{000000} 4.6 & {\cellcolor[HTML]{66BD63}} \color[HTML]{000000} 32.2 &  \\
\texttt{G-\allowbreak{}CFA\_\allowbreak{}Pyrenew-\allowbreak{}Pyrenew\_\allowbreak{}E\_\allowbreak{}RSV} & Scientist-inspired & {\cellcolor[HTML]{EEF8A8}} \color[HTML]{000000} 27.5 & {\cellcolor[HTML]{1B9950}} \color[HTML]{000000} 3.3 & {\cellcolor[HTML]{A5D86A}} \color[HTML]{000000} 34.1 &  \\
\texttt{G-\allowbreak{}UGA\_\allowbreak{}flucast-\allowbreak{}INFLAenza} & Scientist-inspired & {\cellcolor[HTML]{EBF7A3}} \color[HTML]{000000} 27.4 & {\cellcolor[HTML]{30A356}} \color[HTML]{000000} 3.5 & {\cellcolor[HTML]{ABDB6D}} \color[HTML]{000000} 34.4 &  \\
\texttt{G-\allowbreak{}CFA\_\allowbreak{}Pyrenew-\allowbreak{}Pyrenew\_\allowbreak{}HE\_\allowbreak{}RSV} & Scientist-inspired & {\cellcolor[HTML]{F1F9AC}} \color[HTML]{000000} 27.6 & {\cellcolor[HTML]{3CA959}} \color[HTML]{000000} 3.6 & {\cellcolor[HTML]{B5DF74}} \color[HTML]{000000} 34.8 &  \\